%% file: main.tex
\documentclass{article} %
\usepackage{arxiv,times}
\usepackage{todonotes}
\input{math_commands.tex} %
\usepackage{hyperref}
\usepackage{cleveref}
\usepackage{url}
\usepackage{thmtools}
\usepackage{booktabs}
\usepackage{wrapfig}
\usepackage{float}
\title{Learning from Failures: When Contrastive Post-Training Beats Success Conditioning}
\author{Soutrik Sarangi\\
Microsoft Research\\
\texttt{soutriksarangi14@gmail.com} \\
\And
Gaurav Sinha \\
Microsoft Research \\
\texttt{gauravsinha@microsoft.com} \\
}
\usepackage{xcolor}

\newcommand{\add}[1]{#1}
\preprintfinalcopy %
\begin{document}
\maketitle

\begin{abstract}
Post-training a language model from rollouts with binary outcomes has two standard recipes: supervised fine-tuning (SFT) on the successful rollouts (success conditioning), and pairwise contrastive learning on success--failure pairs. In practice contrastive and SFT losses are combined, and the combination is often reported to work best, but there is little theory of when and why. We analyze the combined family: a margin-based contrastive term plus a weighted success conditioning term on successful rollouts using a contextual bandit setup with binary rewards. We characterize when adding the contrastive term improves on success conditioning, from expected reward to ranking-based metrics (commonly used in generative retrieval). \add{
We study both unweighted and DPO-style reference-reweighted variants of contrastive losses, and characterize how the base policy's alignment with action rewards determines their relative performance.} We validate our theoretical results via synthetic experiments and also in the real-world context of generative retrieval. 
\end{abstract}

\section{Introduction}
\label{sec:intro}

In reinforcement learning with binary outcome feedback, each trajectory receives a reward of $1$ indicating success or $0$ indicating failure. A policy can be trained by maximizing likelihood on successful trajectories alone or by contrasting successful and failed trajectories. Success-only fine-tuning has been studied for language
models~\citep{Zelikman2022,Singh2024,russo2026success}. Related works on preference-optimization methods learn from comparisons between
preferred and dispreferred responses~\citep{dpo2023,simpo,Zhao2023,Yuan2023RRHF,Hong2024ORPO,Xu2024CPO}. 
When we have explicit binary feedback (instead of only preferences), combining the two objectives of contrastive learning and success-only fine-tuning has also been of interest, yielding gains in applications such as reasoning \citep{pang2024iterativereasoningpreferenceoptimization} and generative retrieval \citep{li2023ltrgr}, motivating a general question: when does a combination of the two ideas, i.e., maximizing likelihood of successful trajectories and contrasting successful and failed trajectories, improve a policy beyond training on successful trajectories alone?

Maximizing likelihood of successful trajectories (also known as success conditioning)~\citep{russo2026success} is already known to provably improve upon the base data collection policy (Corollary $4.5$ in \citep{russo2026success}). At the same time, prior work such as \citet{dpo2023} demonstrates how for applications such as language model alignment, one could contrast a positive and negative answer via preference models (i.e., positive preferred over negative) and obtain strict improvements over the base policy. However, it is unclear if the techniques in the above literature can be generalized to  characterize when the exact minimizer of a linear combination of success conditioning loss and contrastive loss can outperform success conditioning (or even the base policy) on the same rollout distribution. This distinction matters even without sampling or optimization error: a margin-based contrastive loss can rank actions correctly yet assign less probability to better actions, and improvement over the base policy need not imply improvement over success conditioning. Motivated by the practical success of such combined losses, e.g., \citep{pang2024iterativereasoningpreferenceoptimization, li2023ltrgr}, we aim to theoretically understand and characterize their performance.

We study this comparison for binary rewards in contextual bandits and finite-horizon MDPs. A fixed base policy $\base$ generates the data, and its success-conditioned policy $\scond$ is the population solution to maximum-likelihood training on successful trajectories. \add{We study two established variants of hinge-based contrastive learning combined with success conditioning: (a) \unweighted, which applies a pairwise hinge to policy log-likelihoods, as in the ranking component of SLiC-HF \citep{Zhao2023}, and (b) \reweighted, which applies a hinge to log-likelihood ratios relative to the base policy, as covered by the generalized preference-optimization framework \citep{Tang2024}. Both objectives include a maximum-likelihood term on successful rollouts, weighted by $\lambda$, and use a contrastive margin $\delta$.} This re-weighting amounts to changing the margin $\delta$ in the usual contrastive loss by the gap between the log-probabilities of the successful and failed trajectories under $\base$, making the margin more informed of the ranking induced by the base policy $\base$. We characterize how these choices affect performance of the learned policies when measured under typical RL metrics such as ``value'' and other ``ranking'' based metrics. Our results identify conditions for guaranteed improvement over the base policy and the success conditioning policy and establish the limits of these guarantees, as summarized below.

\vspace{-3pt}
\subsection{Our Contributions}
\vspace{-3pt}
Let $\base$ denote the base policy, $\base^+$ its success-conditioned
policy, and $\piunw,\pirw$ the population
minimizers of the unweighted and reweighted objectives, respectively. Let $\delta\geq 0$ be the margin and $\lambda\geq0$ be the scalar multiplier of the success conditioning loss. We refer the reader to Section \ref{sec:setup} for some of the RL and contextual bandit terminology and notation used below. 

\textbf{(1) Improvement over success conditioning -} For contextual bandits, we establish thresholds $\delta^\star$,  $\lambda^\star$ such that for $\delta\geq \delta^\star$ and $\lambda\geq \lambda^\star$ and for every state $\state$, $V_{\piunw}(s),\, V_{\pirw}(s) \ge V_{\base^+}(s)$. The inequalities are strict when $\operatorname{Var}_{a\sim\base^+(\cdot\mid s)}[\Q(s,a)]>0$. We also construct counterexamples with small values of the margin $\delta$ to show that neither comparison holds unconditionally.

\textbf{(2) The role of reference-policy reweighting -} For finite-horizon MDPs, for every $\lambda, \delta \geq 0$, the \reweighted policy $\pirw$ satisfies $V_{\pirw}(s)\ge V_{\base}(s)$. In contrast, for the \unweighted policy $\piunw$ we construct counterexamples where such a guarantee does not hold. In contextual bandits, we further show $V_{\pirw}(s) \ge V_{\piunw}(s)$ for every state $\state$, when $\base(\cdot\mid s)$ and $Q(s,\cdot)$ have the same ordering; the inequality reverses when their orderings are opposite.

\textbf{(3) Limits of generalization to finite-horizon MDPs -} We construct a finite-horizon MDP for which $V_{\base^+}(\mu)> \max\!\left\{ V_{\piunw}(\mu), V_{\pirw}(\mu) \right\}$, for every finite $\lambda\ge0$ and every $\delta\ge0$. A separate construction shows that even when the orderings of $\base(\cdot\mid s)$ and $Q(s,\cdot)$ are the same, the comparison between $\pirw$ and $\piunw$ (mentioned for contextual bandits above) need not extend to MDPs.

\textbf{(4) Ranking guarantees distinct from value guarantees -} We show that for sufficiently large $\lambda, \delta$, $\pirw$ also improves over $\base^+$ in other metrics such as precision and recall that are relevant for applications such as ranking and retrieval. This aligns with our experiments in Section \ref{sec:expt} and the performance gains that have been observed for such combined losses in the retrieval literature \citep{li2023ltrgr}. 

\textbf{(5) Empirical evidence -}
We perform semi-synthetic bandit and MS~MARCO generative-retrieval experiments in Section \ref{sec:expt} and examine the effects of $\lambda$, $\delta$, and the ordering of base probabilities relative to rewards. We demonstrate gains from combined objectives and distinguish the objective preferred for expected reward
from that preferred for retrieval ranking.
\vspace{-3pt}
\subsection{Related Work}
\vspace{-3pt}
\paragraph{Learning from successful and failed outcomes.}
STaR and ReST$^{\mathrm{EM}}$ train on successful model-generated
solutions \citep{Zelikman2022,Singh2024}, while
\citet{russo2026success} characterizes success conditioning as a conservative policy-improvement update. Negative examples can improve probability redistribution \citep{Tajwar2024} and, when constructed at the
step level, credit assignment \citep{Setlur2024}.
In robotics, failed demonstrations constrain inverse
reinforcement learning \citep{Shiarlis2016}, while
GCR and TW-CRL use success--failure contrasts to learn rewards
\citep{Biza2025, Li2025}. These works establish
the utility of learning from both outcomes, but do not
characterize the population minimizers of our combined
policy objective relative to success conditioning.
We provide this comparison directly, without an intermediate
learned reward model: we identify conditions for strict
improvement, establish the role of the base policy, and
construct counterexamples showing why bandit guarantees
need not extend to sequential decisions.

\paragraph{Contrastive objectives and ranking.}
DPO uses reference-policy likelihood ratios
\citep{dpo2023}, SLiC-HF combines a likelihood-margin
loss with cross-entropy regularization \citep{Zhao2023},
and Iterative Reasoning Preference Optimization adds
likelihood training on correct responses to DPO
\citep{pang2024iterativereasoningpreferenceoptimization}. GPO unifies several preference
losses and analyzes their regularization
\citep{Tang2024}. Our contribution is not the
combination of these ingredients, but explicit conditions
under which reference reweighting and the contrastive
margin yield higher value than success-only training.
In generative retrieval, MINDER learns multiple identifier
views \citep{minder2023}, and LTRGR supplements generation
with ranking supervision \citep{li2023ltrgr}.
Beyond these application-specific methods, we establish
ranking-inversion guarantees and separate them from
expected-reward guarantees, explaining why the loss
preferred for policy value need not be preferred for retrieval.

\vspace{-5pt}
\section{Preliminaries}
\vspace{-5pt}
\label{sec:setup}
In this section, we introduce key definitions and results used throughout the rest of the paper. The set $\{1,\ldots, n\}$ is denoted by $[n]$.

\textbf{Finite-horizon MDPs.}
A finite-horizon MDP is a tuple
$(\stateset,\actset,\tkern,S_0,\initlaw,\termset)$,
where $\stateset,\actset$ are finite state and action sets,
$S_0\sim\initlaw$, $\tkern(\state'\mid\state,\act)$ is the
transition probability, and $\termset\subseteq\stateset$ is the
set of terminal states. Each episode terminates within $\horizon$
decisions, producing a trajectory
$\traj=(\bigstate_0,\bigact_0,\ldots,
\bigstate_{h-1},\bigact_{h-1},\bigstate_h)$
with $h\le\horizon$ and $\bigstate_h\in\termset$.
We write $\bigstate_i(\traj),\bigact_i(\traj)$ for its states
and actions. The episode receives a binary outcome
$\reward(\traj)\in\{0,1\}$ whose conditional distribution depends
only on the terminal state; temporal history
can be included in the state when necessary.

A policy $\pol(\cdot\mid\state)$ is a probability distribution
over $\actset$, and $\Prob_\pol$ denotes the trajectory law
induced by $\initlaw,\pol,\tkern$.
Its value and action-value functions are the success probabilities
when starting from $\state$, and when additionally taking
$\act$ as the first action before following $\pol$, respectively:
\[
\val{\pol}(\state)
=\Prob_\pol\!\left(\reward(\traj)=1
  \mid\bigstate_0=\state\right),
\qquad
\Q_\pol(\state,\act)
=\Prob_\pol\!\left(\reward(\traj)=1
  \mid\bigstate_0=\state,\bigact_0=\act\right).
\]
Both take values in $[0,1]$, and
$\val{\pol}(\initlaw)
=\E_{\state\sim\initlaw}\val{\pol}(\state)$.
For notational convenience, terminated trajectories may be padded
with dummy steps (with the same outcome) giving every
trajectory length $\horizon$.

\textbf{Contextual bandits -} A contextual bandit is the special case $\horizon=1$: each episode consists of one decision at a context $\bigstate_0\sim\initlaw$, so $\Q_\pol(\state,\act)=\Q(\state,\act)$ does not depend on the policy and $\val{\pol}(\state)=\sum_\act\pol(\act\mid\state)\Q(\state,\act)$. In the online version a learner interacts over $T$ rounds, observing a context, playing an action and receiving a binary reward, with the goal of estimating a policy $\pol^{T}$ of high value, or equivalently of small simple regret $\max_\pol\val{\pol}(\initlaw)-\val{\pol^{T}}(\initlaw)$.

\textbf{Offline contextual bandits -} In this setting we assume that the learner only gets to see the trajectories $\{(\state_t, \act_t, r_t) : t\in [T]\}$ generated using a prior data collection policy $\base$, i.e., each $a_t \sim \base(\cdot\mid \bigstate_t = \state_t)$. The learner can use these trajectories to estimate $\pol^T$ with a goal of minimizing the regret defined above.

\textbf{Language models as contextual bandits -}
Let $\vocab$ be a vocabulary of tokens. A Language Model (LM) is an auto-regressive policy $\pol(\bigresp\mid\bigctx)$ which 
when given an input text context $\state$ sampled from some distribution $\initlaw$, sequentially generates tokens $\resp_1,\dots,\resp_\horizon \in \vocab$ such that $\resp_h\sim \pol(\bigresp = \resp_h \mid \bigctx = (\state, \resp_{<h}))$. Given an input text $\state$ and the generated response $\bar\resp = (\resp_1,\ldots,\resp_\horizon)$, one can use a suitable reward model $\reward(\state, \bar\resp) \sim\mathrm{Bernoulli}(\Q(\state,\bar\resp))$ to measure the quality of the response as specified by some downstream application ~\citep{helpfulharmlessassistant}. There is a natural contextual bandit problem associated with an LM where the action set $\actset$ comprises all possible responses.

We call a state $\state\in \stateset$ \emph{non-degenerate} under  a policy $\pol$ if $\val{\pol}(\state)\in (0,1)$ and if the policy $\pol$ has full support over all actions at that state. Unless otherwise specified, we assume all states to be non-degenerate in the rest of the paper, terminal states exempt. The following definitions and observations apply both in the cases of MDPs and contextual bandits.

\begin{definition}[Reward conditioned policies]
\label{def:reward-conditioned-policies}
For a policy $\pol$, we define the success conditioned policy $\pi^+$ following \citep{russo2026success} as $\pol^+(\act\mid\state) = \Prob_\pol(\bigact_t(\traj)=\act\mid\bigstate_t(\traj)=\state, \reward(\traj)=1)$. In other words, it is the probability of selecting $\act$ from the state $\state$ under a given policy $\pol$ with the condition that $(\state, \act)$ belongs to a successful trajectory. Similarly, we can define the failure conditioned policy as $\pol^-(\act\mid\state) = \Prob_\pol(\bigact_t(\traj)=\act\mid\bigstate_t(\traj)=\state, \reward(\traj)=0)$.
\end{definition}

We now make some observations regarding the success and failure conditioned trajectories, similar to \citep{russo2026success}, stated in the following lemma.

\begin{restatable}[Closed form expressions for reward conditioned policies]{lemma}{lemsucfail}\label{lem:sucfail}
    For an MDP policy $\pol$ and a state $\state$ non-degenerate under $\pol$ we have:
    \begin{equation}\label{eq:cond}
\pol^+(\act\mid\state)=\frac{\pol(\act\mid\state)\,\Q_{\pol}(\state,\act)}{\val{\pol}(\state)},
\,
\pol^-(\act\mid\state)=\frac{\pol(\act\mid\state)\,\bigl(1-\Q_\pol(\state,\act)\bigr)}{1-\val{\pol}(\state)},
\end{equation}
\end{restatable}

The above closed form expressions are applications of Bayes' rule, proofs of which are provided in Appendix~\ref{app:proofs}. Moreover, we can also write $\pol^+$ and $\pol^-$ as minimizers of next-action prediction loss on successful and failed trajectories.

\begin{restatable}[Reward conditioning as minimizer of next-action prediction loss]{lemma}{lemsucfailnext}\label{lem:sucfailnext}
Let $\pol$ be a policy of a finite-horizon MDP. If $\val{\pol}(\initlaw)\in (0,1)$ then we have:
\begin{equation}\label{eq:condsft}
\begin{split}
\pol^{+}&\in\argmin_{\hat\pol}\Bigl\{-\E_{\traj\sim {\Prob_\pol}}\Bigl[\sum_{h=0}^{\horizon-1}\ln\hat\pol(\bigact_h(\traj)\mid\bigstate_h(\traj))\Bigm|\reward(\traj)=1\Bigr]\Bigr\},\\
\pol^{-}&\in\argmin_{\hat\pol}\Bigl\{-\E_{\traj\sim {\Prob_\pol}}\Bigl[\sum_{h=0}^{\horizon-1}\ln\hat\pol(\bigact_h(\traj)\mid\bigstate_h(\traj))\Bigm|\reward(\traj)=0\Bigr]\Bigr\}.
\end{split}
\end{equation}
where the minimization is over the space of all possible policies. Note that here the expectation is with respect to the distribution $\Prob_\pol$ on trajectories that arise as a result of sampling actions as per policy $\pi$ and transitioning states as per the state transition distribution $P$.
\end{restatable}

\textbf{Policy optimization via contrastive loss -} We consider MDPs with a base policy $\base$, from which trajectories are rolled out (with state transitions as per $P$) and data is collected. These observed rollouts give us state, action pairs from both successful and unsuccessful trajectories. Recent work in Reinforcement Learning (RL) and Supervised Fine-Tuning (SFT), such as~\citet{pang2024iterativereasoningpreferenceoptimization}, utilizes the collection of observed rollouts to further fine-tune the base policy $\pol$ to a new policy $\newpol$ via contrastive learning based approaches. Among such formulations, the Direct Preference Optimization (DPO) loss from \citep{dpo2023} and the triplet hinge loss from \citep{li2023ltrgr} have received considerable attention lately due to the performance improvements observed in practical settings. The DPO loss re-weighs the log-likelihoods of the optimizing policy, i.e., $\ln\newpol(\cdot\mid\state)$ with those of the base policy (i.e., $\ln\pol(\cdot\mid \state)$) in the contrastive term while the triplet loss directly uses the  \unweighted log-likelihoods of the optimizer. Following their approach, we now formally define the \unweighted and \reweighted losses as follows:
\begin{align}
    &\label{eq:loss:unw:traj}
    \lunw(\newpol, \pol)\coloneqq\mathop{\E}\limits_{\substack{(\state, \act_+)\sim\traj_+\\(\state, \act_-)\sim\traj_-}}\left[\marg + \ln\newpol(\act_-\mid\state) - \ln\newpol(\act_+\mid\state)\right]_+ - \lambda\mathop{\E}\limits_{(\state, \act_+)\sim\traj_+}\ln\newpol(\act_+\mid\state)\\
    &\label{eq:loss:rw:traj}
    \lrw(\newpol, \pol)\coloneqq\mathop{\mathlarger\E}\limits_{\substack{(\state, \act_+)\sim\traj_+\\(\state, \act_-)\sim\traj_-}}\left[\marg + \ln\frac{\newpol(\act_-\mid\state)}{\pol(\act_-\mid\state)} - \ln\frac{\newpol(\act_+\mid\state)}{\pol(\act_+\mid\state)}\right]_+ - \lambda\mathop{\E}\limits_{(\state, \act_+)\sim\traj_+}\ln\newpol(\act_+\mid\state)
\end{align}

We draw a successful rollout $\tau^+\sim P^\pi(\cdot\mid R=1)$.
For each decision step $t$, set $(s,a^+)=(S_t,A_t)$ and draw
$a^-\sim\pi^-(\cdot\mid s)$ independently of the successful
rollout conditional on $s$. Sum the hinge over these steps. Conditioning on the state of the step turns \cref{eq:loss:unw:traj,eq:loss:rw:traj} into the \emph{state-wise form}:
\begin{align}
    &\label{eq:loss:unw}
\lunw(\newpol,\pol)=\mathop{\E}\limits_{\state\sim\nu}\mathop{\E}\limits_{\substack{\act_+\sim\pol^{+}(\cdot\mid\state)\\\act_-\sim\pol^{-}(\cdot\mid\state)}}\hinge{\marg+\ln\newpol(\act_-\mid\state)-\ln\newpol(\act_+\mid\state)}-\lam\mathop{\E}\limits_{\state\sim\nu}\mathop{\E}\limits_{\act_+\sim\pol^{+}(\cdot\mid\state)}\ln\newpol(\act_+\mid\state),\\
    &\label{eq:loss:rw}
\lrw(\newpol,\pol)=\mathop{\E}\limits_{\state\sim\nu}\mathop{\E}\limits_{\substack{\act_+\sim\pol^{+}(\cdot\mid\state)\\\act_-\sim\pol^{-}(\cdot\mid\state)}}\hinge{\marg+\ln\frac{\newpol(\act_-\mid\state)}{\pol(\act_-\mid\state)}-\ln\frac{\newpol(\act_+\mid\state)}{\pol(\act_+\mid\state)}}-\lam\mathop{\E}\limits_{\state\sim\nu}\mathop{\E}\limits_{\act_+\sim\pol^{+}(\cdot\mid\state)}\ln\newpol(\act_+\mid\state),
\end{align}
where the states are distributed according to the success-conditioned visitation measure $\nu(\state)\coloneqq\occ^{+}(\state)=\occ_{\pol,\initlaw}(\state)\val{\pol}(\state)/\val{\pol}(\initlaw)$. We write $\mathcal L_{\unw,\state}(\newpol;\pol)$ and $\mathcal L_{\rw,\state}(\newpol;\pol)$ for the two \emph{state-wise} losses. Note that the objectives $\lunw, \lrw$ are positively weighted sums of state-wise losses with the common coefficients $(\lam,\marg)$, and the state losses depend on disjoint coordinates of $\newpol$. Hence, over all tabular policies, $\newpol$ minimizes \eqref{eq:loss:unw} (resp.\ \eqref{eq:loss:rw}) if and only if $\newpol(\cdot\mid\state)$ minimizes the state loss at every $\state$. This is more formally proved in \cref{lem:statewise} in Appendix~\ref{app:prelim}. We define $\piunw(\pol)$ and $\pirw(\pol)$ to be the minimizers (with respect to $\newpol$) of $\lunw(\newpol, \pol)$ (from \cref{eq:loss:unw}) and $\lrw(\newpol, \pol)$ (from \cref{eq:loss:rw}) respectively. When the infimum of a state loss is not attained (for $\lam>0$ this happens exactly when some action has $\Q_\pol(\state,\act)=0$), we use $\piunw, \pirw$ to denote the limit of the policies induced by a minimizing sequence of \cref{eq:loss:unw,eq:loss:rw}. For $\lam>0$ this limit exists, is unique, and is supported on the actions with $\Q_\pol(\state,\act)>0$ (Appendix~\ref{app:prelim}). For $\lam=0$ minimizers need not be unique, and statements at $\lam=0$ are made for every minimizer.

\vspace{-5pt}
\section{Main results}\label{sec:main}
\vspace{-5pt}

In this section we present our main technical results. Unless otherwise specified, we will assume a base policy $\base$. For this base policy, we can define the success and failure conditioning policies $\scond$ and $\fcond$ (as per Definition \ref{def:reward-conditioned-policies}) and the unweighted and re-weighted contrastive policies $\piunw(\base)$ and $\pirw(\base)$ as minimizers of losses $\lunw$ (\ref{eq:loss:unw}) and $\lrw$ (\ref{eq:loss:rw}) respectively. Since the base policy $\base$ is fixed for this section, for brevity, we will denote $\piunw(\base)$ and $\pirw(\base)$ by $\piunw$ and $\pirw$ respectively. The goal of this section is to compare the performance of $\piunw$ and $\pirw$ with that of $\scond$ and $\base$ under different metrics. Note that, from Corollary $4.5$ in \cite{russo2026success}, we already know that, in any given finite-horizon MDP, $V_{\scond}(s)\geq V_{\base}(s)$, for every state $s$. Here $V_\pol(s)$ is the value of a policy $\pol$ at state $s$ (defined in Section \ref{sec:setup}).

We now give a brief overview of our results in this section.

\textbf{(1)} $\piunw, \pirw$ vs.\ $\base$ - In Theorem \ref{thm:rw-base}, we obtain a result similar to Corollary $4.5$ of \cite{russo2026success} for $\pirw$, i.e., we show that, for any given episodic MDP and any fixed $\lambda, \delta \geq 0$, $V_{\pirw}(s) \geq V_{\base}(s)$ for every state $\state$. Unfortunately, this result does not hold in general for the unweighted policy $\piunw$. In \cref{ex:unw-base}, we construct a contextual bandit instance where this inequality fails. We strengthen this example in Lemma \ref{lem:unw-base}, by showing that under simple assumptions on the base policy $\base$, for all $\lambda, \delta$ sufficiently small, the value $V_{\piunw}$ of the corresponding unweighted policy is strictly smaller than $V_{\base}$.

\textbf{(2)} $\piunw, \pirw$ vs.\ $\scond$ - Interestingly for $\lambda, \delta$ large enough, in Theorem \ref{thm:beat-sft}, we show that for contextual bandits, both $V_{\piunw}(\state) \ge V_{\scond}(\state)$ and $V_{\pirw}(\state) \ge V_{\scond}(\state)$ for every $\state$, strictly if and only if $\Q(\state,\cdot)$ is non-constant on the positive actions.  Unfortunately, the situation for smaller values of $\lambda, \delta$ is not as clear. In fact, since Lemma \ref{ex:unw-base} exhibits $V_{\piunw} < V_{\base}$, for this same example we obtain that $V_{\piunw} < V_{\scond}$ since $V_{\base}(s) \leq V_{\scond}(s)$ for all states $\state$ (by Corollary $4.5$, \cite{russo2026success}). In Lemma \ref{lem:rw-sft}, we construct a contextual bandit instance where for any given $\lambda\geq 0$ and $\delta$ sufficiently small, we are able to show that $V_{\pirw} < V_{\scond}$. In summary, if one can appropriately tune $\lambda, \delta$, the contrastive loss based policies $\piunw, \pirw$ are better than $\scond$ in terms of value.

\textbf{(3)} $\piunw$ vs.\ $\pirw$ - In Theorem \ref{thm:comono}, for contextual bandit instances, we are able to identify situations where one of these policies always dominates the other. More precisely, when $\base$ and $Q_{\base}$ are aligned (anti-aligned respectively), then $V_{\pirw} \geq V_{\piunw}$ ($V_{\piunw} \geq V_{\pirw}$ respectively).
\vspace{-3pt}
\subsection{$\pirw$ dominates $\base$ in MDPs}
\vspace{-3pt}
In case of learning with the \reweighted loss $\lrw$, the ``re-weighting'' by $\ln\base$ introduces a form of an adaptive margin adjustment for the different $(\state, \act_+, \act_-)$ tuples. The contrastive term (first expectation in Eq.~\eqref{eq:loss:rw}) increases the margin of separation more in tuples where the base policy $\base$ already puts a higher weight on the successful action. In such cases, the higher margin induced by $\ln\base(\act_+\mid\state) - \ln\base(\act_-\mid\state)$ encourages a higher separation and makes the learned policy improve further over the base. This phenomenon is observed for any finite-horizon MDP. We state this formally in the following theorem. Proof can be found in Appendix \ref{app:proofs}.

\begin{restatable}{theorem}{thmrwbase}\label{thm:rw-base}
Let $\lambda, \delta$ be any scalars $\geq 0$ and $\lrw$ be the corresponding re-weighted loss (\cref{eq:loss:rw}). For any finite-horizon MDP with non-degenerate states, let $\pirw$ be the state-wise minimizer of $\lrw$. Then, we have $\val{\pirw}(\state)\ge\val{\base}(\state)$ for every state $\state$.
\end{restatable}

\subsection{Large $\lambda, \delta$ imply $\piunw, \pirw$ dominate $\scond$ in contextual bandits}
Note that, as $\lam \to\infty$, both the normalized losses $\lunw/\lam, \lrw/\lam$ converge to the success conditioned loss (top row in Eq.~\eqref{eq:condsft}) whose optimizer is the success conditioned policy $\scond$. 
The question however arises whether the contrastive terms in $\lunw, \lrw$ help in improving $\piunw, \pirw$ strictly beyond $\scond$.  We show that, for contextual bandits, for appropriately large $\lambda$ and $\delta$, the value function corresponding to both $\piunw, \pirw$ can improve strictly beyond the value function of $\scond$ in the case of contextual bandits. The result is formally stated below. Proof can be found in Appendix \ref{app:proofs}.

\begin{restatable}{theorem}{thmbeatsft}\label{thm:beat-sft}
Consider a contextual bandit problem and any non-degenerate context $\state$ with full-support $\base(\cdot\mid\state)$ and $V_{\base}(\state) \in (0,1)$. Then there exist scalars $\lam^*_\rw,\lam^*_\unw > 0$, $\marg^*_\rw, \marg^*_\unw \geq 0$ (depending on $\base, \Q$), such that for all $\lam\geq\max\{\lam^*_\rw,\lam^*_\unw\},\, \marg\geq\max\{\marg^*_\rw, \marg^*_\unw\}$, the corresponding policies $\piunw$ and $\pirw$ satisfy the following: \emph{(1)} $\val{\pirw}(\state) \geq \val{\scond}(\state)$; \emph{(2)} $\val{\piunw}(\state) \geq \val{\scond}(\state)$. In particular, both the value function differences $\val{\pirw} - \val{\scond}$ and $\val{\piunw} - \val{\scond}$ can be lower bounded by a constant proportional to $\Var_{\act\sim\scond(\cdot\mid\state)}(\Q(\state, \act))$. The inequality is strict if and only if $\Q(\state,\cdot)$ is non-constant on the actions with $\Q(\state,\act)>0$.
\end{restatable}

\add{\textbf{Case of deterministic rewards.} If $\Q(\state,\act)\in\{0,1\}$ for every action, as for a deterministic verifier of complete responses, and $0<\Vb<1$, then for every $\lam>0$ and every finite $\marg\ge0$ both minimizers equal $\scond(\cdot\mid\state)$ and have value one. They are supported on the positive actions, where $\fcond$ vanishes, so the contrastive terms are constant there and the SFT term in the combined loss is uniquely minimized at $\scond$. The strict improvement of Theorem~\ref{thm:beat-sft} therefore requires unequal success probabilities among the positive-quality actions.}

\vspace{-3pt}
\subsection{$\piunw$ can be worse than both $\base$ and $\scond$}
\vspace{-3pt}
\begin{restatable}{lemma}{lemunwbasepure}
    \label{ex:unw-base}
Consider a contextual bandit instance with two actions $\act_1, \act_2$ and a single state $\state$. Using definitions in Section \ref{sec:setup}, we know that $Q(\state, \act) = \Prob(\reward(\state, \act) = 1 | \bigstate_0=\state, \bigact_0 = \act)$. We assume that $ 1 > Q(\state, \act_1) > Q(\state, \act_2) > 0$. For a base policy $\pi$ with $\pi(\act_1\mid\state) > \pi(\act_2\mid\state) > 0$ and for the purely contrastive case ($\lam$ = 0), if $\marg < \ln\left(\frac{\pi(\act_1\mid\state)}{\pi(\act_2\mid\state)}\right)$, then $\val{\piunw} < \val{\pol}$.
\end{restatable}

 The training data might contain ``misdirectional labels'' due to stochasticity of reward. Comparisons favoring $\act_1$ increase the log-probability gap until it reaches $\marg$ and saturates, while comparisons favoring $\act_2$ continue to push the gap down. The pure contrastive objective is thus minimized when
$\ln\piunw(\act_1\mid\state) - \ln\piunw(\act_2\mid\state)=\marg.$ If the base policy's log-probability gap already exceeds $\marg$, this update actually reduces the probability of choosing the better action. A stronger version with positive $\lam$ is stated and proved in \cref{lem:unw-base} of Appendix~\ref{app:proofs:guarantees}.

In these examples, the base policy $\base$ already assigns higher weight to the higher value actions, with more than the level of separation the margin $\delta$ demands in the contrastive loss component and as a result $\piunw$ fails to improve on $\base$. This motivates the need to define the re-weighted loss $\lrw$ (\cref{eq:loss:rw}) which adapts the margin according to $\base$ in its contrastive term.

\vspace{-3pt}
\subsection{$\pirw$ can be worse than $\scond$}
\vspace{-3pt}
Similar to the previous contextual bandit instances, in Lemma \ref{lem:rw-sft}, we construct an instance where $\pirw$ is worse than the success conditioning policy $\scond$. Proof can be found in Appendix \ref{app:proofs}.

\begin{restatable}{lemma}{lemrwsft}
\label{lem:rw-sft}
Consider a two-action contextual bandit with a single state $\state$ and base policy $\base$ with full support. Suppose we have two actions $\act_1,\act_2$, with $1>\Q(\state,\act_1)>\Q(\state,\act_2)>0$. Then, for any $\lam\geq0$ and any margin
$0\leq\marg<\ln\bigl(\Q(\state,\act_1)/\Q(\state,\act_2)\bigr)$, the optimizer $\pirw$ of the corresponding \reweighted hinge loss satisfies $\val{\pirw}<\val{\scond}$.
\end{restatable}

The above suggests that, if the margin $\marg$ in the contrastive losses $\lrw, \lunw$ is not high enough, the respective optimizers cannot beat $\scond$ in terms of expected reward. 
In contrast, recall from Theorem \ref{thm:beat-sft} that for high enough margin and appropriate $\lam$, both $\pirw$ and $\piunw$ can improve over $\scond$.

\vspace{-3pt}
\subsection{Comparison between $\piunw$ and $\pirw$}
\vspace{-3pt}
Theorems \ref{thm:rw-base}, \ref{thm:beat-sft} and Lemmas \ref{lem:unw-base}, \ref{lem:rw-sft} give rise to the natural question: which of the two losses ($\lunw, \lrw$) should one use? On the one hand, we have seen in Theorem \ref{thm:rw-base} that $\pirw$ is at least always guaranteed to improve over the base policy $\base$ regardless of the hyperparameters $\lambda, \delta$ involved in its definition. On the other hand, we have seen in Theorem \ref{thm:beat-sft} that both the losses $\lunw, \lrw$ (for appropriately chosen $\lambda, \delta$) can beat $\scond$.

In the following result, we identify a situation, which can help in clearly determining between the two choices. 

\begin{definition}[Aligned/ Anti-aligned policies]
\label{def:aligned}
We say that a policy $\pol$ is \emph{\comon} if for every state $\state$ and actions $\act, \act^\prime$, 
\[
(\Q(\state, \act)-\Q(\state, \act^\prime))(\pol(\act\mid \state)-\pol(\act^\prime\mid \state))\ge0.
\]
If for every state $\state$ and all pairs of actions $\act, \act^\prime$, the reverse inequality holds, then we call the policy $\pol$ \emph{\noncomon}.
\end{definition}

Theorem \ref{thm:comono} uses Definition \ref{def:aligned} to select between $\piunw, \pirw$. Proof can be found in Appendix \ref{app:proofs}.

\begin{restatable}{theorem}{thmcomono}\label{thm:comono}
Consider a contextual bandit with non-degenerate states and a base policy $\base$. For any fixed scalars $\lambda>0,\, \delta \geq 0$, let $\piunw$ and $\pirw$ be the corresponding \unweighted and \reweighted policies as defined in Section \ref{sec:setup}. 
If  $\base$ is \comon then $\val{\pirw}(\state)\ge\val{\piunw}(\state)$ for every state $\state$. Conversely, if $\base$ and $\Q$ are \noncomon then  $\val{\pirw}(\state)\le\val{\piunw}(\state)$ for every state $\state$.
\end{restatable}

Therefore, if $\base$ already puts higher weight on higher reward actions, then $\pirw$ with appropriately chosen $\lam, \marg$ is guaranteed to be better than both $\scond$ and  $\piunw$.

\vspace{-3pt}
\subsection{Generalizing our results from Contextual bandits to MDPs}
\vspace{-3pt}
\label{sec:MDPs}
While we were able to show Theorem \ref{thm:rw-base} for general finite-horizon MDPs, Theorems \ref{thm:beat-sft} and $\ref{thm:comono}$  were proved only in the case of contextual bandits. While this already helps us in important practical settings such as fine-tuning language models to maximize rewards, it's not clear to what extent these results might hold for general RL problems involving finite-horizon MDPs.
Below we demonstrate constructions of MDPs where these theorems fail. In Lemma \ref{lem:mdp-sft}, we construct an MDP such that for any values of the hyperparameters $\lambda, \delta\geq 0$, the corresponding contrastive policies $\piunw$ and $\pirw$ are inferior to $\scond$. Details of the construction and the complete proof are presented in Appendix \ref{app:proofs}.   

\begin{restatable}{lemma}{lemmdpsft}
\label{lem:mdp-sft}
There exists a finite-horizon tabular MDP (with a DAG structure) with binary terminal reward, a full-support base policy $\base$, and an
initial distribution $\mu$, such that the following holds simultaneously
for $\pirw$ and $\piunw$. For every finite
$\lam\ge0$, every $\marg\ge0$, every
statewise optimizer $\pimix$ of the contrastive-plus-SFT objective (from \cref{eq:loss:unw,eq:loss:rw}) satisfies $V_{\scond}(\mu)-V_{\pimix}(\mu) > 0$.
\end{restatable}

\begin{remark}
    While the one-step reward $\Q(\state,\act)$ is independent of the deployed policy for contextual bandits, for trajectory rollout in MDPs it is dependent on the policy since $\Q_{\pol}(\state,\act)=\sum_{\state'\in\stateset}\Prob(\state'\mid\state,\act)\val{\pol}(\state')$. Thus, one-step guarantees obtained by minimizing state-wise action-wise losses do not directly give us trajectory-level guarantees in the general case. It should be noted that, as $\lam\to\infty$, both the optimizers $\pirw$ and $\piunw$ tend to $\scond$ anyways, but they need not always approach from ``above'' in the case of MDPs, and the entire trajectory of $\val{\pimix},\,\lambda: 0\to \infty$ can stay below $\val{\scond}$.
\end{remark}

 In Lemma \ref{lem:mdp-comono} below we show that even Theorem~\ref{thm:comono} does not hold for general episodic MDPs. In particular, we show that $\piunw$ can beat $\pirw$ in expected trajectory level value despite the base policy $\base$ being aligned with rewards. Proof is provided in Appendix \ref{app:proofs}. This also critically uses the dependence between the action value function $\Q_\pol$ and the deployed policy $\pol$ that was highlighted in the remark above.

\begin{restatable}{lemma}{lemmdpcomono}
\label{lem:mdp-comono}
There exists a finite-horizon MDP with two decision states,
binary terminal rewards, and a full-support base policy $\base$ (that is strictly aligned with $\Qb$ at both states) and some hyperparameters $\lam,\marg>0$
 such that the state-wise hinge optimizers satisfy $\val{\piunw}(\state_0)>\val{\pirw}(\state_0).$
\end{restatable}

\vspace{-3pt}
\subsection{Optimization of metrics other than value}\label{sec:retrieval}
\vspace{-3pt}
In practical applications, value is not the only objective a policy needs to maximize. For example, in generative retrieval \citep{li2023ltrgr, tay2022transformermemorydifferentiablesearch},
the policy $\pol(\cdot\mid q)$ assigns probabilities to document
identifiers given a text query $q$, and these scores are used to
construct a ranked list of retrieved documents. The policy $\pol$ takes as input a text query $q$ and outputs a list $\widehat{\mathcal{D}}_K$ of $K$ documents from a corpus $\mathcal{D}$ using their likelihood given by $\pol(\cdot\mid q)$ (distribution over the document corpus). Given a binary reward $\reward(q, d)$ (e.g., relevance) distributed as a Bernoulli variable with mean $\Q(q,d)$, the value of $\pol$ is $V_{\pol} = \E_{q\sim \initlaw}\sum_{d\in \mathcal{D}}\Q(q,d)\pi(d\mid q)$. This does not capture the key retrieval metrics that one cares about. For example, one might want to maximize one of the following metrics. Let $\mathcal{D}(q)$ be the set of all documents relevant to query $q$ (say all $d\in\mathcal{D}$ such that $\Q(q,d)>\eta$ for some fixed threshold $\eta$). For each $k\in [K]$, define Recall$@k = \E_{q\sim\initlaw}\left(|\mathcal{D}(q)\cap\widehat{\mathcal{D}}_k|/|\mathcal{D}(q)|\right)$, and Precision$@k = \E_{q\sim\initlaw}\left(|\mathcal{D}(q)\cap\widehat{\mathcal{D}}_k|/k\right)$. Here $\widehat{\mathcal{D}}_k\subset\widehat{\mathcal{D}}_K$ are the top $k$ documents in $\widehat{\mathcal{D}}_K$ as per the likelihood $\pol(\cdot\mid q)$.  We observe that these metrics exhibit two properties: $(a)$ they are a function of the quality sequence $(\Q(\state,\ell_1),\dots,\Q(\state,\ell_N))$ only, where the list $\ell\coloneqq(\ell_1,\cdots,\ell_N)$ is an ordering of the actions and $(b)$ they do not decrease when two entries are swapped so that the one with the larger quality moves up. We call a metric with these two properties \emph{swap-monotone}. Next, we prove that $\piunw$ and $\pirw$ improve such swap monotone metrics in contextual bandit settings (such as our retrieval problem).

\textbf{Inversions and ranking using a policy -} For any contextual bandit problem, fix a state $\state$ and assume the action set $\actset = \{\act_1, \ldots, \act_N\}$ is ordered such that $\Q(\state, \act_1) \geq \Q(\state, \act_2) \geq\cdots\geq\Q(\state, \act_N)$.
Considering the score vector given by likelihood $\pol(\cdot\mid\state)$, the \emph{inversions} are the pairs it gets strictly wrong, namely $\Inv(\pol)=\bigl\{(i,j):\ \Q(\state, \act_i)>\Q(\state, \act_j),\ \pol(\act_i\mid\state)<\pol(\act_j\mid\state)\bigr\}.$ Inversions order rankings without reference to a metric: It is easy to see that if $\Inv(\pol_1)\subseteq\Inv(\pol_2)$, and $\pol_1$ does not tie any two actions of distinct quality then the list of actions sorted by $\pol_1(\cdot\mid\state)$ is at least as good as the list sorted by $\pol_2(\cdot\mid\state)$ for \emph{every} swap-monotone metric at once (we prove this formally in the proof of Theorem~\ref{thm:rankings}). This is because a list whose inversions contain those of another can be turned into it by adjacent swaps, each moving a higher-quality action above a lower-quality one; by $(a)$ and $(b)$ no such swap lowers a swap-monotone metric. And thus at the state $\state$, $\pol_1$ has a higher swap monotone score than $\pol_2$. The list sorted by relevance has no inversions and is therefore optimal for all these metrics simultaneously, which is the probability ranking principle of \citet{robertson1977prp}. Next, we show that $\piunw$ is always guaranteed to improve over $\scond$ and $\pirw$ is always guaranteed to improve over $\base$ in terms of minimizing inversions.

\begin{restatable}{theorem}{thmrankings}\label{thm:rankings}
Consider a contextual bandit with a non-degenerate state $\state$ and actions $\act_1,\cdots,\act_n$ such that $\Q(\state, \act_1) \geq \cdots \geq \Q(\state, \act_n)$. Then the following holds:
\begin{enumerate}
\item Fix any $\lam>0$ and let $\piunw$ and $\pirw$ be the \unweighted and \reweighted contrastive policies respectively. Then, $\Inv(\piunw)\subseteq\Inv(\scond)\subseteq\Inv(\base)$ and $\Inv(\pirw)\subseteq\Inv(\base)$.

\item $\exists\lam^*>0,\marg^*>0$ such that $\forall \lam \geq \lam^*$ and $\marg \geq \marg^*$, we have $\Inv(\pirw) \subseteq\Inv(\scond)$.
\end{enumerate}
\end{restatable}

\subsection{Connections to preference optimization and traditional RL}
So far we have explored how hinge based contrastive losses with an SFT anchor term compare to just training with SFT. However, in the RLHF literature of preference optimization, DPO style contrastive losses like \citep{dpo2023,simpo,dpop} already explore reward maximization using contrastive learning. We make the observation that the softplus function $x\mapsto\tau\ln\left(1+e^{x/\tau}\right)$ converges to the hinge map $x\mapsto[x]_+$ as $\tau\downarrow0$. We thus naturally ask the question if our results for hinge-based contrastive loss extend to the softplus based contrastive losses as well, and if that can be connected to DPO style contrastive training. We discuss more on this in Appendix~\ref{app:softplus}. 

One more natural question is to ask the connection of contrastive learning which provably improves value, with traditional RL methods, which are designed to maximize value. We already know from \cite{russo2026success} that the optimizer of the SFT loss can be represented as a TRPO. We show that one can say the same about the contrastive losses too for specific regimes of $\lam, \marg$ which we discuss in Appendix~\ref{app:proofs:trpo}

\input{sections/experiments.tex}

\newpage
\subsection*{AI use statement}
We used generative AI tools (specifically, LLMs like ChatGPT and Claude) extensively for editing the text, for assistance in developing and checking mathematical arguments and proofs, and for assistance with experimental code. The tools are not part of the objectives or models analyzed.  The authors have reviewed all AI-assisted material and take responsibility for the mathematical and empirical claims as well as soundness of all code and experimental results.

\bibliography{references/references}
\bibliographystyle{references/author_year}
\newpage
\input{sections/proofs.tex}

\end{document}

%% file: math_commands.tex
\usepackage{amsmath,amssymb,amsthm,mathtools}
\usepackage{thmtools,thm-restate}
\usepackage{bm}
\usepackage{xspace}
\usepackage{relsize}

\declaretheorem[name=Theorem,numberwithin=section]{theorem}

\declaretheorem[name=Lemma,sibling=theorem]{lemma}

\declaretheorem[name=Definition,sibling=theorem,style=definition]{definition}

\declaretheorem[name=Remark,sibling=theorem,style=remark]{remark}

\newcommand{\actset}{\mathcal A}          
\newcommand{\stateset}{\mathcal S}        
\newcommand{\termset}{\mathcal T}         
\newcommand{\vocab}{\mathcal V}           
\newcommand{\ctx}{x}
\newcommand{\bigctx}{X}
\newcommand{\act}{a}
\newcommand{\state}{s}
\newcommand{\resp}{y} 
\newcommand{\bigresp}{Y} 
\newcommand{\horizon}{H}
\newcommand{\initlaw}{\mu}                
\newcommand{\tkern}{P}                    
\newcommand{\reward}{\mathcal{R}}

\newcommand{\pol}{\pi}
\newcommand{\base}{\pi_{0}}               
\newcommand{\scond}{\base^{+}}               
\newcommand{\fcond}{\base^{-}}                
\newcommand{\rw}{\mathrm{rw}}
\newcommand{\unw}{\mathrm{unw}}
\newcommand{\pirw}{{\pol_{\rw}}}            
\newcommand{\piunw}{{\pol_{\unw}}}          
\newcommand{\pimix}{\pol_{\lam}}          

\newcommand{\Q}{Q}                        
\newcommand{\Qb}{Q_{0}}                   

\newcommand{\Vb}{V_{0}}                   
\newcommand{\val}[1]{V_{#1}}              

\newcommand{\occ}{d}     
\newcommand{\traj}{\tau}
\newcommand{\bigstate}{S}
\newcommand{\bigact}{A}

\newcommand{\newpol}{\hat\pi}

\newcommand{\lam}{\lambda}                
\newcommand{\marg}{\delta}                
\newcommand{\temp}{\tau}                  
\newcommand{\score}{u}
\newcommand{\hinge}[1]{[#1]_{+}}
\newcommand{\sig}{\sigma}                 

\newcommand{\unweighted}{\text{un-weighted}\xspace}
\DeclareRobustCommand{\reweighted}{\text{re-weighted}\xspace}
\newcommand{\lunw}{\mathcal{L}_{\unw}}
\newcommand{\lrw}{\mathcal{L}_{\rw}}
\newcommand{\comon}{\text{aligned}\xspace}
\newcommand{\noncomon}{\text{anti-aligned}\xspace}

\newcommand{\Inv}{\operatorname{Inv}}

\newcommand{\E}{\mathbb E}
\newcommand{\Prob}{\mathbb P}

\newcommand{\Var}{\operatorname{Var}}
\newcommand{\Cov}{\operatorname{Cov}}
\newcommand{\KL}{\operatorname{KL}}

\newcommand{\osc}{\operatorname{osc}}
\newcommand{\ind}{\mathbf 1}

\DeclareMathOperator*{\argmin}{arg\,min}

\newcommand{\inner}[2]{\langle #1,#2\rangle}
\newcommand{\norm}[1]{\lVert #1\rVert}
\newcommand{\abs}[1]{\lvert #1\rvert}

\newcommand{\lctr}{\mathcal L_{\mathrm{ctr}}}
\newcommand{\lsft}{\mathcal L_{\mathrm{SFT}}}
\newcommand{\Ent}{\operatorname{Ent}}

%% file: sections/experiments.tex
\vspace{-7.5pt}
\section{Experiments: applications in Generative Retrieval}
\vspace{-7.5pt}
\label{sec:expt}
In this section, we empirically compare the policies $\piunw, \pirw, \base^+, \base$ in the framework of generative retrieval, where training must improve the passages returned by the model and their ranking. We organize the evaluation around three
questions: \emph{(1)} Do $\piunw, \pirw$ improve over either $\base$ or $\base^+$? \emph{(2)} How does the comparison between $\piunw$ and $\pirw$ change as the base policy $\base$ changes? \emph{(3)} Do the policies $\piunw, \pirw$ lead to better
retrieval performance? These are asking the question of how well our theoretical guarantees from \cref{thm:beat-sft,thm:comono,thm:rankings} hold in a real-world setting, like generative retrieval. We also validate our theory in a synthetic setup, where we can directly control the \emph{alignment} between $\base$ and $\Q$, which is discussed in Appendix~\ref{subsec:expt:semisynthetic}.

\textbf{Retrieval setup.}
We use a compact-trie implementation of MINDER~\citep{minder2023}, with a
BART-large generator~\citep{bart2020} that produces pseudo-query identifiers
for the 8.84 million passages in MS MARCO~\citep{msmarco}.
We start from three checkpoints of the \emph{ground-truth relevance-supervised training trajectory}:
\textsc{Early} at step 1,000, \textsc{Midpoint} at step 3,000, and
\textsc{Final}. Appendix~\ref{app:expt} gives the complete details. The frozen Qwen3-Reranker-0.6B~\citep{qwen3embedding2025} assigns a soft reward to passages; the identifier reward $\overline \Q(q, z)$ is the mean reward of
its associated passages. Let $\base(z\mid q)$ denote the base model's
full-sequence probability normalized over the selected pool. We draw positive and negative identifiers independently from $\scond(\cdot\mid q)$ and $\fcond(\cdot\mid q)$, respectively.

\textbf{Objectives and methodology.}
We fine-tune the BART model using the contrastive losses $\lrw$~\cref{eq:loss:rw} and $\lunw$~\cref{eq:loss:unw} and the base policy's probability of selecting identifier $z$ for query $q$. We use the convex combination loss, with weight $\omega\in[0,1)$ ($\lam = \omega/(1-\omega)$ for the corresponding SFT weight $\lam$ from \cref{eq:loss:rw,eq:loss:unw}) to avoid numerical issues. We see that \emph{Early}, \emph{Midpoint}, and \emph{Final} have increasing expected rewards, since the model gets more aligned with the reward during training. \add{The choice of the three checkpoints from the same training trajectory is deliberate since this is a real-world analogue of $\base$ getting more ``aligned'' with the rewards and we ask the question whether the guarantees of Theorem~\ref{thm:comono} would hold under a more realistic setting like this. An ambitious goal with this is to explore if different loss functions can be used at different stages of a post-training trajectory, as the alignment changes.} Using each one of the 3 checkpoints as $\base$, we fine-tune the BART model using both the contrastive losses $\lunw, \lrw$ at $\omega\in\{0,1/9,\ldots,8/9\}$ and margins $\delta\in\{0.1,2.575,5.05,7.525,10\}$, and train one shared Reward-SFT endpoint ($\omega=1$). \emph{Reward-SFT is trained using samples from $\pi^+_0$, not the relevance labels.}

\begin{figure}[h]
\vspace{-10pt}
    \centering
    \includegraphics[width=0.80\linewidth]{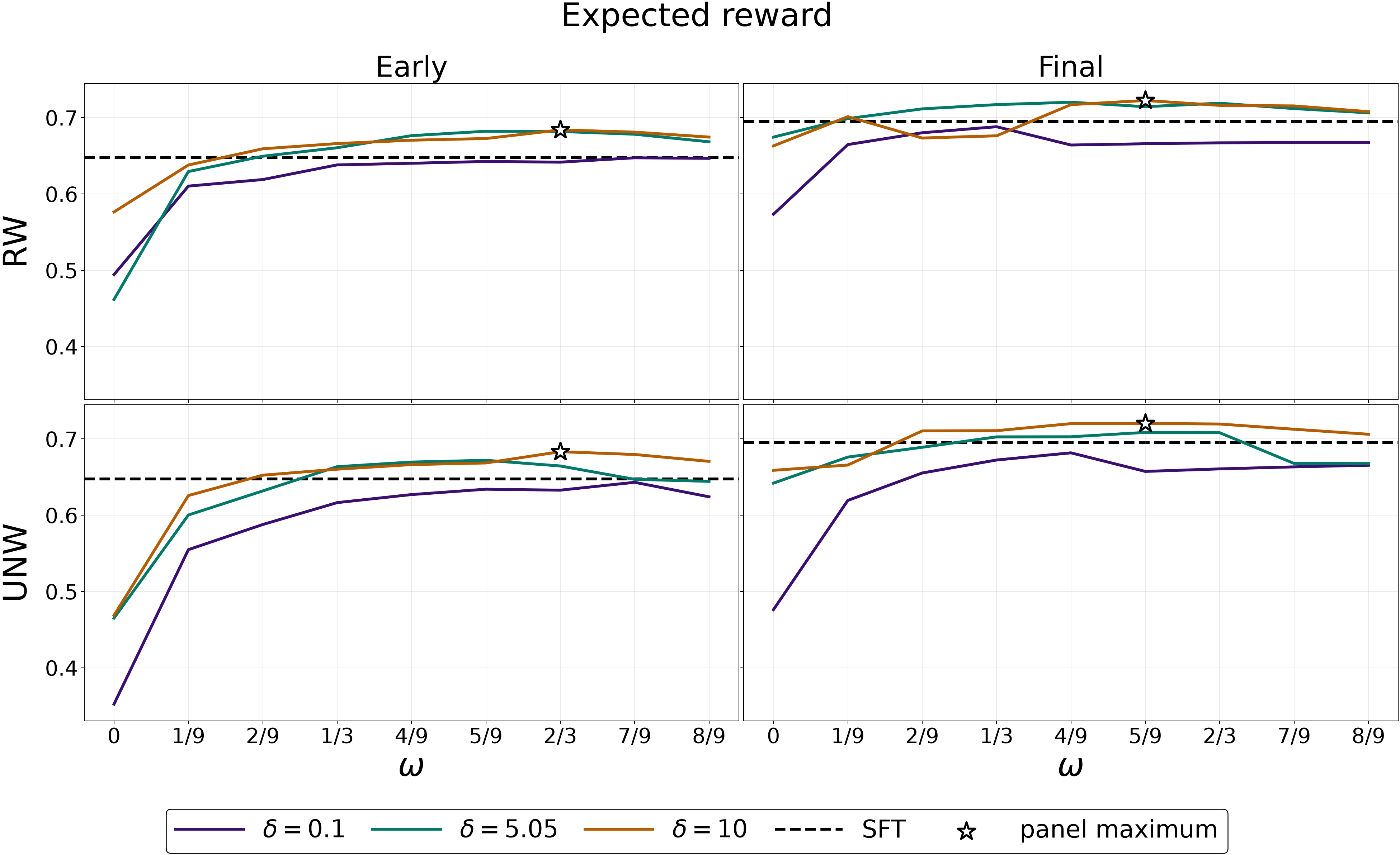}
    \caption{\add{$\val{\pirw}$ and $\val{\piunw}$ vs $\omega$ across different initial $\base$, 3-seed avg. Each curve is one margin $\delta$. Dashed line is Reward SFT}}
    \label{fig:reward_plot_minder_main}
    \vspace{-15pt}
\end{figure}
\paragraph{Expected reward of contrastive and SFT losses}\add{We first show how the expected Qwen rewards $\val{\pirw}, \val{\piunw}$, obtained by beam-free sampling with replacement, vary across this grid for the Early and Final checkpoints and compare to the reward of $\scond$ in \cref{fig:reward_plot_minder_main}}.
We make the following observations: \textbf{(1)} Both $\val{\pirw}$ and $\val{\piunw}$ go over $\val{\pi_{\mathrm{SFT}}}$ as $\omega\to 1$ (i.e., $\lambda \to \infty$), indicating improvement at higher values of the SFT weight. \textbf{(2)} High-margin curves have higher reward than low-margin ones. These two observations validate \cref{thm:beat-sft} in a real-world empirical setup where we observe that high-margin contrastive losses show significant gains in reward over SFT.

\begin{figure}[h]
\vspace{-10pt}
    \centering
    \includegraphics[width=0.80\linewidth]{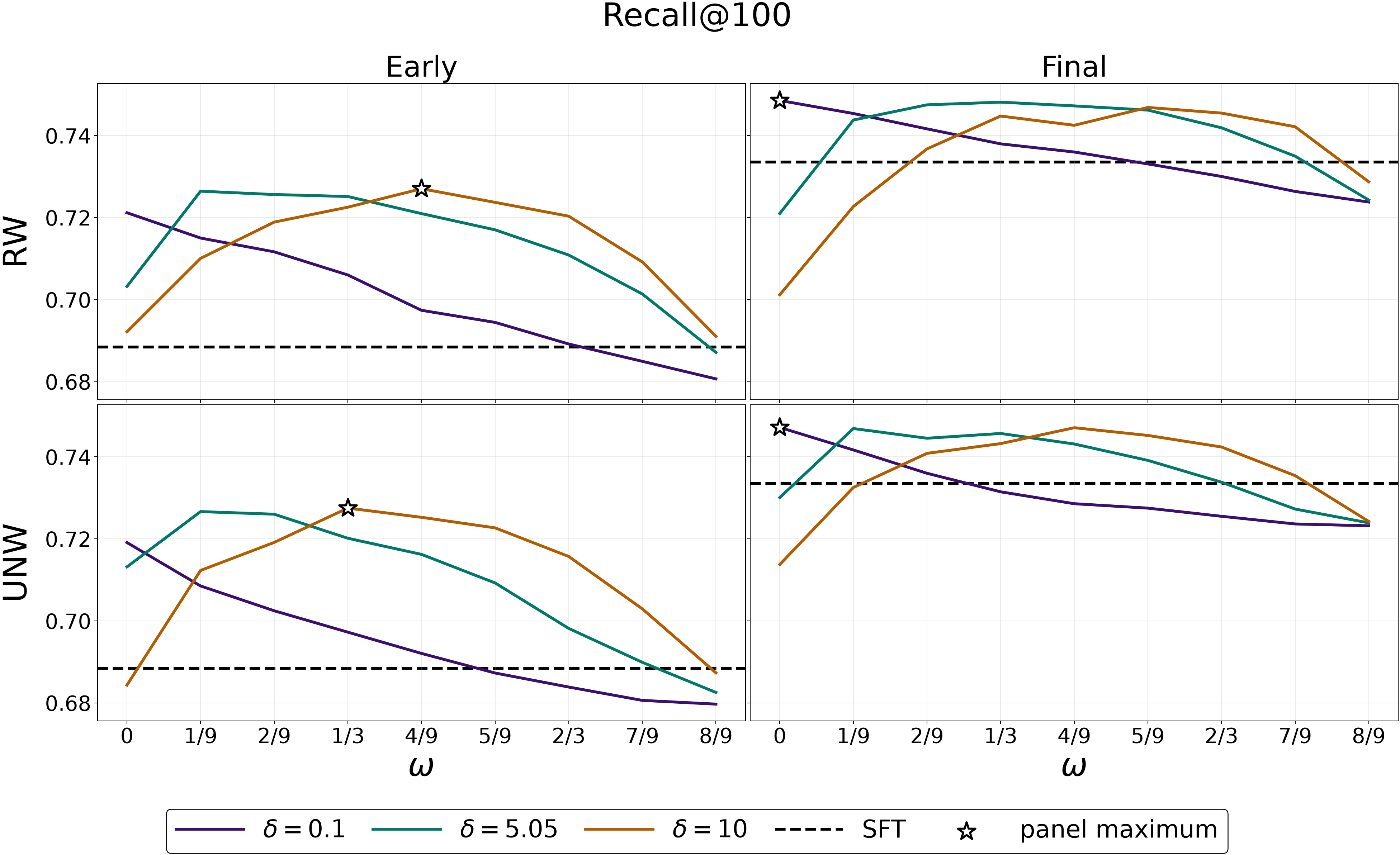}
    \caption{\add{Recall@100 vs $\omega$ across different initial $\base$, 3-seed avg. Each curve is one margin $\delta$. Dashed line is Reward SFT.}}
    \label{fig:recall_plot_minder_main}
\end{figure}

\textbf{Retrieval comparison of contrastive and SFT losses -} Here we wish to see if the gains of contrastive losses over SFT hold for retrieval metrics as well. We give the Recall@100 vs $\omega$ plot in \cref{fig:recall_plot_minder_main}. We observe that: \textbf{(1)} For $2/4$ loss--checkpoint combinations, the best performance was found for $\omega\in(0,1)$ indicating the usefulness in combining the two losses, especially when the policy and reward are not fully aligned (like in the early checkpoint) \textbf{(2)} Low-margin curves start high and degrade fast; higher-margin curves are bell-shaped. This is significant, because this implies that to achieve an optimal trade-off between reward and ranking, a higher margin is preferable, since the low-margin, low-$\omega$ points are higher in Recall but have low reward, whereas high-margin curves with appropriate $\lam$ achieve both high reward and ranking accuracy---best performance overall.

\vspace{-5pt}
\vspace{-5pt}
\begin{table}[H]
  \centering
  \small
  \setlength{\tabcolsep}{4pt}
  \renewcommand{\arraystretch}{1.08}
  \caption{\add{Avg beam-slot reward and expected (sampled) reward (mean $\pm$ SEM across 3 seeds)}.}
  \resizebox{\textwidth}{!}{
   \add{\begin{tabular}{lrrrrrr}
    \toprule
    & \multicolumn{3}{c}{Avg.\ beam reward}
    & \multicolumn{3}{c}{Expected reward} \\
    \cmidrule(lr){2-4}
    \cmidrule(lr){5-7}
    Method
      & Early & Midpoint & Final
      & Early & Midpoint & Final \\
    \midrule
    Base policy ($\pi_0$)
      & 0.3940 $\pm$ 0.0011 & 0.4376 $\pm$ 0.0009 & 0.4743 $\pm$ 0.0012
      & 0.5296 $\pm$ 0.0010 & 0.5761 $\pm$ 0.0008 & 0.6308 $\pm$ 0.0011 \\
    Reward-SFT
      & 0.3862 $\pm$ 0.0009 & 0.4171 $\pm$ 0.0010 & 0.4539 $\pm$ 0.0013
      & 0.6473 $\pm$ 0.0008 & 0.6701 $\pm$ 0.0011 & 0.6951 $\pm$ 0.0009 \\
    \addlinespace[3pt]
    DPO
      & 0.4611 $\pm$ 0.0012 & 0.4740 $\pm$ 0.0007 & \underline{0.5109} $\pm$ 0.0010
      & 0.4902 $\pm$ 0.0011 & 0.6031 $\pm$ 0.0009 & 0.6510 $\pm$ 0.0012 \\
    RankNet
      & 0.4520 $\pm$ 0.0008 & 0.4718 $\pm$ 0.0011 & 0.5096 $\pm$ 0.0009
      & 0.4033 $\pm$ 0.0013 & 0.4482 $\pm$ 0.0010 & 0.5148 $\pm$ 0.0008 \\
    SimPO
      & 0.4601 $\pm$ 0.0010 & 0.4744 $\pm$ 0.0010 & 0.5099 $\pm$ 0.0008
      & 0.3575 $\pm$ 0.0009 & 0.5554 $\pm$ 0.0011 & 0.6198 $\pm$ 0.0010 \\
    DPO+SFT
      & 0.4605 $\pm$ 0.0011 & 0.4742 $\pm$ 0.0009 & 0.5102 $\pm$ 0.0010
      & 0.5265 $\pm$ 0.0012 & 0.6145 $\pm$ 0.0007 & 0.6566 $\pm$ 0.0011 \\
      \midrule
    $\pi_{\mathrm{rw}}$
      & \textbf{0.4703} $\pm$ 0.0009 &\textbf{0.4861} $\pm$ 0.0011 & \textbf{0.5148} $\pm$ 0.0008
      & \textbf{0.6844} $\pm$ 0.0010 & \textbf{0.7060} $\pm$ 0.0012 & \textbf{0.7225} $\pm$ 0.0010 \\
    $\pi_{\mathrm{unw}}$
      & \underline{0.4650 $\pm$ 0.0011} & \underline{0.4763} $\pm$ 0.0008 & 0.5068 $\pm$ 0.0011
      & \underline{0.6832} $\pm$ 0.0009 & \underline{0.7019} $\pm$ 0.0009 & \underline{0.7204} $\pm$ 0.0012 \\
    \bottomrule
  \end{tabular}}}
  \label{tab:minder-main-reward}
\end{table}
\vspace{-5pt}
\textbf{Comparison between $\pirw$ and $\piunw$ as $\base$ gets better -} \add{We report the average beam-slot reward (a more practically relevant metric in terms of generative retrieval) as well as the average sampled reward (a better estimate of $\val{\pol}$) for baselines and the grid-winner (best across all $\lam, \marg$ in validation) of the two hinge-based contrastive losses across checkpoints in \cref{tab:minder-main-reward}. We observe that: \textbf{(1)} $\pirw$ shows better performance than $\piunw$ in both beam average and Monte Carlo rewards, with \emph{$\val{\pirw} - \val{\piunw}$ being higher in the final checkpoint than the initial across both rewards}. \textbf{(2)} Reward-SFT also improves with $\base$. Note that Reward-SFT \emph{does not improve over $\base$ in beam reward, but significantly improves in Monte Carlo expected reward.} \textbf{(3)} Across checkpoints and metrics, the overall winner lies in the two contrastive classes.}
\vspace{-5pt}
\section{Conclusion}
\vspace{-5pt}
We establish when losses combining contrastive training with success conditioning improve on success
conditioning and how this depends on the hyperparameters
and base policy. Our results distinguish reward improvement
from ranking improvement and identify which guarantees extend
from bandits to MDPs. Experiments in generative retrieval
illustrate these distinctions, showing that learning from
failures must be matched to the evaluation objective.

%% file: sections/proofs.tex
\appendix

\section{Limitations}
\label{app:limitations}

\paragraph{Tabular settings and practical learning}
Our main guarantees concern exact state-wise population minimizers over
unrestricted policies. They identify when the \reweighted{} and
\unweighted{} objectives can yield higher value than the base or
positive-only policy, and how this depends on reward variance and
reference--reward alignment. They do not characterize the sample
complexity or convergence of finite-data gradient-based training, nor
bound the approximation error of an arbitrary shared policy class.
Consequently, our results do not generalize to every empirical optimizer or training
iterate.

\paragraph{Shared parametrizations}
 With shared parameters, changes at one state affect others,
and the unrestricted optimum may not be representable. Existing
policy-gradient analyses illustrate how guarantees for parametric
policies depend on assumptions about approximation, sampling, or
overparameterization~\cite{agarwal2021policygradient,wang2019neuralpg};
they also make very specific assumptions on the parametrization and do not work with general arbitrary classes of parametrized policies.
Extending our comparative guarantees to finite feedback while jointly
controlling estimation, approximation, and optimization error remains
an important direction for future work.

\paragraph{Deterministic verification}
Under deterministic binary verification, exact success conditioning
already attains reward one whenever success has positive probability.
Strict improvement over that ideal policy is therefore impossible.
Such cases involve reasoning tasks like math, rather than human-feedback based reward or preference data involving stochasticity which our analysis is focused on. Our stochastic-reward analysis does not establish that
separate practical claim.

\section{Proofs of technical results}
\label{app:proofs}
Throughout the section, we have used $\ctx$ and $\state$ interchangeably to denote a state/context. Generally, $\act$ or $\act_i$ denotes actions.

\subsection{Proofs of Preliminary Lemmas}
\label{app:prelim}

\lemsucfail*
\begin{proof}
Consider a trajectory started from $\bigstate_0=\ctx$ and
subsequently generated by $\pol$.
By the definition of the policy and the action-value function, $\Prob_{\pol}\bigl(
    \bigact_0=\act,\reward(\traj)=1
    \mid \bigstate_0=\ctx
\bigr)=\pol(\act\mid\ctx)\Q_{\pol}(\ctx,\act).$
Similarly, since the terminal reward is binary,
$\Prob_{\pol}\bigl(
    \bigact_0=\act,\reward(\traj)=0
    \mid \bigstate_0=\ctx
\bigr)
=
\pol(\act\mid\ctx)
\bigl(1-\Q_{\pol}(\ctx,\act)\bigr).
$
The corresponding probabilities of the conditioning events are
\begin{align*}
\Prob_{\pol}\bigl(
    \reward(\traj)=1\mid\bigstate_0=\ctx
\bigr)
=\val{\pol}(\ctx),
\,
\Prob_{\pol}\bigl(
    \reward(\traj)=0\mid\bigstate_0=\ctx
\bigr)
=1-\val{\pol}(\ctx).
\end{align*}
Both denominators are positive by the assumed non-degeneracy
of $\ctx$. Bayes' rule therefore gives
\begin{align*}
\pol^+(\act\mid\ctx)
&=
\Prob_{\pol}\bigl(
    \bigact_0=\act
    \mid \bigstate_0=\ctx,\reward(\traj)=1
\bigr)
=
\frac{\pol(\act\mid\ctx)\Q_{\pol}(\ctx,\act)}
     {\val{\pol}(\ctx)},\\
\pol^-(\act\mid\ctx)
&=
\Prob_{\pol}\bigl(
    \bigact_0=\act
    \mid \bigstate_0=\ctx,\reward(\traj)=0
\bigr)
=
\frac{\pol(\act\mid\ctx)
      \bigl(1-\Q_{\pol}(\ctx,\act)\bigr)}
     {1-\val{\pol}(\ctx)}.
\end{align*}
By the Markov property, the same conditional action laws apply
at every visit to $\ctx$.
\end{proof}

Now, for the next lemma, for notational convenience we pad every trajectory to length $\horizon$: terminal states are absorbing, the only action available at a terminal state is a dummy action $\bot$, which every policy takes with probability one, and the outcome is unchanged. A trajectory is thus $\traj=(\bigstate_0,\bigact_0,\ldots,\bigstate_{\horizon-1},\bigact_{\horizon-1},\bigstate_{\horizon})$, with $\bigstate_t\in\termset$ and $\bigact_t=\bot$ once the episode has terminated. We call a step $(\bigstate_t,\bigact_t)$ with $\bigstate_t\notin\termset$ a \emph{decision step}. Since $\hat\pol(\bot\mid\state)=1$ at every terminal state $\state$ for every policy $\hat\pol$, the padded steps contribute $\ln\hat\pol(\bigact_t\mid\bigstate_t)=0$ to every log-likelihood sum below, and policies are compared and optimized at decision states only. The occupancy of $\pol$ from $\initlaw$ is $\occ_{\pol,\initlaw}(\state)\coloneqq\E_{\pol,\initlaw}\sum_{t<\horizon}\ind\{\bigstate_t=\state\}$ for a decision state $\state$, the expected number of visits to $\state$; since a trajectory has at most $\horizon$ decision steps, $\sum_{\state}\occ_{\pol,\initlaw}(\state)\le\horizon$.

\lemsucfailnext*
\begin{proof}
We treat the first problem; the second is identical with $\reward(\traj)=0$, $1-\Q_\pol$ and $1-\val{\pol}$ in place of $\reward(\traj)=1$, $\Q_\pol$ and $\val{\pol}$. The conditioning event has probability $\Prob_\pol(\reward(\traj)=1)=\val{\pol}(\initlaw)>0$, and every summand $-\ln\hat\pol(\bigact_t\mid\bigstate_t)$ is nonnegative, with the conventions $-\ln0=+\infty$ and $0\cdot(+\infty)=0$; hence the objective is well defined in $[0,+\infty]$, and expectations of sums of nonnegative terms may be taken term by term.

For a decision state $\ctx$ and $t<\horizon$, let $E^{t}_\ctx=\{t<\horizon,\ \bigstate_t=\ctx\}$ be the event that $\ctx$ is visited at time $t$, and suppose that $\Prob_\pol(E^{t}_\ctx)>0$. By the Markov property, conditionally on $E^{t}_\ctx$ the continuation $(\bigact_t,\bigstate_{t+1},\dots,\bigstate_{\horizon})$ together with the outcome $\reward(\traj)$ is distributed as a trajectory started from $\bigstate_0=\ctx$ and generated by $\pol$, together with its outcome. Hence, by the definition of the policy and of the action-value function, exactly as in the proof of Lemma~\ref{lem:sucfail},
\begin{equation}\label{eq:visit-law}
\Prob_\pol\bigl(\bigact_t=\act,\ \reward(\traj)=1\bigm|E^{t}_\ctx\bigr)=\pol(\act\mid\ctx)\,\Q_\pol(\ctx,\act),
\,
\Prob_\pol\bigl(\reward(\traj)=1\bigm|E^{t}_\ctx\bigr)=\val{\pol}(\ctx),
\end{equation}
the second identity by summing the first over $\act$.

Since a trajectory occupies exactly one state at each decision step,
\begin{equation*}
-\sum_{t<\horizon}\ln\hat\pol(\bigact_t\mid\bigstate_t)=\sum_{\ctx}\sum_{t<\horizon}\ind\{E^{t}_\ctx\}\bigl(-\ln\hat\pol(\bigact_t\mid\ctx)\bigr),
\end{equation*}
where the outer sum runs over the decision states and the $(\ctx,t)$ summand is $0$ off $E^{t}_\ctx$. Fix $\ctx$ and $t$. If $\Prob_\pol(E^{t}_\ctx)=0$, the $(\ctx,t)$ summand vanishes almost surely. Otherwise, splitting $E^{t}_\ctx$ according to the value of $\bigact_t$ and using \cref{eq:visit-law},
\begin{align*}
\E_\pol\bigl[\ind\{E^{t}_\ctx\}\bigl(-\ln\hat\pol(\bigact_t\mid\ctx)\bigr)\bigm|\reward(\traj)=1\bigr]
&=\sum_{\act}\frac{\Prob_\pol\bigl(E^{t}_\ctx,\ \bigact_t=\act,\ \reward(\traj)=1\bigr)}{\val{\pol}(\initlaw)}\bigl(-\ln\hat\pol(\act\mid\ctx)\bigr)\\
&=\frac{\Prob_\pol(E^{t}_\ctx)}{\val{\pol}(\initlaw)}\sum_{\act}\pol(\act\mid\ctx)\,\Q_\pol(\ctx,\act)\bigl(-\ln\hat\pol(\act\mid\ctx)\bigr).
\end{align*}
Summing over $t$ and $\ctx$, and writing $\occ_{\pol}(\ctx)=\E_\pol\sum_{t<\horizon}\ind\{\bigstate_t=\ctx\}=\sum_{t<\horizon}\Prob_\pol(E^{t}_\ctx)$ for the expected number of visits to $\ctx$, we obtain
\begin{equation}\label{eq:sft-decomp}
\begin{split}
-\E_{\traj\sim\pol}\Bigl[\sum_{t<\horizon}\ln\hat\pol(\bigact_t\mid\bigstate_t)\Bigm|\reward(\traj)=1\Bigr]
&=\sum_{\ctx}\frac{\occ_{\pol}(\ctx)}{\val{\pol}(\initlaw)}\sum_{\act}\pol(\act\mid\ctx)\,\Q_\pol(\ctx,\act)\bigl(-\ln\hat\pol(\act\mid\ctx)\bigr)\\
&=\sum_{\ctx:\,\occ^{+}(\ctx)>0}\occ^{+}(\ctx)\sum_{\act}\pol^{+}(\act\mid\ctx)\bigl(-\ln\hat\pol(\act\mid\ctx)\bigr),
\end{split}
\end{equation}
where $\occ^{+}(\ctx)=\occ_{\pol}(\ctx)\,\val{\pol}(\ctx)/\val{\pol}(\initlaw)$. Indeed, if $\val{\pol}(\ctx)=0$ then $\pol(\act\mid\ctx)\Q_\pol(\ctx,\act)=0$ for every $\act$, so the $\ctx$ term vanishes; otherwise $\pol(\act\mid\ctx)\Q_\pol(\ctx,\act)=\val{\pol}(\ctx)\,\pol^{+}(\act\mid\ctx)$, where $\pol^{+}(\cdot\mid\ctx)$ is given by the first formula in \cref{eq:cond}, which only requires $\val{\pol}(\ctx)>0$ (by \cref{eq:visit-law} and Bayes' rule, it is the conditional law of $\bigact_t$ given $E^{t}_\ctx$ and $\reward(\traj)=1$ for every $t$ with $\Prob_\pol(E^{t}_\ctx)>0$). By \cref{eq:visit-law}, $\occ^{+}(\ctx)=\E_\pol[\sum_{t<\horizon}\ind\{\bigstate_t=\ctx\}\mid\reward(\traj)=1]$ is the expected number of visits to $\ctx$ on successful trajectories; at the states with $\occ^{+}(\ctx)=0$ the objective does not depend on $\hat\pol(\cdot\mid\ctx)$.

For each $\ctx$ with $\occ^{+}(\ctx)>0$, the inner sum in \cref{eq:sft-decomp} is the cross-entropy
\begin{equation*}
\sum_{\act}\pol^{+}(\act\mid\ctx)\bigl(-\ln\hat\pol(\act\mid\ctx)\bigr)=\Ent\bigl(\pol^{+}(\cdot\mid\ctx)\bigr)+\KL\bigl(\pol^{+}(\cdot\mid\ctx)\,\big\|\,\hat\pol(\cdot\mid\ctx)\bigr),
\end{equation*}
where $\Ent$ is the Shannon entropy and the divergence equals $+\infty$ when $\hat\pol(\cdot\mid\ctx)$ vanishes at an action charged by $\pol^{+}(\cdot\mid\ctx)$. The objective is therefore
\begin{equation*}
\sum_{\ctx}\occ^{+}(\ctx)\,\Ent\bigl(\pol^{+}(\cdot\mid\ctx)\bigr)+\sum_{\ctx}\occ^{+}(\ctx)\,\KL\bigl(\pol^{+}(\cdot\mid\ctx)\,\big\|\,\hat\pol(\cdot\mid\ctx)\bigr).
\end{equation*}
The first sum does not depend on $\hat\pol$ and is finite, since each entropy is at most $\ln|\actset|$ and $\sum_\ctx\occ^{+}(\ctx)=\E_\pol\bigl[\sum_{t<\horizon}\ind\{\bigstate_t\notin\termset\}\bigm|\reward(\traj)=1\bigr]\le\horizon$. By Gibbs' inequality the second is nonnegative and vanishes at $\hat\pol=\pol^{+}$. Hence $\pol^{+}$ is a minimizer, which is \cref{eq:condsft}. (Gibbs' inequality also shows that every minimizer agrees with $\pol^{+}$ at every state with $\occ^{+}(\ctx)>0$.)
\end{proof}

\begin{restatable}[State-wise form of the objectives]{lemma}{lemstatewise}\label{lem:statewise}
Let $\occ^{+}(\state)\coloneqq\E_\pol\bigl[\sum_{t<\horizon}\ind\{\bigstate_t=\state\}\bigm|\reward(\traj)=1\bigr]=\occ_{\pol,\initlaw}(\state)\val{\pol}(\state)/\val{\pol}(\initlaw)$ be the expected number of visits to $\state$ on a successful rollout. For every policy $\newpol$,
\begin{align}
\label{eq:loss:unw:app}
\lunw(\newpol,\pol)&=\sum_{\state}\occ^{+}(\state)\left(\mathop{\E}\limits_{\substack{\act_+\sim\pol^{+}(\cdot\mid\state)\\\act_-\sim\pol^{-}(\cdot\mid\state)}}\hinge{\marg+\ln\newpol(\act_-\mid\state)-\ln\newpol(\act_+\mid\state)}-\lam\mathop{\E}\limits_{\act_+\sim\pol^{+}(\cdot\mid\state)}\ln\newpol(\act_+\mid\state)\right),\\
\label{eq:loss:rw:app}
\lrw(\newpol,\pol)&=\sum_{\state}\occ^{+}(\state)\left(\mathop{\E}\limits_{\substack{\act_+\sim\pol^{+}(\cdot\mid\state)\\\act_-\sim\pol^{-}(\cdot\mid\state)}}\hinge{\marg+\ln\frac{\newpol(\act_-\mid\state)}{\pol(\act_-\mid\state)}-\ln\frac{\newpol(\act_+\mid\state)}{\pol(\act_+\mid\state)}}-\lam\mathop{\E}\limits_{\act_+\sim\pol^{+}(\cdot\mid\state)}\ln\newpol(\act_+\mid\state)\right)
\end{align}
where $\act_+$ and $\act_-$ are independent and the sums run over the decision states; at every state with $\occ^{+}(\state)>0$ and $\val{\pol}(\state)<1$ one has $\val{\pol}(\state)\in(0,1)$, so that $\pol^{\pm}(\cdot\mid\state)$ exist. If instead $\traj_-$ is drawn independently of $\traj_+$ and a pair is formed by every two steps at which the two rollouts are in the same state, then the contrastive term carries the weight $\occ^{+}(\state)\occ^{-}(\state)$, with $\occ^{-}(\state)\coloneqq\occ_{\pol,\initlaw}(\state)(1-\val{\pol}(\state))/(1-\val{\pol}(\initlaw))$, while the likelihood term keeps $\occ^{+}(\state)$.
\end{restatable}

\begin{proof}
For a function $f$ of a state and an action, the computation leading to \cref{eq:sft-decomp}, with $-\ln\hat\pol(\act\mid\ctx)$ replaced by $f(\ctx,\act)$, gives
\begin{equation}\label{eq:step-decomp}
\E_\pol\Bigl[\sum_{t<\horizon}f(\bigstate_t,\bigact_t)\Bigm|\reward(\traj)=1\Bigr]=\sum_{\ctx}\occ^{+}(\ctx)\sum_{\act}\pol^{+}(\act\mid\ctx)\,f(\ctx,\act):
\end{equation}
by \cref{eq:visit-law}, $\Prob_\pol(\bigact_t=\act,\reward(\traj)=1\mid E^{t}_\ctx)=\pol(\act\mid\ctx)\Q_\pol(\ctx,\act)=\val{\pol}(\ctx)\,\pol^{+}(\act\mid\ctx)$ at every visit, so summing over $t$ and dividing by $\Prob_\pol(\reward(\traj)=1)=\val{\pol}(\initlaw)$ produces the weight $\sum_t\Prob_\pol(E^{t}_\ctx)\val{\pol}(\ctx)/\val{\pol}(\initlaw)=\occ^{+}(\ctx)$. The likelihood term of \cref{eq:loss:unw:traj} is \cref{eq:step-decomp} with $f(\ctx,\act)=-\ln\newpol(\act\mid\ctx)$.

For the contrastive term, condition on the step. Given that the step is at $\ctx$ with action $\act_+$, the failure action is the action at $\ctx$ of a failed rollout conditioned to visit $\ctx$; by \cref{eq:visit-law} with $\reward(\traj)=0$ in place of $\reward(\traj)=1$, its law at any visit is $\pol(\act\mid\ctx)\bigl(1-\Q_\pol(\ctx,\act)\bigr)/\bigl(1-\val{\pol}(\ctx)\bigr)=\pol^{-}(\act\mid\ctx)$, independently of $\act_+$. Hence the contrastive term is \cref{eq:step-decomp} with
\[
f(\ctx,\act_+)=\sum_{\act_-}\pol^{-}(\act_-\mid\ctx)\,\hinge{\marg+\ln\newpol(\act_-\mid\ctx)-\ln\newpol(\act_+\mid\ctx)},
\]
a function of the successful step alone. Adding the two terms gives \cref{eq:loss:unw:app}; the same argument with the reweighted scores gives \cref{eq:loss:rw:app}.

For the independent pairing, with $g$ the hinge and $\traj_\pm$ independent, the contrastive term is
\[
\E\Bigl[\sum_{t,t'}\ind\{\bigstate_t(\traj_+)=\bigstate_{t'}(\traj_-)\}\,g\bigl(\bigstate_t(\traj_+),\bigact_t(\traj_+),\bigact_{t'}(\traj_-)\bigr)\Bigr].
\] Writing the indicator as $\sum_\ctx\ind\{\bigstate_t(\traj_+)=\ctx\}\ind\{\bigstate_{t'}(\traj_-)=\ctx\}$ and using independence, this equals
\[
\sum_\ctx\sum_{\act_+,\act_-}g(\ctx,\act_+,\act_-)\,\E_{+}\Bigl[\sum_t\ind\{\bigstate_t=\ctx,\bigact_t=\act_+\}\Bigr]\,\E_{-}\Bigl[\sum_{t'}\ind\{\bigstate_{t'}=\ctx,\bigact_{t'}=\act_-\}\Bigr],
\]
where $\E_\pm$ are the expectations under the two conditioned laws; by \cref{eq:step-decomp} and its failure analogue, the two factors are $\occ^{+}(\ctx)\pol^{+}(\act_+\mid\ctx)$ and $\occ^{-}(\ctx)\pol^{-}(\act_-\mid\ctx)$.
\end{proof}

\begin{remark}
    We write $\mathcal L_{\unw,\state}(\newpol;\pol)$ and $\mathcal L_{\rw,\state}(\newpol;\pol)$ for the brackets in \cref{eq:loss:unw:app} and \cref{eq:loss:rw:app}, the \emph{state losses}. The weight $\occ^{+}(\state)>0$ multiplies both terms of the state loss, so the coefficients $(\lam,\marg)$ are the same at every state, and $\mathcal L_{\star,\state}$ depends on $\newpol$ only through $\newpol(\cdot\mid\state)$. Hence, over all tabular policies, $\newpol$ minimizes $\lunw$ (resp.\ $\lrw$) if and only if $\newpol(\cdot\mid\state)$ minimizes the state loss at every state with $\occ^{+}(\state)>0$; the minimizers do not depend on the weight, and the same holds for any positive weight $\nu$ in place of $\occ^{+}$, e.g.\ $\nu=\initlaw$ in a contextual bandit. We call such policies \emph{state-wise minimizers}. (Under the independent pairing of the lemma the objective still decouples, but the likelihood term at $\state$ then has coefficient $\lam/\occ^{-}(\state)$; Lemma~\ref{lem:mdp-sft} concerns a common $\lam$, which is why we fix the conditional pair law.) We define $\piunw(\pol)$ and $\pirw(\pol)$ to be state-wise minimizers of $\lunw(\cdot,\pol)$ and $\lrw(\cdot,\pol)$. 
    
    When the infimum of a state loss is not attained, $\pi_\star(\cdot\mid\state)$ ($\star\in\{\unw,\rw\}$) denotes the limit of the policies induced by a minimizing sequence of scores; for $\lam>0$ this limit exists and is unique, and is supported on the positive-quality actions (see the reduction in the proof of Theorem~\ref{thm:comono}, and Lemma~\ref{lem:hinge-optimizer} on its range). For $\lam=0$ minimizers need not be unique, and statements at $\lam=0$ are made for every minimizer.

\end{remark}

\subsection{Auxiliary lemmas}

\begin{lemma}[covariance identity]\label{lem:cov}
For a probability vector $p$ and vectors $f,g$, $\Cov_p(f,g)=\tfrac12\sum_{\act,\act'}p_\act p_{\act'}(f_\act-f_{\act'})(g_\act-g_{\act'})$. In particular $\Cov_p(f,g)\ge0$ whenever $f$ is a nondecreasing function of $g$, with strict inequality if $f$ is strictly increasing in $g$ and $g$ is nonconstant on the support of $p$.
\end{lemma}
\begin{proof}
Expand the right-hand side and use $\sum_\act p_\act=1$.
\end{proof}

\begin{lemma}[performance-difference identity]\label{lem:pdl}
In an episodic MDP, for any two policies $\pol,\pol'$ and any initial law,
$\val{\pol}(\initlaw)-\val{\pol'}(\initlaw)=\sum_{\state}\occ_{\pol,\initlaw}(\state)\,\inner{\pol(\cdot\mid\state)-\pol'(\cdot\mid\state)}{\Q_{\pol'}(\state,\cdot)}$. In particular $\abs{\val{\pol}(\initlaw)-\val{\pol'}(\initlaw)}\le\tfrac12\sum_{\state}\occ_{\pol,\initlaw}(\state)\norm{\pol(\cdot\mid\state)-\pol'(\cdot\mid\state)}_1$, and $\sum_{\state}\occ_{\pol,\initlaw}(\state)=\E_{\pol,\initlaw}T\le\horizon$.
\end{lemma}
\begin{proof}
Let $(S_0,A_0,\dots,S_T)$ be a trajectory of $\pol$ from $\initlaw$, $S_T$ terminal. Extend $\val{\pol'}$ to terminal states by $\val{\pol'}(\state)=\Prob(\reward=1\mid\state)$, so that $\E\val{\pol'}(S_T)=\val{\pol}(\initlaw)$ and $\E\val{\pol'}(S_0)=\val{\pol'}(\initlaw)$. Telescoping, $\val{\pol'}(S_T)-\val{\pol'}(S_0)=\sum_{t<T}\bigl(\val{\pol'}(S_{t+1})-\val{\pol'}(S_t)\bigr)$, a finite sum; conditioning each increment on $(S_t,A_t)$ and using $\E[\val{\pol'}(S_{t+1})\mid S_t,A_t]=\Q_{\pol'}(S_t,A_t)$ and $\sum_\act\pol'(\act\mid\state)\Q_{\pol'}(\state,\act)=\val{\pol'}(\state)$,
\begin{align*}
\val{\pol}(\initlaw)-\val{\pol'}(\initlaw)&=\sum_{t\ge0}\E\bigl[\ind\{t<T\}\bigl(\Q_{\pol'}(S_t,A_t)-\val{\pol'}(S_t)\bigr)\bigr]\\
&=\sum_{\state}\occ_{\pol,\initlaw}(\state)\Bigl(\sum_\act\pol(\act\mid\state)\Q_{\pol'}(\state,\act)-\val{\pol'}(\state)\Bigr) ,
\end{align*}
which is the claim \citep{kakade2002cpi}. The bound follows because each inner product is at most $\tfrac12\norm{\pol(\cdot\mid\state)-\pol'(\cdot\mid\state)}_1\osc(\Q_{\pol'}(\state,\cdot))$ and $\osc\le1$.
\end{proof}

\begin{restatable}[Hinge optimizer at large margin]{lemma}{lemhingeoptimizer}
\label{lem:hinge-optimizer}
Consider a contextual bandit and fix a \emph{non-degenerate} context $\ctx$ and a base policy $\base$. Let $\mathcal A_{+}\coloneqq\{\act: \Q(\ctx, \act) > 0\}$, define $q_{\min}=\min_{\act\in \mathcal A_{+}}\Q(\ctx, \act)$ and define

 \begin{equation}
     \label{eq:lampos}
     \lambda^*=\max_{\act\in \mathcal A_{+}}\left(\frac{\fcond(\act\mid\ctx)}{\scond(\act\mid\ctx)}-\fcond(\mathcal A_{+}\mid\ctx)\right)
 =\frac{\val{\base}/q_{\min}-\base(\mathcal A_{+}\mid\ctx)}{1-\val{\base}}.
 \end{equation}

For $\lambda>\lambda^*$, we define:
\begin{equation}
 \pol^\lambda(\act\mid\ctx)=
 \begin{cases}
 \displaystyle\scond(\act\mid\ctx)+\frac{\fcond(\mathcal A_{+}\mid\ctx)\scond(\act\mid\ctx)-\fcond(\act\mid\ctx)}{\lambda},&\act\in \mathcal A_{+},\\[2mm]
 0,&\act\notin \mathcal A_{+}.
 \end{cases}
 \label{eq:explicit-policy}
\end{equation}
This is a probability distribution, strictly positive on $\mathcal A_{+}$.
For either readout, let
\[
 D_\rw(\lambda)\coloneqq\osc_{\act\in \mathcal A_{+}}
    \left\{\log\pol^\lambda(\act\mid\ctx) - \log\base(\act\mid\ctx)\right\},\,
 D_\unw(\lambda)\coloneqq\osc_{\act\in \mathcal A_{+}}
    \log\pol^\lambda(\act\mid\ctx).
\]
If $\delta\ge \max\{D_\rw(\lambda), D_\unw(\lambda)\}$, then every policy $\pol$ satisfies
\begin{equation}
 \lrw(\pol)-\inf_{\hat\pol}\lrw(\hat\pol)
 \ge\lambda\KL\bigl(\pol^\lambda\Vert \pol\bigr),\,\lunw(\pol)-\inf_{\hat\pol}\lunw(\hat\pol)
 \ge\lambda\KL\bigl(\pol^\lambda\Vert \pol\bigr)
 \label{eq:hinge-certificate}
\end{equation}
If $\mathcal A_{+}$ is the entire action set, the unique optimal policy is $\pol^\lambda$.
If $\mathcal A_{+}$ is proper, no finite score vector minimizes the risk, but every minimizing sequence induces policies converging to $\pol^\lambda$.
\end{restatable}

\begin{proof}
The definitions of the success- and failure-conditioned policies give,
for every $\act\in \mathcal A_{+}$,
\[
\frac{\fcond(\act\mid\ctx)}{\scond(\act\mid\ctx)}
-\fcond(\mathcal A_{+}\mid\ctx)
=
\frac{
\val{\base}/\Q(\ctx,\act)-\base(\mathcal A_{+}\mid\ctx)}
{1-\val{\base}}.
\]
Taking the maximum proves the stated expression for $\lambda^*$.
Moreover,
$\val{\base}\ge q_{\min}\base(\mathcal A_{+}\mid\ctx)$,
so $\lambda^*\ge0$. For every $\act\in \mathcal A_{+}$,
\[
\frac{\pol^\lambda(\act\mid\ctx)}
     {\scond(\act\mid\ctx)}
=
1-\frac1\lambda
\left(
\frac{\fcond(\act\mid\ctx)}{\scond(\act\mid\ctx)}
-\fcond(\mathcal A_{+}\mid\ctx)
\right)
\ge
1-\frac{\lambda^*}{\lambda}>0.
\]
Since $\scond(\mathcal A_{+}\mid\ctx)=1$, summing
\cref{eq:explicit-policy} over $\mathcal A_{+}$ gives
\[
\sum_{\act\in \mathcal A_{+}}\pol^\lambda(\act\mid\ctx)
=
1+
\frac{\fcond(\mathcal A_{+}\mid\ctx)-\fcond(\mathcal A_{+}\mid\ctx)}{\lambda}
=1.
\]
Thus $\pol^\lambda$ is a probability distribution with support $\mathcal A_{+}$.

We next derive a lower bound for each loss.
Every hinge is at least its affine argument.
Apply this inequality to the pairs whose two actions belong to $\mathcal A_{+}$,
and discard the nonnegative hinge terms whose failure action lies
outside $\mathcal A_{+}$.
There are no terms with a success action outside $\mathcal A_{+}$, because
$\scond$ vanishes there.

For the unweighted loss, collecting the coefficients of
$\log\pol(\act\mid\ctx)$ gives
\[
\fcond(\act\mid\ctx)
-
\bigl(\lambda+\fcond(\mathcal A_{+}\mid\ctx)\bigr)
\scond(\act\mid\ctx)
=
-\lambda\pol^\lambda(\act\mid\ctx).
\]
Consequently,
\[
\lunw(\pol)
\ge
\delta\fcond(\mathcal A_{+}\mid\ctx)
-
\lambda\sum_{\act\in \mathcal A_{+}}
\pol^\lambda(\act\mid\ctx)\log\pol(\act\mid\ctx).
\]
For the reweighted loss, the same calculation leaves the additional
base-policy term:
\begin{align*}
\lrw(\pol)
\ge{}&
\delta\fcond(\mathcal A_{+}\mid\ctx)
-
\lambda\sum_{\act\in \mathcal A_{+}}
\pol^\lambda(\act\mid\ctx)\log\pol(\act\mid\ctx)
\\
&+
\lambda\sum_{\act\in \mathcal A_{+}}
\bigl(
\pol^\lambda(\act\mid\ctx)-\scond(\act\mid\ctx)
\bigr)
\log\base(\act\mid\ctx).
\end{align*}
In both inequalities, use
\begin{align*}
-\lambda\sum_{\act\in \mathcal A_{+}}
\pol^\lambda(\act\mid\ctx)\log\pol(\act\mid\ctx)
=
-\lambda\sum_{\act\in \mathcal A_{+}}
\pol^\lambda(\act\mid\ctx)
\log\pol^\lambda(\act\mid\ctx) +
\lambda\KL\bigl(\pol^\lambda\Vert\pol\bigr).
\end{align*}
It remains to show that the resulting policy-independent lower bounds
are the respective infima.

First suppose that $\mathcal A_{+}$ is the entire action set.
At $\pol=\pol^\lambda$, the margin condition ensures that every
unweighted score difference and every reweighted score difference
has absolute value at most $\delta$.
Hence every hinge equals its affine argument, and both lower bounds
are attained.
Subtracting these attained bounds proves
\cref{eq:hinge-certificate}.
Since $\lambda>0$, equality at the minimum forces
$\KL(\pol^\lambda\Vert\pol)=0$, hence $\pol=\pol^\lambda$.
This proves uniqueness.

Now suppose that $\mathcal A_{+}$ is proper.
Consider strictly positive policies $\hat\pol$ of the form
\[
\hat\pol(\act\mid\ctx)=
\begin{cases}
\hat\pol(\mathcal A_{+}\mid\ctx)\pol^\lambda(\act\mid\ctx),
&\act\in \mathcal A_{+},\\[1mm]
\displaystyle
\frac{1-\hat\pol(\mathcal A_{+}\mid\ctx)}
     {1-\base(\mathcal A_{+}\mid\ctx)}
\base(\act\mid\ctx),
&\act\notin \mathcal A_{+},
\end{cases}
\]
and let $\hat\pol(\mathcal A_{+}\mid\ctx)$ increase to one.

Within $\mathcal A_{+}$, their score differences are exactly those of
$\pol^\lambda$, for both losses.
The margin condition therefore makes every internal hinge equal
its affine argument.
Meanwhile, the score of every action outside $\mathcal A_{+}$ tends to
$-\infty$, whereas the scores on $\mathcal A_{+}$ remain finite.
Thus every hinge pairing a success action in $\mathcal A_{+}$ with a failure
action outside $\mathcal A_{+}$ is eventually zero.

It follows that both affine lower bounds are equalities along
these policies once $\hat\pol(\mathcal A_{+}\mid\ctx)$ is sufficiently close
to one. Furthermore,
\[
\KL\bigl(\pol^\lambda\Vert\hat\pol\bigr)
=
-\log\hat\pol(\mathcal A_{+}\mid\ctx)
\longrightarrow0.
\]
Therefore the policy-independent lower bounds are again the
respective infima. Subtracting them proves
\cref{eq:hinge-certificate}.

Every finite-score policy is strictly positive and hence assigns
positive mass outside $\mathcal A_{+}$.
It therefore differs from $\pol^\lambda$, so the KL divergence in
\cref{eq:hinge-certificate} is strictly positive.
Neither loss can attain its infimum at finite scores.

Finally, along any minimizing sequence for either loss,
\cref{eq:hinge-certificate} forces the KL divergence from
$\pol^\lambda$ to tend to zero.
Pinsker's inequality then gives convergence of the induced
policies to $\pol^\lambda$ in total variation.
\end{proof}

\begin{lemma}[Margin thresholds uniform in $\lam$]\label{lem:uniform-margin}
In the setting of Lemma~\ref{lem:hinge-optimizer}, with $\mathcal A_{+}=\{\act\in\actset:\Q(\state,\act)>0\}$ the set of positive-quality actions and $\lam^*$ as in \cref{eq:lampos}, define
\begin{equation}\label{eq:thresholds}
\begin{aligned}
\bar\lambda(\state)&\coloneqq\max\{2\lam^*,2\},\\
\bar\delta_{\rw}(\state)&\coloneqq\ln\frac{\max_{\act\in\mathcal A_{+}}\Q(\state,\act)}{\min_{\act\in\mathcal A_{+}}\Q(\state,\act)}+\ln 3,\quad
\bar\delta_{\unw}(\state)\coloneqq\ln\frac{\max_{\act\in\mathcal A_{+}}\base(\act\mid\state)\Q(\state,\act)}{\min_{\act\in\mathcal A_{+}}\base(\act\mid\state)\Q(\state,\act)}+\ln 3 .
\end{aligned}
\end{equation}
Then $D_{\rw}(\lam)\le\bar\delta_{\rw}(\state)$ and $D_{\unw}(\lam)\le\bar\delta_{\unw}(\state)$ for every $\lam\ge\bar\lambda(\state)$ (where $D_\rw, D_\unw$ are as defined in \cref{lem:hinge-optimizer}). Hence, for $\lam\ge\bar\lambda(\state)$, $\pirw(\cdot\mid\state)=\pol^{\lam}(\cdot\mid\state)$ whenever $\marg\ge\bar\delta_{\rw}(\state)$, and $\piunw(\cdot\mid\state)=\pol^{\lam}(\cdot\mid\state)$ whenever $\marg\ge\bar\delta_{\unw}(\state)$; in particular $\pirw(\cdot\mid\state)=\piunw(\cdot\mid\state)=\pol^{\lam}(\cdot\mid\state)$ whenever $\lam\ge\bar\lambda(\state)$ and $\marg\ge\max\{\bar\delta_{\rw}(\state),\bar\delta_{\unw}(\state)\}$. These thresholds are sufficient conditions, not claims of necessity.
\end{lemma}
\begin{proof}
Fix the state $\state$. For $\act\in\mathcal A_{+}$, \cref{eq:explicit-policy} gives
\[
\frac{\pol^{\lam}(\act\mid\state)}{\scond(\act\mid\state)}=1-\frac1\lam\Bigl(\frac{\fcond(\act\mid\state)}{\scond(\act\mid\state)}-\fcond(\mathcal A_{+}\mid\state)\Bigr).
\]
By \cref{eq:lampos}, $\frac{\fcond(\act\mid\state)}{\scond(\act\mid\state)}-\fcond(\mathcal A_{+}\mid\state)\in[-\fcond(\mathcal A_{+}\mid\state),\lam^*]\subseteq[-1,\lam^*]$, so for $\lam\ge\max\{2\lam^*,2\}$ the ratio lies in $[\tfrac12,\tfrac32]$ and $\ln\pol^{\lam}(\act\mid\state)-\ln\scond(\act\mid\state)$ takes values in an interval of length $\ln3$. Hence the oscillation over $\mathcal A_{+}$ of $\ln\pol^{\lam}$ (respectively, of $\ln\pol^{\lam}-\ln\base$) exceeds that of $\ln\scond$ (respectively, of $\ln\scond-\ln\base$) by at most $\ln3$, that is, $D_{\unw}(\lam)\le\osc_{\mathcal A_{+}}\ln\scond+\ln3$ and $D_{\rw}(\lam)\le\osc_{\mathcal A_{+}}(\ln\scond-\ln\base)+\ln3$. Since $\ln\scond(\act\mid\state)=\ln\base(\act\mid\state)+\ln\Q(\state,\act)-\ln\val{\base}(\state)$ on $\mathcal A_{+}$ by \cref{eq:cond}, these two oscillations are $\ln\bigl(\max_{\mathcal A_{+}}\base\Q/\min_{\mathcal A_{+}}\base\Q\bigr)$ and $\ln\bigl(\max_{\mathcal A_{+}}\Q/\min_{\mathcal A_{+}}\Q\bigr)$, which gives the two bounds. The remaining claims follow from Lemma~\ref{lem:hinge-optimizer}: for $\lam\ge\bar\lambda(\state)>\lam^*$ and $\marg\ge\bar\delta_{\rw}(\state)\ge D_{\rw}(\lam)$, its certificate for the \reweighted\ loss shows that every minimizing sequence converges to $\pol^{\lam}$, so $\pirw(\cdot\mid\state)=\pol^{\lam}(\cdot\mid\state)$; likewise for the \unweighted\ loss.
\end{proof}

\subsection{Proofs concerning comparisons of $\scond, \base, \pirw, \piunw$}
\label{app:proofs:guarantees}

\thmrwbase*
\begin{proof}
Fix a decision state $\state$. First suppose that the state-wise
minimum is attained at finite scores. We claim that
\[
\Qb(\state,\act)>\Qb(\state,\act')
\implies
\frac{\pirw(\act\mid\state)}{\base(\act\mid\state)}
\ge
\frac{\pirw(\act'\mid\state)}{\base(\act'\mid\state)}.
\]

Suppose otherwise. The reweighted score of $\act$ is then strictly
smaller than that of $\act'$. Increase the unnormalized reweighted
score of $\act$ at rate $1/\base(\act\mid\state)$ and decrease that
of $\act'$ at rate $1/\base(\act'\mid\state)$, keeping the other
unnormalized scores fixed and renormalizing the resulting policy.
We show that the right directional derivative of $\lrw$ along
this perturbation is strictly negative.

Consider first the contrastive terms involving a third action.
When the third action is the success action, the hinge argument
involving $\act$ as the failure action is strictly smaller than
the corresponding argument involving $\act'$.
Moreover,
\[
\frac{\fcond(\act\mid\state)}{\base(\act\mid\state)}
=
\frac{1-\Qb(\state,\act)}{1-\val{\base}(\state)}
<
\frac{1-\Qb(\state,\act')}{1-\val{\base}(\state)}
=
\frac{\fcond(\act'\mid\state)}{\base(\act'\mid\state)}.
\]
The perturbation increases the smaller hinge argument and
decreases the larger one. Since hinge slopes are nonnegative
and nondecreasing, the sum of these two directional derivatives
is nonpositive.

When the third action is the failure action, the hinge argument
involving $\act$ as the success action is strictly larger than
the corresponding argument involving $\act'$.
Here,
\[
\frac{\scond(\act\mid\state)}{\base(\act\mid\state)}
=
\frac{\Qb(\state,\act)}{\val{\base}(\state)}
>
\frac{\Qb(\state,\act')}{\val{\base}(\state)}
=
\frac{\scond(\act'\mid\state)}{\base(\act'\mid\state)}.
\]
The perturbation decreases the larger hinge argument and
increases the smaller one. Their combined directional derivative
is again nonpositive. These comparisons remain valid at hinge
kinks: the right slope at a strictly smaller argument is at most
the left slope at a strictly larger argument.

The self-pair terms are constant. For the two terms pairing
$\act$ and $\act'$ with each other, the hinge with success action
$\act$ and failure action $\act'$ is strictly active, because
$\delta\ge0$ and the score of $\act$ is smaller.
Its slope is therefore one, while the slope of the reverse
hinge is at most one. Their combined directional derivative is
at most
\begin{align*}
&
\left(
\frac{1}{\base(\act\mid\state)}
+
\frac{1}{\base(\act'\mid\state)}
\right)
\left(
\scond(\act'\mid\state)\fcond(\act\mid\state)
-
\scond(\act\mid\state)\fcond(\act'\mid\state)
\right)
\\
&\quad =
-
\frac{
\bigl(\base(\act\mid\state)+\base(\act'\mid\state)\bigr)
\bigl(\Qb(\state,\act)-\Qb(\state,\act')\bigr)}
{\val{\base}(\state)\bigl(1-\val{\base}(\state)\bigr)}
<0.
\end{align*}
Thus the contrastive term has strictly negative directional
derivative.

The SFT term, including its coefficient $\lambda$, has
directional derivative
\[
\lambda\left(
\frac{\pirw(\act\mid\state)}{\base(\act\mid\state)}
-
\frac{\pirw(\act'\mid\state)}{\base(\act'\mid\state)}
-
\frac{\Qb(\state,\act)-\Qb(\state,\act')}
     {\val{\base}(\state)}
\right)
\le0.
\]
Consequently, the right directional derivative of the entire
loss is strictly negative, contradicting optimality.
This proves the claimed ordering for every $\lambda,\delta\ge0$.

We next address limiting optimal policies. If $\pirw$ assigns
positive probability only to actions with
$\Qb(\state,\act)=1$, then
\[
\sum_{\act}\pirw(\act\mid\state)\Qb(\state,\act)
=1\ge\val{\base}(\state).
\]
Otherwise, $\pirw$ retains an action whose base quality is
strictly below one. Every action with positive base quality
must then also have positive probability under $\pirw$:
if such an action were omitted, its reweighted score relative
to the retained action would tend to $-\infty$ along the
minimizing sequence. The hinge pairing the omitted success
action with the retained failure action has a strictly positive
weight and would diverge, contradicting minimization.

Hence, in this remaining case, every omitted action has base
quality zero. The limiting policy minimizes the loss restricted
to its positive-probability actions. Indeed, conditioning any
strictly positive policy on those actions leaves their internal
hinges unchanged, removes nonnegative external hinge terms,
and cannot increase the SFT risk. Conversely, any strictly
positive policy on the retained actions can be extended by
vanishing mass on the omitted actions; their failure scores
then tend to $-\infty$, so the external hinges vanish.
The full and restricted infima therefore coincide, and
continuity on the retained support proves restricted optimality.

The preceding directional argument applies on that support.
Every omitted action has both zero base quality and zero
likelihood ratio, so the same ordering extends to all actions.

Whenever this ordering applies, normalization gives
\begin{align*}
&\sum_{\act}
\bigl(\pirw(\act\mid\state)-\base(\act\mid\state)\bigr)
\Qb(\state,\act)
\\
&\quad =
\frac12\sum_{\act,\act'}
\base(\act\mid\state)\base(\act'\mid\state)
\left(
\frac{\pirw(\act\mid\state)}{\base(\act\mid\state)}
-
\frac{\pirw(\act'\mid\state)}{\base(\act'\mid\state)}
\right)
\left(
\Qb(\state,\act)-\Qb(\state,\act')
\right)
\ge0.
\end{align*}
Together with the case where all retained actions have quality
one, this proves, at every decision state,
\[
\sum_{\act}\pirw(\act\mid\state)\Qb(\state,\act)
\ge
\val{\base}(\state).
\]

Now, we fix a decision state $\state$ and apply the performance-difference identity of Lemma~\ref{lem:pdl} with $\initlaw$ the point mass at $\state$, $\pol=\pirw$ and $\pol'=\base$ to get:
\[
\val{\pirw}(\state)-\val{\base}(\state)=\sum_{\state'}\occ_{\pirw,\initlaw}(\state')\,\inner{\pirw(\cdot\mid\state')-\base(\cdot\mid\state')}{\Qb(\state',\cdot)}\ \ge\ 0,
\]
because each inner product on the right is the state-wise inequality just proved, at $\state'$, and the occupancies are nonnegative. This holds at every state, with no assumption on the order in which states are visited.
\end{proof}

\thmbeatsft*
\begin{proof}
Fix the context $\state$ and call an action $\act$ \emph{positive-quality} if
$\Q(\state,\act)>0$.  By definition,
$\scond(\act\mid\state)=\base(\act\mid\state)\Q(\state,\act)/\val{\base}(\state)$ and
$\fcond(\act\mid\state)=\base(\act\mid\state)\bigl(1-\Q(\state,\act)\bigr)/\bigl(1-\val{\base}(\state)\bigr)$,
so $\scond(\cdot\mid\state)$ is supported exactly on the positive-quality actions, and
\begin{equation}\tag{a}
\sum_{\act:\,\Q(\state,\act)>0}\base(\act\mid\state)\,\Q(\state,\act)=\val{\base}(\state),
\quad
\sum_{\act:\,\Q(\state,\act)>0}\fcond(\act\mid\state)
=\frac{\sum_{\act:\,\Q(\state,\act)>0}\base(\act\mid\state)-\val{\base}(\state)}{1-\val{\base}(\state)} .
\end{equation}

Let $\mathcal A_{+}$ be the set of all actions with positive rewards. By Lemma~\ref{lem:uniform-margin}, for $\lam\ge\bar\lambda(\state)$
     and $\marg\ge\bar\delta_\star(\state)$, Lemma~\ref{lem:hinge-optimizer}
     applies to $\mathcal L_{\star,\state}$, so $\pi_\star(\cdot\mid\state)
     =\pol^{\lam}(\cdot\mid\state)$: it puts zero mass off $\mathcal A_{+}$ and,
     on every $\act\in\mathcal A_{+}$, it puts the mass:
     
\begin{equation}\tag{b}
\pi_\star(\act\mid\state)=\scond(\act\mid\state)
+\frac1\lam\Bigl(\scond(\act\mid\state)\sum_{\act':\,\Q(\state,\act')>0}\fcond(\act'\mid\state)
-\fcond(\act\mid\state)\Bigr).
\end{equation}
The same expression serves both losses, so it suffices to prove the claims for the
policy (b).  Substituting the definitions and the second identity in (a), the bracket
in (b) equals
\begin{align*}
&\frac{\base(\act\mid\state)}{\val{\base}(\state)\bigl(1-\val{\base}(\state)\bigr)}
\biggl(
\Q(\state,\act)\Bigl(\sum_{\act':\,\Q(\state,\act')>0}\base(\act'\mid\state)-\val{\base}(\state)\Bigr)
-\val{\base}(\state)\bigl(1-\Q(\state,\act)\bigr)
\biggr)
\\
&=
\frac{\base(\act\mid\state)\Bigl(\Q(\state,\act)\sum_{\act':\,\Q(\state,\act')>0}\base(\act'\mid\state)-\val{\base}(\state)\Bigr)}
{\val{\base}(\state)\bigl(1-\val{\base}(\state)\bigr)} .
\end{align*}
Multiplying by $\Q(\state,\act)$, summing over the positive-quality actions, and
using the first identity in (a), the value of the policy (b) exceeds
$\val{\scond}(\state)$ by exactly
\begin{equation}\tag{c}
\frac{1}{\lam\,\val{\base}(\state)\bigl(1-\val{\base}(\state)\bigr)}
\biggl(
\sum_{\act':\,\Q(\state,\act')>0}\base(\act'\mid\state)
\sum_{\act:\,\Q(\state,\act)>0}\base(\act\mid\state)\,\Q(\state,\act)^2
-\val{\base}(\state)^2
\biggr).
\end{equation}

We first prove the inequalities in the items 1 and 2.  By the Cauchy--Schwarz inequality with the positive weights
$\base(\act\mid\state)$ on the positive-quality actions,
\begin{align*}
\val{\base}(\state)^2
=\Bigl(\sum_{\act:\,\Q(\state,\act)>0}\base(\act\mid\state)\,\Q(\state,\act)\Bigr)^2
\le
\sum_{\act':\,\Q(\state,\act')>0}\base(\act'\mid\state)
\sum_{\act:\,\Q(\state,\act)>0}\base(\act\mid\state)\,\Q(\state,\act)^2,
\end{align*}
with equality if and only if $\Q(\state,\cdot)$ is constant on the positive-quality
actions.  Hence (c) is nonnegative, which is
$\val{\pirw}(\state)\geq\val{\scond}(\state)$ and
$\val{\piunw}(\state)\geq\val{\scond}(\state)$ on the respective ranges of
$(\lam,\marg)$, with strict inequality whenever $\Q(\state,\cdot)$ is non-constant on
the positive-quality actions.

Now we prove the variance bound.  The variance of a random variable is at most its second
moment about any fixed number.  Taking the number
$\val{\base}(\state)\big/\sum_{\act':\,\Q(\state,\act')>0}\base(\act'\mid\state)$,
and using $\scond(\act\mid\state)=\base(\act\mid\state)\Q(\state,\act)/\val{\base}(\state)$
together with $\Q(\state,\act)\le1$,
\begin{align*}
\Var_{\act\sim\scond(\cdot\mid\state)}\bigl(\Q(\state,\act)\bigr)
&\le\frac{1}{\val{\base}(\state)}
\sum_{\act:\,\Q(\state,\act)>0}\base(\act\mid\state)\,\Q(\state,\act)
\Bigl(\Q(\state,\act)-\frac{\val{\base}(\state)}{\sum_{\act':\,\Q(\state,\act')>0}\base(\act'\mid\state)}\Bigr)^2\\
&\le\frac{1}{\val{\base}(\state)}
\sum_{\act:\,\Q(\state,\act)>0}\base(\act\mid\state)
\Bigl(\Q(\state,\act)-\frac{\val{\base}(\state)}{\sum_{\act':\,\Q(\state,\act')>0}\base(\act'\mid\state)}\Bigr)^2 .
\end{align*}
Expanding the square and using (a), the last sum equals
\begin{align*}
\sum_{\act:\,\Q(\state,\act)>0}\base(\act\mid\state)\,\Q(\state,\act)^2
-\frac{\val{\base}(\state)^2}{\sum_{\act':\,\Q(\state,\act')>0}\base(\act'\mid\state)},
\end{align*}
and therefore
\begin{align*}
&\sum_{\act':\,\Q(\state,\act')>0}\base(\act'\mid\state)
\sum_{\act:\,\Q(\state,\act)>0}\base(\act\mid\state)\,\Q(\state,\act)^2
-\val{\base}(\state)^2\\
&\ge
\val{\base}(\state)\sum_{\act':\,\Q(\state,\act')>0}\base(\act'\mid\state)\;
\Var_{\act\sim\scond(\cdot\mid\state)}\bigl(\Q(\state,\act)\bigr).
\end{align*}
Inserting this into (c), and then using
$\val{\base}(\state)\le\sum_{\act':\,\Q(\state,\act')>0}\base(\act'\mid\state)$
(by (a) and $\Q(\state,\act)\le1$), we obtain, on the
respective ranges of $(\lam,\marg)$:
\begin{align*}
\min&\left\{\val{\pi_\rw}(\state)-\val{\scond}(\state), \val{\pi_\unw}(\state)-\val{\scond}(\state)\right\}\\
&\ge
\frac{\sum_{\act':\,\Q(\state,\act')>0}\base(\act'\mid\state)}{\lam\bigl(1-\val{\base}(\state)\bigr)}\,
\Var_{\act\sim\scond(\cdot\mid\state)}\bigl(\Q(\state,\act)\bigr)\\
&\ge
\frac{\val{\base}(\state)}{\lam\bigl(1-\val{\base}(\state)\bigr)}\,
\Var_{\act\sim\scond(\cdot\mid\state)}\bigl(\Q(\state,\act)\bigr),
\end{align*}
which is the claimed lower bound; it is strictly positive exactly when
$\Q(\state,\cdot)$ is non-constant on the positive-quality actions.
\end{proof}

\begin{restatable}{lemma}{lemunwbase}\label{lem:unw-base}
Consider a contextual bandit instance with a single state $\state$ and two actions $\act_1,\act_2$ with $1 >\Q(\state,\act_1)>\Q(\state, \act_2)>0$. Let the base policy $\base$ be such that $\base(\act_1\mid\ctx)>\base(\act_2\mid\ctx)>0$. Then there exist $\lam^*>0,\,\marg^*>0$ such that $\val{\piunw}<\val{\base}$ for every $\marg<\marg^*$ and $\lam<\lam^{*}$.
\end{restatable}
\begin{proof}
We write $\Q_1$ and $\Q_2$ in short to denote $\Q(\ctx, \act_1)$ and $\Q(\ctx, \act_2)$ respectively. Write $A=\scond(1)\fcond(2)$ and $B=\scond(2)\fcond(1)$ for the probabilities of a correctly and of a wrongly ordered pair, and $d_0=\log(\base(1)/\base(2))$ which means $\base(1)=\sig(d_0)$. Note that we have $A/B=\Q_1(1-\Q_2)/(\Q_2(1-\Q_1))>1$.

Let $\score_\unw$ be the optimizer for $\ln\newpol$ in $\lunw$ in \cref{eq:loss:unw}, giving $\newpol(\act\mid\ctx)=\frac{\exp\left(\score_\unw(\ctx, \act)\right)}{\sum_{\act'\in\actset}\exp\left(\score_\unw(\ctx, \act')\right)}$. We write $\score_1, \score_2$ to denote $\score_\unw(\ctx, \act_1), \score_\unw(\ctx, \act_2)$ respectively. Writing the score gap $g\coloneqq\score_1-\score_2$, the \unweighted policy has mass $\sig(g)$ on the better action, and its combined risk, up to a constant, can be written as
\begin{equation*}
\lunw(g)=A\hinge{\marg-g}+B\hinge{\marg+g}+\lam\bigl(\log(1+e^{g})-\scond(1)g\bigr):
\end{equation*}
 The self-pairs  in the contrastive term add constant terms, and $\KL(\scond\|\sig(g))=\log(1+e^{g})-\scond(1)g+\mathrm{const}$. 
 
 $\lunw$ is convex and $\lunw(g)\to\infty$ if $|g|\to\infty$, which means it has interior optimizers. For $\lam > 0$ the objective is strictly convex and the optimizer $\hat g$ is unique. By convexity all of them lie strictly to the left of $d_0$ if and only if the left derivative of $\lunw$ at $d_0$ is positive. Note that, for the two action bandit, for any given policy $\pol$ we have: $\val{\pol}-\Vb=(\pol(1)-\base(1))(\Q_1-\Q_2)$. Now, by our ``score-gap'' notation, we have: $\base(1)=\sig(d_0)$ and $\newpol(1)=\sig(\hat g)$ for the learned \unweighted policy $\newpol$ and its corresponding ``score-gap'' $\hat g$. This gives us the equivalence: $$\val{\piunw}<\Vb\,\iff\,\sig(\hat g) < \sig(d_0)\,\iff\, \hat g < d_0.$$

Now, suppose $d_0>\marg$, i.e.\ $\base(1)>\sig(\marg)$. Near $d_0$ the hinge $\hinge{\marg-g}$ vanishes and $\hinge{\marg+g}$ has slope $1$, so the contrastive part contributes $B$ to the left derivative of $\lunw$ at $d_0$, while the KL anchor contributes $\lam(\sig(d_0)-\scond(1))=-\lam(\scond(1)-\base(1))$ with $\scond(1)-\base(1)=\base(1)(\Q_1-\Vb)/\Vb=\base(1)\base(2)(\Q_1-\Q_2)/\Vb>0$. Hence we can write the positivity condition on the left-derivative of $\lunw$ as follows:
\begin{equation*}
\lunw^{',-}(d_0)=B-\lam(\scond(1)-\base(1))>0\,\iff\,\lam<\lam^{*}=\frac{B}{\scond(1)-\base(1)}=\frac{\Q_2(1-\Q_1)}{(1-\Vb)(\Q_1-\Q_2)},
\end{equation*}
where the last expression uses $B=\base(1)\base(2)\,\Q_2(1-\Q_1)/(\Vb(1-\Vb))$. Thus, we see that, defining $\marg^*\coloneqq\ln(\base(1)/\base(2))$ and $\lam^*\coloneqq\frac{\Q_2(1-\Q_1)}{(1-\Vb)(\Q_1-\Q_2)}$ gives us that $\val{\piunw} < \val{\base},\,\forall\,(\lam,\marg)\in(0,\lam^*)\times(0,\marg^*)$. 

For $\lam=0$ the objective $A\hinge{\marg-g}+B\hinge{\marg+g}$ has slope $-A$ on $(-\infty,-\marg)$, $B-A<0$ on $(-\marg,\marg)$ and $B>0$ on $(\marg,\infty)$, so its unique minimizer is $g=\marg<d_0$; this includes $\marg=0$. Hence the conclusion holds on $[0,\lam^*)\times[0,\marg^*)$.

\end{proof}

\lemrwsft*

\begin{proof}
Every policy on the two actions has $\pol(\act_2\mid\state)=1-\pol(\act_1\mid\state)$,
hence
$\val{\pol}=\Q(\state,\act_2)+\pol(\act_1\mid\state)\bigl(\Q(\state,\act_1)-\Q(\state,\act_2)\bigr)$.
Since $\Q(\state,\act_1)>\Q(\state,\act_2)$, the claim $\val{\pirw}<\val{\scond}$ is
equivalent to
\begin{equation}\tag{$\ast$}
\pirw(\act_1\mid\state)\;<\;\scond(\act_1\mid\state)
=\frac{\base(\act_1\mid\state)\,\Q(\state,\act_1)}{\val{\base}} .
\end{equation}

By definition of the \reweighted hinge loss, the positive action is drawn from
$\scond$ and the negative action from the failure-conditioned policy
$\act_i\mapsto\base(\act_i\mid\state)\bigl(1-\Q(\state,\act_i)\bigr)/\bigl(1-\val{\base}\bigr)$,
independently.  Expanding the expectation over the four pairs, the loss of a policy
$\pol$ equals
\begin{align*}
&\scond(\act_1\mid\state)\,
\frac{\base(\act_2\mid\state)\bigl(1-\Q(\state,\act_2)\bigr)}{1-\val{\base}}
\Bigl[\marg-\ln\frac{\pol(\act_1\mid\state)}{\base(\act_1\mid\state)}
+\ln\frac{\pol(\act_2\mid\state)}{\base(\act_2\mid\state)}\Bigr]_+\\
&{}+\scond(\act_2\mid\state)\,
\frac{\base(\act_1\mid\state)\bigl(1-\Q(\state,\act_1)\bigr)}{1-\val{\base}}
\Bigl[\marg+\ln\frac{\pol(\act_1\mid\state)}{\base(\act_1\mid\state)}
-\ln\frac{\pol(\act_2\mid\state)}{\base(\act_2\mid\state)}\Bigr]_+\\
&{}-\lam\Bigl(\scond(\act_1\mid\state)\ln\pol(\act_1\mid\state)
+\scond(\act_2\mid\state)\ln\pol(\act_2\mid\state)\Bigr)
\;+\;\text{a constant},
\end{align*}
where the constant is the contribution of the two pairs with equal positive and
negative action, whose hinge argument is $\marg\geq0$.

The two hinge arguments are $\marg$ minus and $\marg$ plus the same quantity
\begin{align*}
\ln\frac{\pol(\act_1\mid\state)}{\pol(\act_2\mid\state)}
-\ln\frac{\base(\act_1\mid\state)}{\base(\act_2\mid\state)},
\end{align*}
which is a continuous, strictly increasing function of $\pol(\act_1\mid\state)$ and,
by $(\ast)$ together with
$\scond(\act_2\mid\state)=\base(\act_2\mid\state)\Q(\state,\act_2)/\val{\base}$,
equals $\ln\bigl(\Q(\state,\act_1)/\Q(\state,\act_2)\bigr)$ at
$\pol(\act_1\mid\state)=\scond(\act_1\mid\state)$.  Because
$\ln\bigl(\Q(\state,\act_1)/\Q(\state,\act_2)\bigr)>\marg\geq0$, every policy with
$\pol(\act_1\mid\state)\geq\scond(\act_1\mid\state)$ --- and, by continuity, every
policy with $\pol(\act_1\mid\state)$ slightly below $\scond(\act_1\mid\state)$ --- has
a negative first hinge argument and a positive second one.  On this range the first
hinge vanishes, the second equals its argument, and the loss is a differentiable
function of $\pol(\act_1\mid\state)$ whose derivative is
\begin{align*}
\frac{\scond(\act_2\mid\state)\,\base(\act_1\mid\state)\bigl(1-\Q(\state,\act_1)\bigr)}
{\bigl(1-\val{\base}\bigr)\,\pol(\act_1\mid\state)\bigl(1-\pol(\act_1\mid\state)\bigr)}
\;+\;\lam\,\frac{\pol(\act_1\mid\state)-\scond(\act_1\mid\state)}
{\pol(\act_1\mid\state)\bigl(1-\pol(\act_1\mid\state)\bigr)}\,,
\end{align*}
the second term following from $\scond(\act_1\mid\state)+\scond(\act_2\mid\state)=1$.
The first term is strictly positive, because $\base$ has full support and
$\Q(\state,\act_1),\Q(\state,\act_2)\in(0,1)$; the second is nonnegative whenever
$\pol(\act_1\mid\state)\geq\scond(\act_1\mid\state)$, because $\lam\geq0$.  Hence the
loss is strictly increasing in $\pol(\act_1\mid\state)$ on
$[\scond(\act_1\mid\state),1)$, and its derivative is continuous and strictly positive
at $\pol(\act_1\mid\state)=\scond(\act_1\mid\state)$, so a policy with
$\pol(\act_1\mid\state)$ slightly below $\scond(\act_1\mid\state)$ has strictly smaller
loss than every policy with $\pol(\act_1\mid\state)\geq\scond(\act_1\mid\state)$.
Therefore no minimizer of the \reweighted hinge loss satisfies
$\pol(\act_1\mid\state)\geq\scond(\act_1\mid\state)$; in particular $(\ast)$ holds for
$\pirw$, which proves the lemma.
\end{proof}

\thmcomono*
\begin{proof}
Both objectives decouple across states, so we fix a state $\state$ and prove the inequality there.  Comonotonicity of $\base$ and $\Q$ at $\state$ says that
\begin{equation}\tag{a}\label{eq:comono:i}
\bigl(\Q(\state,\act)-\Q(\state,\act')\bigr)\bigl(\base(\act\mid\state)-\base(\act'\mid\state)\bigr)\ \ge\ 0\quad\text{for all }\act,\act'\in\actset,
\end{equation}
and the second hypothesis of the theorem reverses this inequality. We treat the aligned case; reversing every sign below gives the other one.

 Write $\mathcal A_{+}\coloneqq\{\act\in\actset:\Q(\state,\act)>0\}$ and $\mathcal L_\state$ for the state loss under consideration (either of the two contrastive losses). By \cref{eq:cond}, $\scond(\cdot\mid\state)$ is supported exactly on $\mathcal A_{+}$ and is positive there, while $\fcond(\cdot\mid\state)$ may vanish on $\mathcal A_{+}$ (at actions with $\Q(\state,\act)=1$) and $\fcond(\mathcal A_{+}\mid\state)\le1$. Let $\mathcal L_{\mathcal A_{+}}$ be $\mathcal L_\state$ with both sums restricted to $\mathcal A_{+}$, without renormalizing the weights $\scond(\act\mid\state)\fcond(\act'\mid\state)$; it is a function of policies on $\mathcal A_{+}$. Let $\pol$ be any policy with full support on $\actset$ and define $\bar\pol\coloneqq\pol(\cdot\mid\state)/\pol(\mathcal A_{+}\mid\state)$ on $\mathcal A_{+}$. Passing from $\pol$ to $\bar\pol$ shifts the scores of all actions in $\mathcal A_{+}$ by the common constant $-\log\pol(\mathcal A_{+}\mid\state)$, in both losses, so the hinge arguments of pairs inside $\mathcal A_{+}$ are unchanged. Since $\scond(\mathcal A_{+}\mid\state)=1$, the success-conditioning term equals $-\lam\sum_{\act\in\mathcal A_{+}}\scond(\act\mid\state)\log\bar\pol(\act)-\lam\log\pol(\mathcal A_{+}\mid\state)$; and the hinges with a failure action outside $\mathcal A_{+}$ are nonnegative, while there are none with a success action outside $\mathcal A_{+}$. Hence
\[
\mathcal L_{\state}(\pol)\ \ge\ \mathcal L_{\mathcal A_{+}}(\bar\pol)-\lam\log\pol(\mathcal A_{+}\mid\state)\ \ge\ \mathcal L_{\mathcal A_{+}}(\bar\pol).
\]
Conversely, extending a positive $\bar\pol$ on $\mathcal A_{+}$ by mass $1-1/n$ on $\bar\pol$ and $1/n$ spread over the complement leaves the internal hinges unchanged, changes the success-conditioning term by $-\lam\log(1-1/n)\to0$, and sends the external hinges to $0$, because the scores outside $\mathcal A_{+}$ tend to $-\infty$ relative to those inside; so $\inf\mathcal L_{\state}=\inf\mathcal L_{\mathcal A_{+}}$.\\
We parametrize the policies on $\mathcal A_{+}$ by scores. Every policy with full support on $\mathcal A_{+}$ can be written, for any fixed $t\in[0,1]$, as
\begin{equation}\tag{b}\label{eq:comono:ii}
\pol^{u}_{t}(\act\mid\state)\coloneqq\frac{\base(\act\mid\state)^{t}\,e^{u(\act)}}{\sum_{\act'\in\mathcal A_{+}}\base(\act'\mid\state)^{t}\,e^{u(\act')}},\quad\act\in\mathcal A_{+},
\end{equation}
for a score $u:\mathcal A_{+}\to\mathbb R$, unique up to an additive constant; explicitly $u=\log\pol^u_t(\cdot\mid\state)-t\log\base(\cdot\mid\state)+\mathrm{const}$, so for $t=0$ the score is the log-policy and for $t=1$ it is the log-likelihood ratio to the base, the two quantities inside the hinges of \cref{eq:loss:unw} and \cref{eq:loss:rw}. Minimizing $\mathcal L_{\mathcal A_{+}}$ over policies on $\mathcal A_{+}$ is therefore minimizing it over $u$, and below, in \cref{eq:comono:iv}, we write $F_t(u)$ for $\mathcal L_{\mathcal A_{+}}(\pol^u_t)$ up to an additive constant, with $t=0$ for the unweighted and $t=1$ for the reweighted loss.\\
We shall show after \cref{eq:comono:iv} that, since $\lam>0$, $F_t$ has a unique minimizer $u_t$ modulo constants, so that $\pol_t\coloneqq\pol^{u_t}_t$ is the unique minimizer of $\mathcal L_{\mathcal A_{+}}$, with full support on $\mathcal A_{+}$, and, in the proof of \cref{eq:comono:vi}, that
\begin{equation}\tag{$\ast$}\label{eq:comono:cert}
F_t(u)-F_t(u_t)\ \ge\ \lam\,\KL\bigl(\pol_t\,\|\,\pol^u_t\bigr)\quad\text{for every score }u,
\end{equation}
i.e.\ $\mathcal L_{\mathcal A_{+}}(\bar\pol)-\mathcal L_{\mathcal A_{+}}(\pol_t)\ge\lam\KL(\pol_t\|\bar\pol)$ for every policy $\bar\pol$ on $\mathcal A_{+}$ (for $\bar\pol$ without full support on $\mathcal A_{+}$ both sides are $+\infty$). Both facts use only that $\scond(\cdot\mid\state)>0$ on $\mathcal A_{+}$, so there is no circularity. Combined with the display above and $\KL(\pol_t\|\pol)=\KL(\pol_t\|\bar\pol)-\log\pol(\mathcal A_{+}\mid\state)$, which holds because $\pol_t$ is supported on $\mathcal A_{+}$, where $\pol(\cdot\mid\state)=\pol(\mathcal A_{+}\mid\state)\,\bar\pol$,
\[
\mathcal L_{\state}(\pol)-\inf\mathcal L_{\state}\ \ge\ \lam\,\KL(\pol_t\|\pol),
\]
so every minimizing sequence of $\mathcal L_\state$ converges to $\pol_t$ in total variation. Thus the minimizers of Section~\ref{sec:setup} are $\piunw(\cdot\mid\state)=\pol_0$ and $\pirw(\cdot\mid\state)=\pol_1$, extended by zero off $\mathcal A_{+}$: they exist, are unique, and have support exactly $\mathcal A_{+}$. Since $\Q$ vanishes off $\mathcal A_{+}$, $\val{\pol}(\state)=\sum_{\act\in\mathcal A_{+}}\pol(\act\mid\state)\Q(\state,\act)$ for both, so it suffices to compare the restricted minimizers; if $\mathcal A_{+}$ is a singleton both are the point mass on it and there is nothing to prove. From now on all sums run over $\mathcal A_{+}$, and $\actset$ in the displays below is to be read as $\mathcal A_{+}$.

Inside the hinge of \cref{eq:loss:unw} the policy enters through $\log\pol(\act^-\mid\state)-\log\pol(\act^+\mid\state)$, and inside the hinge of \cref{eq:loss:rw} through $\log\frac{\pol(\act^-\mid\state)}{\base(\act^-\mid\state)}-\log\frac{\pol(\act^+\mid\state)}{\base(\act^+\mid\state)}$; at $\pol=\pol^u_0$ the first difference equals $u(\act^-)-u(\act^+)$, and at $\pol=\pol^u_1$ so does the second, the normalizer of \cref{eq:comono:ii} cancelling. Hence both hinge terms equal
\begin{equation}\tag{c}\label{eq:comono:iii}
C(u)\coloneqq\sum_{\act,\act'\in\actset}\scond(\act\mid\state)\,\fcond(\act'\mid\state)\,\bigl[\marg+u(\act')-u(\act)\bigr]_+ ,
\end{equation}
while the supervised term at $\pol^u_t$ is
\begin{align*}
    -\lam\sum_{\act}\scond(\act\mid\state)\log\pol^u_t(\act\mid\state)
=&\lam\Bigl[\log\sum_{\act}\base(\act\mid\state)^{t}e^{u(\act)}-\sum_{\act}\scond(\act\mid\state)\,u(\act)\Bigr]\\
-&\lam t\sum_{\act}\scond(\act\mid\state)\log\base(\act\mid\state),
\end{align*}
whose last term does not depend on $u$. Up to additive constants, the objectives \cref{eq:loss:unw} and \cref{eq:loss:rw} at $\state$ are therefore the members $t=0$ and $t=1$ of the family
\begin{equation}\tag{iv}\label{eq:comono:iv}
F_t(u)\coloneqq C(u)+\lam\Bigl[\log\sum_{\act\in\actset}\base(\act\mid\state)^{t}e^{u(\act)}-\sum_{\act\in\actset}\scond(\act\mid\state)\,u(\act)\Bigr],\quad t\in[0,1],
\end{equation}
with $\piunw(\cdot\mid\state)=\pol^u_0$ for a minimizer $u$ of $F_0$ and $\pirw(\cdot\mid\state)=\pol^u_1$ for a minimizer $u$ of $F_1$. 

Every $F_t$ is unchanged when a constant is added to $u$, so we minimize over the scores with $\sum_\act u(\act)=0$. Two properties of $F_t$ on this set give the existence and the uniqueness of the minimizer. The first is that $F_t$ is \emph{coercive}: $F_t(u)\to+\infty$ as $\max_\act|u(\act)|\to\infty$. This is what turns the infimum into a minimum. A sublevel set $\{u:\sum_\act u(\act)=0,\ F_t(u)\le c\}$ of a coercive function is bounded, and that of a continuous function is closed, so it is compact; a continuous function attains its minimum on a nonempty compact set; and with $c=F_t(0)$ this minimum is the minimum over all zero-sum scores, every score outside the sublevel set having $F_t>c$. Coercivity of $F_t$ comes from an explicit bound. With $\bar u=\max_\act u(\act)$ and $\underline u=\min_\act u(\act)$, the pair formed by an action attaining $\underline u$ as the success and one attaining $\bar u$ as the failure gives $C(u)\ge\min_\act\scond(\act\mid\state)\,\min_\act\fcond(\act\mid\state)\,(\marg+\bar u-\underline u)$, and the bracket in \cref{eq:comono:iv} is at least $t\min_\act\log\base(\act\mid\state)+\min_\act\scond(\act\mid\state)\,(\bar u-\underline u)$, because the logarithm of the sum is at least $t\log\base(\act\mid\state)+\bar u$ at an action attaining $\bar u$, while $\sum_\act\scond(\act\mid\state)u(\act)\le\bar u-\scond(\act\mid\state)(\bar u-\underline u)$ at an action attaining $\underline u$. Hence
\[
F_t(u)\ \ge\ \lam\,t\min_\act\log\base(\act\mid\state)+\lam\min_\act\scond(\act\mid\state)\,(\bar u-\underline u),
\]
and on the zero-sum scores $\bar u-\underline u\ge\max_\act|u(\act)|$, since $\underline u\le0\le\bar u$; as $\scond>0$ on $\mathcal A_{+}$ and $\lam>0$, the right side tends to $+\infty$ with $\max_\act|u(\act)|$. (The bound on $C$ alone is of this kind only when $\fcond>0$ on $\mathcal A_{+}$, which is not assumed here; it is what is used in the case $\lam=0$ below, where that holds.) The second property is strict convexity. $F_t$ is convex, and for $\lam>0$ strictly convex on the zero-sum scores, because the Hessian of the bracket at $u$ is the covariance matrix of $\pol^u_t$, positive definite on nonconstant vectors as $\pol^u_t$ has full support, and a nonzero zero-sum vector is not constant; two distinct minimizers would therefore give a smaller value at their midpoint. So for $\lam>0$ each $F_t$ has a unique minimizer $u_t$ modulo constants, $\pirw$ and $\piunw$ are well defined, and both have full support on $\mathcal A_{+}$.

For $\lam>0$ we move $t$ from $0$ to $1$ in \cref{eq:comono:iv} and follow the value of the minimizer. At the hinge the minimizer is not a differentiable function of $t$, so we smooth inside the proof and pass to the limit at the end. For $\varepsilon>0$ replace $[z]_+$ in \cref{eq:comono:iii} by $\varepsilon\log(1+e^{z/\varepsilon})$, which is convex, has derivative $\sigma(z/\varepsilon)$ with $\sigma(z)=(1+e^{-z})^{-1}$, and satisfies
\begin{equation}\tag{v}\label{eq:comono:v}
0<\varepsilon\log(1+e^{z/\varepsilon})-[z]_+=\varepsilon\log\bigl(1+e^{-|z|/\varepsilon}\bigr)\le\varepsilon\log2 .
\end{equation}
Write $C_\varepsilon$ and $F_{t,\varepsilon}$ for $C$ and $F_t$ after this replacement. Since $C_\varepsilon\ge C$, the argument above applies verbatim: $F_{t,\varepsilon}$ is smooth, strictly convex and coercive on $\{\sum_\act u(\act)=0\}$, with a unique minimizer $u_{t,\varepsilon}$ there; let $\pol_{t,\varepsilon}\coloneqq\pol^{u_{t,\varepsilon}}_t$, a policy with full support. Two facts finish the proof. First, for $t\in\{0,1\}$, with $\pol_0\coloneqq\piunw(\cdot\mid\state)$ and $\pol_1\coloneqq\pirw(\cdot\mid\state)$,
\begin{equation}\tag{vi}\label{eq:comono:vi}
\sum_{\act\in\actset}\pol_t(\act)\log\frac{\pol_t(\act)}{\pol_{t,\varepsilon}(\act\mid\state)}\ \le\ \frac{\varepsilon\log2}{\lam}.
\end{equation}
Second, for every fixed $\varepsilon>0$ the map $t\mapsto\pol_{t,\varepsilon}$ is continuously differentiable and
\begin{equation}\tag{vii}\label{eq:comono:vii}
\frac{d}{dt}\val{\pol_{t,\varepsilon}}(\state)=\sum_{\{\act,\act'\}}\omega_{\act\act'}(t)\,\bigl(\Q(\state,\act)-\Q(\state,\act')\bigr)\bigl(\log\base(\act\mid\state)-\log\base(\act'\mid\state)\bigr),\quad\omega_{\act\act'}(t)\ge0,
\end{equation}
the sum running over unordered pairs of distinct actions. Since $\log$ is increasing, \cref{eq:comono:i} makes every term of \cref{eq:comono:vii} nonnegative, and integrating over $t\in[0,1]$ gives $\val{\pol_{1,\varepsilon}}(\state)\ge\val{\pol_{0,\varepsilon}}(\state)$ for every $\varepsilon>0$. By \cref{eq:comono:vi} and Pinsker's inequality, $\pol_{t,\varepsilon}\to\pol_t$ as $\varepsilon\downarrow0$ for $t\in\{0,1\}$, so
\[
\val{\pirw}(\state)-\val{\piunw}(\state)=\lim_{\varepsilon\downarrow0}\bigl(\val{\pol_{1,\varepsilon}}(\state)-\val{\pol_{0,\varepsilon}}(\state)\bigr)\ \ge\ 0 .
\]

For \cref{eq:comono:vi}, fix $t\in\{0,1\}$ and let $u_t$ be the minimizer of $F_t$ on $\{\sum_\act u(\act)=0\}$, so that $\pol_t=\pol^{u_t}_t$. Since \cref{eq:comono:v} gives $0\le F_{t,\varepsilon}-F_t\le\varepsilon\log2$ pointwise (the pair weights $\scond(\act\mid\state)\fcond(\act'\mid\state)$ sum to at most one),
\[
F_t(u_{t,\varepsilon})-F_t(u_t)\ \le\ F_{t,\varepsilon}(u_{t,\varepsilon})-F_t(u_t)\ \le\ F_{t,\varepsilon}(u_t)-F_t(u_t)\ \le\ \varepsilon\log2 .
\]
In the other direction, $C$ is convex and finite and the bracket in \cref{eq:comono:iv} is differentiable with gradient $\pol^u_t-\scond(\cdot\mid\state)$, so minimality of $u_t$ provides a subgradient $g$ of $C$ at $u_t$ with $g+\lam\bigl(\pol_t-\scond(\cdot\mid\state)\bigr)=0$. For every score $u$, convexity gives $C(u)\ge C(u_t)+\sum_\act g(\act)\bigl(u(\act)-u_t(\act)\bigr)$, and the log-partition function satisfies the exact identity
\[
\log\sum_{\act}\base(\act\mid\state)^{t}e^{u(\act)}-\log\sum_{\act}\base(\act\mid\state)^{t}e^{u_t(\act)}-\sum_\act\pol_t(\act)\bigl(u(\act)-u_t(\act)\bigr)=\sum_\act\pol_t(\act)\log\frac{\pol_t(\act)}{\pol^u_t(\act\mid\state)},
\]
as one sees by writing $\log\pol_t(\act)-\log\pol^u_t(\act\mid\state)=u_t(\act)-u(\act)+\log\sum_{\act'}\base(\act'\mid\state)^te^{u(\act')}-\log\sum_{\act'}\base(\act'\mid\state)^te^{u_t(\act')}$ and averaging under $\pol_t$. Adding the two, we get:
\begin{align*}
    F_t(u)-F_t(u_t)\ &\ge\ \sum_\act\bigl(g(\act)+\lam\pol_t(\act)-\lam\scond(\act\mid\state)\bigr)\bigl(u(\act)-u_t(\act)\bigr)+\lam\sum_\act\pol_t(\act)\log\frac{\pol_t(\act)}{\pol^u_t(\act\mid\state)}\\
  &=\lam\sum_\act\pol_t(\act)\log\frac{\pol_t(\act)}{\pol^u_t(\act\mid\state)} 
\end{align*}
and $u=u_{t,\varepsilon}$ gives \cref{eq:comono:vi}.

For \cref{eq:comono:vii}, fix $\varepsilon>0$ and let $t$ range over an open interval containing $[0,1]$. The minimizer is characterized by the stationarity condition
\begin{equation}\tag{viii}\label{eq:comono:viii}
\nabla C_\varepsilon(u_{t,\varepsilon})+\lam\bigl(\pol_{t,\varepsilon}(\cdot\mid\state)-\scond(\cdot\mid\state)\bigr)=0 ,
\end{equation}
a smooth equation in $(u,t)$ between zero-sum vectors whose derivative in $u$ is $K+\lam\bigl(\operatorname{diag}(\pol_{t,\varepsilon}(\cdot\mid\state))-\pol_{t,\varepsilon}(\cdot\mid\state)\,\pol_{t,\varepsilon}(\cdot\mid\state)^{\top}\bigr)$, with $K\coloneqq\nabla^2C_\varepsilon(u_{t,\varepsilon})$. Differentiating \cref{eq:comono:iii} twice and grouping the two orientations of each pair shows that $K$ is the graph Laplacian given by:
\begin{align*}
    (Kv)(\act)&=\sum_{\act'\ne\act}k_{\act\act'}\bigl(v(\act)-v(\act')\bigr),\\
k_{\act\act'}&=\frac{1}{\varepsilon}\Bigl[\scond(\act\mid\state)\fcond(\act'\mid\state)\,\sigma'\Bigl(\tfrac{\marg+u_{t,\varepsilon}(\act')-u_{t,\varepsilon}(\act)}{\varepsilon}\Bigr)+\scond(\act'\mid\state)\fcond(\act\mid\state)\,\sigma'\Bigl(\tfrac{\marg+u_{t,\varepsilon}(\act)-u_{t,\varepsilon}(\act')}{\varepsilon}\Bigr)\Bigr]\\
       &\ge0 ,
\end{align*}
so $K$ is symmetric, $K\mathbf 1=0$ and $v^\top Kv=\sum_{\{\act,\act'\}}k_{\act\act'}\bigl(v(\act)-v(\act')\bigr)^2\ge0$; the second summand of the derivative is the covariance matrix of $\pol_{t,\varepsilon}(\cdot\mid\state)$, positive definite on nonzero zero-sum vectors. By the implicit function theorem, $t\mapsto u_{t,\varepsilon}$ and hence $t\mapsto\pol_{t,\varepsilon}$ are continuously differentiable; write $\dot u$ and $\dot{\pol}$ for their derivatives at a fixed $t$. Since $\pol_{t,\varepsilon}(\cdot\mid\state)$ is the softmax of $u_{t,\varepsilon}+t\log\base(\cdot\mid\state)$, the chain rule gives $\dot{\pol}=\bigl(\operatorname{diag}(\pol_{t,\varepsilon})-\pol_{t,\varepsilon}\pol_{t,\varepsilon}^\top\bigr)\bigl(\dot u+\log\base(\cdot\mid\state)\bigr)$, that is, $\dot{\pol}=\operatorname{diag}(\pol_{t,\varepsilon}(\cdot\mid\state))\,y$ for the centred vector
\[
y\coloneqq\dot u+\log\base(\cdot\mid\state)-\Bigl(\sum_\act\pol_{t,\varepsilon}(\act\mid\state)\bigl(\dot u(\act)+\log\base(\act\mid\state)\bigr)\Bigr)\mathbf 1 .
\]
Differentiating \cref{eq:comono:viii} in $t$ gives $K\dot u+\lam\dot{\pol}=0$, and $K\dot u=K\bigl(y-\log\base(\cdot\mid\state)\bigr)$ because $K\mathbf 1=0$; so
\[
M y=K\log\base(\cdot\mid\state),\quad M\coloneqq K+\lam\operatorname{diag}\bigl(\pol_{t,\varepsilon}(\cdot\mid\state)\bigr).
\]
$M$ is symmetric positive definite, as $v^\top Mv=v^\top Kv+\lam\sum_\act\pol_{t,\varepsilon}(\act\mid\state)\,v(\act)^2>0$ for $v\ne0$; let $G\coloneqq M^{-1}$. Then $y=GK\log\base(\cdot\mid\state)$ and
\begin{align*}
    &\dot{\pol}=\operatorname{diag}(\pol_{t,\varepsilon})\,G\,\bigl(M-\lam\operatorname{diag}(\pol_{t,\varepsilon})\bigr)\log\base(\cdot\mid\state)=\Omega\,\log\base(\cdot\mid\state),\\
 &\Omega\coloneqq\operatorname{diag}(\pol_{t,\varepsilon})-\lam\operatorname{diag}(\pol_{t,\varepsilon})\,G\operatorname{diag}(\pol_{t,\varepsilon}).
\end{align*}
Three properties of $\Omega$ give \cref{eq:comono:vii}. It is symmetric. Its rows sum to zero: $M\mathbf 1=\lam\,\pol_{t,\varepsilon}(\cdot\mid\state)$ because $K\mathbf 1=0$, hence $G\,\pol_{t,\varepsilon}(\cdot\mid\state)=\mathbf 1/\lam$ and $\Omega\mathbf 1=\pol_{t,\varepsilon}(\cdot\mid\state)-\operatorname{diag}(\pol_{t,\varepsilon})\mathbf 1=0$. And its off-diagonal entries $\Omega_{\act\act'}=-\lam\,\pol_{t,\varepsilon}(\act\mid\state)\,\pol_{t,\varepsilon}(\act'\mid\state)\,G_{\act\act'}$ are nonpositive, because $G$ is entrywise nonnegative: if $x$ is a column of $G$, then $Mx$ is a unit vector, and if $x$ had a negative minimum at a coordinate $\act$, then $(Mx)(\act)=\lam\pol_{t,\varepsilon}(\act\mid\state)\,x(\act)+\sum_{\act'\ne\act}k_{\act\act'}\bigl(x(\act)-x(\act')\bigr)<0$, which is impossible. For a symmetric matrix with zero row sums, $f^\top\Omega g=-\sum_{\{\act,\act'\}}\Omega_{\act\act'}\bigl(f(\act)-f(\act')\bigr)\bigl(g(\act)-g(\act')\bigr)$ for all vectors $f,g$, so
\[
\frac{d}{dt}\val{\pol_{t,\varepsilon}}(\state)=\sum_\act\Q(\state,\act)\,\dot{\pol}(\act)=\Q(\state,\cdot)^\top\,\Omega\,\log\base(\cdot\mid\state)
\]
is \cref{eq:comono:vii} with $\omega_{\act\act'}(t)=\lam\,\pol_{t,\varepsilon}(\act\mid\state)\,\pol_{t,\varepsilon}(\act'\mid\state)\,G_{\act\act'}\ge0$.

\textbf{The case of $\lam=0$.} Assume $0<\Q(\state,\act)<1$ for every $\act$. Then $\scond(\cdot\mid\state)$ and $\fcond(\cdot\mid\state)$ have full support, $\mathcal A_{+}=\actset$, and the reduction above is not needed. A policy with zero mass at some action has infinite loss, because the pair with that action as the success and any action of positive mass as the failure has an infinite hinge and positive weight; so both minimizations run over policies with full support, which are the $\pol^u_t$ of \cref{eq:comono:ii}. The identities \cref{eq:comono:ii}--\cref{eq:comono:iv} are algebraic and do not involve $\lam$, so in the score $u$ both losses equal the common hinge objective $C$ of \cref{eq:comono:iii} up to additive constants, the \unweighted loss being read out as $\pol^u_0$ and the \reweighted loss as $\pol^u_1$. $C$ is convex, and coercive on $\{\sum_\act u(\act)=0\}$ by the bound $C(u)\ge\min_\act\scond(\act\mid\state)\,\min_\act\fcond(\act\mid\state)\,(\marg+\bar u-\underline u)$ noted after \cref{eq:comono:iv}, whose constant is now positive; hence its minimizers modulo constants form a nonempty compact convex set $U^*$, the \unweighted minimizers are exactly the $\pol^u_0$ and the \reweighted minimizers exactly the $\pol^u_1$ with $u\in U^*$. Fix $u\in U^*$ and write $p\coloneqq\pol^u_0$. By \cref{eq:comono:ii}, $\pol^u_1(\act\mid\state)=p(\act)\base(\act\mid\state)\big/\sum_{\act'}p(\act')\base(\act'\mid\state)$, so
\begin{align*}
\val{\pol^u_1}(\state)-\val{\pol^u_0}(\state)
&=\frac{\sum_{\act}p(\act)\base(\act\mid\state)\Q(\state,\act)-\bigl(\sum_{\act}p(\act)\Q(\state,\act)\bigr)\bigl(\sum_{\act}p(\act)\base(\act\mid\state)\bigr)}{\sum_{\act}p(\act)\base(\act\mid\state)}\\
&=\frac{\tfrac12\sum_{\act,\act'}p(\act)\,p(\act')\bigl(\Q(\state,\act)-\Q(\state,\act')\bigr)\bigl(\base(\act\mid\state)-\base(\act'\mid\state)\bigr)}{\sum_{\act}p(\act)\base(\act\mid\state)}\ \ge\ 0,
\end{align*}
the second equality by Lemma~\ref{lem:cov} and the inequality by \cref{eq:comono:i}; in the \noncomon\ case every sign reverses. Taking the supremum and the infimum over $u\in U^*$ on both sides gives the result in this case.
\end{proof}

\newpage

\subsection{MDP Counterexamples}
\label{app:proofs:mdpcounter}
\lemmdpsft*
\begin{proof}
Fix $\varepsilon\in(0,1/5)$.  The MDP has decision states $u,v,y$ and two absorbing
states, success and failure; an entry $\mathrm{Bernoulli}(q)$ means immediate absorption
in success with probability $q$ and in failure otherwise.
\[
\begin{array}{c|c|c|c}
\text{State} & \text{Action }1 & \text{Action }2 & \base(1\mid\state)\\ \hline
u & \mathrm{Bernoulli}(1-\varepsilon) & \text{move to }y & 1/2\\
v & \mathrm{Bernoulli}(1-3\varepsilon) & \mathrm{Bernoulli}(\varepsilon/2) & 1/2\\
y & \text{success} & \text{failure} & \varepsilon
\end{array}
\]
Every episode contains at most two decisions.  The action values of $\base$ and its
state values are
\begin{align*}
\Q(y,\cdot)&=(1,0),\ \val{\base}(y)=\varepsilon;\quad
\Q(u,\cdot)=(1-\varepsilon,\varepsilon),\ \val{\base}(u)=\tfrac12;\quad\\
\Q(v,\cdot)&=(1-3\varepsilon,\varepsilon/2),\ \val{\base}(v)=\tfrac{2-5\varepsilon}{4},    
\end{align*}

where $\Q(u,2)=\val{\base}(y)$ because action $2$ moves to $y$ and $\base$ is followed
from there.  Every decision state has base value in $(0,1)$ and $\base$ has full
support.  The initial distribution is
\[
\mu(u)=\frac{2-7\varepsilon}{2-5\varepsilon},\quad
\mu(v)=\frac{2\varepsilon}{2-5\varepsilon},\quad
\mu(y)=0,
\quad\text{so that}\quad
\mu(u)\,\varepsilon=\mu(v)\Bigl(1-\frac{7\varepsilon}{2}\Bigr).
\]

First we analyze the optimizers at $u$ and $v$.  For $s\in\{u,v\}$ write
$\alpha_s=\scond(1\mid s)$, $\beta_s=\fcond(1\mid s)$, $p_s=\pimix(1\mid s)$.  From the
values above,
\[
\alpha_u=1-\varepsilon,\quad \beta_u=\varepsilon,\quad
\alpha_v=1-\frac{\varepsilon}{2-5\varepsilon},\quad
\beta_v=\frac{6\varepsilon}{2+5\varepsilon},
\]
and, since $1<2-5\varepsilon<2$ and $2+5\varepsilon<3$, we have
$\varepsilon/2<1-\alpha_v<\varepsilon$ and $\beta_v>2\varepsilon$, hence
\begin{equation}\label{eq:mdp-ineq}
\alpha_v>\alpha_u,\quad
(1-\alpha_v)\beta_v>\varepsilon^2=(1-\alpha_u)\beta_u,\quad
\alpha_v-\beta_v<1-\tfrac{5\varepsilon}{2}<1-2\varepsilon=\alpha_u-\beta_u .
\end{equation}
The base is uniform at $u$ and $v$, so the reweighted and unweighted scores there
differ by the constant $\ln2$ and both losses have the same optimizer.  As a function
of the log-odds of action $1$, the state loss is convex, and its derivative is
$\lam\bigl(\pimix(1\mid s)-\alpha_s\bigr)$ plus the hinge contribution
\[
-\alpha_s(1-\beta_s)\ \text{on }(-\infty,-\marg),\quad
\beta_s-\alpha_s\ \text{on }(-\marg,\marg),\quad
(1-\alpha_s)\beta_s\ \text{on }(\marg,\infty).
\]
For negative log-odds, $\pimix(1\mid s)\le\tfrac12<\alpha_s$ and $\beta_s<\alpha_s$
make this derivative negative, so the minimizer has non-negative log-odds.
On $(0,\marg)$,
at the kink $\marg$, and on $(\marg,\infty)$ the stationarity conditions are
$p_s-\alpha_s=(\alpha_s-\beta_s)/\lam$;
$p_s=\sigma(\marg)$ with $-(1-\alpha_s)\beta_s/\lam\le\sigma(\marg)-\alpha_s\le(\alpha_s-\beta_s)/\lam$;
and $p_s-\alpha_s=-(1-\alpha_s)\beta_s/\lam$.  Exactly one of them holds, by
convexity, and therefore
\begin{equation}\label{eq:mdp-median}
p_s-\alpha_s=\operatorname{median}\Bigl\{-\frac{(1-\alpha_s)\beta_s}{\lam},\;
\sigma(\marg)-\alpha_s,\;\frac{\alpha_s-\beta_s}{\lam}\Bigr\},
\end{equation}
where for $\lam=0$ the outer entries are read as $\mp\infty$: the pure hinge is
minimized exactly at log-odds $\marg$, because
$\alpha_s(1-\beta_s)>(1-\alpha_s)\beta_s$.  By \cref{eq:mdp-ineq}, each entry of
\cref{eq:mdp-median} at $u$ strictly exceeds the corresponding entry at $v$, and the
median strictly increases under a strict componentwise increase; hence
\begin{equation}\label{eq:mdp-gap}
(p_u-\alpha_u)-(p_v-\alpha_v)>0\quad\text{for every }\lam,\marg\ge0 .
\end{equation}

Now we compute the value function.  At $y$, $\scond(1\mid y)=\varepsilon\cdot1/\varepsilon=1$, so
$\val{\scond}(y)=1\ge\val{\pimix}(y)$ for whatever policy $\pimix$ uses at $y$.  Some
choice is needed there: for $\lam>0$ the loss at $y$ has infimum $0$, approached only
as $\pimix(1\mid y)\to1$, and for $\lam=0$ its minimizers are not unique; the argument
uses only $\val{\pimix}(y)\le1$.  The one-step Bellman equations give
\begin{align*}
\val{\scond}(u)-\val{\pimix}(u)
&=(\alpha_u-p_u)(1-\varepsilon)+(1-\alpha_u)-(1-p_u)\val{\pimix}(y)\\
&=\varepsilon\,(p_u-\alpha_u)+(1-p_u)\bigl(1-\val{\pimix}(y)\bigr),\\
\implies\val{\scond}(v)-\val{\pimix}(v)
&=(\alpha_v-p_v)\Bigl(1-\frac{7\varepsilon}{2}\Bigr).
\end{align*}
Weighting by $\mu$ and using $\mu(u)\varepsilon=\mu(v)(1-7\varepsilon/2)$,
\[
\val{\scond}(\mu)-\val{\pimix}(\mu)
=\mu(u)\Bigl[\varepsilon\bigl((p_u-\alpha_u)-(p_v-\alpha_v)\bigr)
+(1-p_u)\bigl(1-\val{\pimix}(y)\bigr)\Bigr]>0
\]
by \cref{eq:mdp-gap} and $\val{\pimix}(y)\le1$.  This holds for both losses,
every finite $\lam,\marg\ge0$, and every admissible choice of the policy at $y$.
\end{proof}

\lemmdpcomono*

\begin{proof}
Fix scalars $0<q<1$, $\tfrac12<c<1$, $\tfrac12<t<1$, and $r$ with
$c+(1-c)q<r<\frac{c+(1-c)q^2}{c+(1-c)q}$.
The interval is nonempty, the difference of its endpoints being
$c(1-c)(1-q)^2/(c+(1-c)q)>0$, and $r<1$ because $q^2<q$.  The MDP has two decision
states and two absorbing states, success and failure; 
\[
\begin{array}{c|c|c|c}
\text{State} & \text{Action }1 & \text{Action }2 & \base(1\mid\state)\\ \hline
\state_0 & \mathrm{Bernoulli}(r) & \text{move to }\state_1 & t\\
\state_1 & \text{success} & \mathrm{Bernoulli}(q) & c
\end{array}
\]

Thus, we see: the action values of the base policy $\base$ and its state values are
\begin{align*}
&\Q(\state_1,\cdot)=(1,\,q),\quad
\Q(\state_0,\cdot)=\bigl(r,\;c+(1-c)q\bigr),\\
&\val{\base}(\state_1)=c+(1-c)q,\quad
\val{\base}(\state_0)=t\,r+(1-t)\bigl(c+(1-c)q\bigr),    
\end{align*}

where $\Q(\state_0,2)=\val{\base}(\state_1)$ because action $2$ moves to $\state_1$ and
$\base$ is followed from there.  At both states action $1$ has the strictly larger
action value, $\Q(\state_0,1)=r>c+(1-c)q=\Q(\state_0,2)$ and
$\Q(\state_1,1)=1>q=\Q(\state_1,2)$, and, since $\base(1\mid\state_0)=t>\tfrac12$ and
$\base(1\mid\state_1)=c>\tfrac12$, also the strictly larger base probability: $\base$
is strictly aligned with $\Q(\state,\cdot)$ at both states.

The construction works as follows.  Under the base critic, action $1$ is the better
action at $\state_0$; but the SFT value of $\state_1$ is given by:
\[
\val{\scond}(\state_1)=\frac{c+(1-c)q^2}{c+(1-c)q}>r.
\]
This exceeds $r$, so once the policy at $\state_1$ is SFT, action $2$ at $\state_0$ is the
better action.  We show that both $\pirw$ and $\piunw$ learn the same score gap at $\state_0$,
and that the reweighted readout, tilted by the aligned base, converts that gap into
more mass on action $1$.

We choose
\[
0<\marg<\log\frac1q,
\quad
0<\lam<\min\Bigl\{\scond(1\mid\state_0)-\fcond(1\mid\state_0),\;
\scond(2\mid\state_0)\,\fcond(1\mid\state_0)\Bigr\}.
\]
The first entry is positive because $r>\val{\base}(\state_0)$ (action $1$ beats the
other action's quality), which gives
$\scond(1\mid\state_0)=tr/\val{\base}(\state_0)>t>t(1-r)/(1-\val{\base}(\state_0))=\fcond(1\mid\state_0)$ (the second is positive because $r$ and $c+(1-c)q$ lie in $(0,1)$).

Since $\Qb(\state_1,1)=1$, the failure-conditioned policy
is $\fcond(\cdot\mid\state_1)=(0,1)$, and the contrastive term reduces to
\[
\scond(1\mid\state_1)\,[\marg+u_2-u_1]_+\;+\;\scond(2\mid\state_1)\,\marg,
\]
which is minimal exactly when $u_1-u_2\ge\marg$.  The SFT policy meets this for both $\lrw$ and $\lunw$: its \unweighted gap is
$\log\frac{\scond(1\mid\state_1)}{\scond(2\mid\state_1)}=\log\frac{c}{(1-c)q}>\log\frac1q>\marg$
(as $c>1-c$), and its \reweighted gap is
$\log\frac{\scond(1\mid\state_1)/c}{\scond(2\mid\state_1)/(1-c)}=\log\frac1q>\marg$.
Since the SFT term is uniquely minimized at $\scond$, both combined optimizers at
$\state_1$ equal $\scond(\cdot\mid\state_1)$; they are unique, have full support, and
\[
\val{\pirw}(\state_1)=\val{\piunw}(\state_1)=\val{\scond}(\state_1).
\]

Write $g$ for the score gap between action $1$ and action $2$ at $\state_0$: $g=\ln\pol(1\mid\state_0)-\ln\pol(2\mid\state_0)$ for the \unweighted\ loss and $g=\ln\frac{\pol(1\mid\state_0)}{\base(1\mid\state_0)}-\ln\frac{\pol(2\mid\state_0)}{\base(2\mid\state_0)}$ for the \reweighted\ loss. In terms of $g$, the contrastive term of either loss at $\state_0$ is:
\[
\scond(1\mid\state_0)\fcond(2\mid\state_0)\,[\marg-g]_+
+\scond(2\mid\state_0)\fcond(1\mid\state_0)\,[\marg+g]_+
+\text{const},
\]
whose left and right derivatives at $g=\marg$ are
$-\bigl(\scond(1\mid\state_0)-\fcond(1\mid\state_0)\bigr)$ and
$\scond(2\mid\state_0)\fcond(1\mid\state_0)$.  The SFT term is strictly convex in
$g$ with derivative $\lam\bigl(\pi(1\mid\state_0)-\scond(1\mid\state_0)\bigr)$, of
absolute value less than $\lam$.  By the choice of $\lam$, the total left derivative
at $\marg$ is negative and the total right derivative positive, so the unique
minimizer of the (strictly convex) objective is $g=\marg$, for both losses.  In
probabilities,
\[
\piunw(1\mid\state_0)=\frac{e^{\marg}}{1+e^{\marg}},
\quad
\pirw(1\mid\state_0)=\frac{t\,e^{\marg}}{1-t+t\,e^{\marg}},
\]
and $\pirw(1\mid\state_0)>\piunw(1\mid\state_0)$ because $t>\tfrac12$.  Since both
policies coincide at $\state_1$,
\[
\val{\piunw}(\state_0)-\val{\pirw}(\state_0)
=\bigl(\pirw(1\mid\state_0)-\piunw(1\mid\state_0)\bigr)
\bigl(\val{\scond}(\state_1)-r\bigr)>0 .
\]
Both optimizers are unique and attained at finite scores, and the conclusion holds
for every $\marg\in(0,\log\tfrac1q)$ and every $\lam$ below the displayed bound.
\end{proof}

\newpage

\subsection{The combined loss is a trust-region step}
\label{app:proofs:trpo}

Trust-region policy optimization \citep{schulman2017trpo} maximizes a linear
surrogate of the return under a divergence constraint to the current policy; in
its practical form the constraint is the quadratic approximation, which for a
tabular policy at a state is the $\chi^2$-divergence
$\chi^2(p\Vert\base)=\sum_\act\bigl(p(\act)-\base(\act\mid\state)\bigr)^2/\base(\act\mid\state)$.
\citet{russo2026success} observed that SFT is exactly such a step: at every state,
$\scond$ maximizes the bandit reward over a $\chi^2$-ball around $\base$.  The
point of this section is that adding the contrastive term does not change the
nature of the step, only its size.  Once the SFT weight $\lam$ and the margin $\marg$ are large enough for
the hinges to be in their linear regime, both combined losses, reweighted and
unweighted alike, produce the same policy, and that policy is again the
maximizer of the bandit reward over a $\chi^2$-ball around $\base$, now with a
radius that shrinks in $\lam$ toward the SFT radius.  Contrastive training, in this
regime, is SFT with a wider trust region.

\begin{restatable}[The combined policy is a Pearson trust-region step]
{theorem}{thmtrpo}
\label{thm:trpo}
Consider a contextual bandit with finitely many actions, a full-support base
policy $\base$, and a non-degenerate state $\state$ at which $\Q(\state,\cdot)$ is
not constant.  There are thresholds
$\lam_*,\marg_*>0$, depending only on $\base$ and $\Q(\state,\cdot)$, with the
following property.  For every $\lam\ge\lam_*$ and $\marg\ge\marg_*$, there exists $R_\lam > 0$ such that
the optimal solutions of the \reweighted and the \unweighted hinge losses solve the following TRPO:
\[
\max_{p\in\Delta_N}\ \sum_{\act}p(\act)\,\Q(\state,\act)
\quad\text{subject to}\quad
\chi^2(p\Vert\base)\le R_\lam^2.
\]

The radius never increases with $\lam$ and $R_\lam^2$ converges to the
SFT radius $\chi^2(\scond\Vert\base)$ as $\lam\to\infty$.
\end{restatable}

\begin{proof}

By Lemma~\ref{lem:uniform-margin}, for $\lam\ge\bar\lambda(\state)$ the policy $\pol^\lam$ of Lemma~\ref{lem:hinge-optimizer} is $\pirw(\cdot\mid\state)$ when $\marg\ge\bar\delta_{\rw}(\state)$ and $\piunw(\cdot\mid\state)$ when $\marg\ge\bar\delta_{\unw}(\state)$, uniquely if $\mathcal A_{+}$ is the whole action set and in the limiting sense of that lemma otherwise.

The rest is a computation with the explicit policy \cref{eq:explicit-policy}.  Summing the definition of
$\fcond$ over $\mathcal A_{+}$ gives
$\fcond(\mathcal A_{+}\mid\state)=\bigl(\base(\mathcal A_{+}\mid\state)-\val{\base}\bigr)/(1-\val{\base})$, and
then, for $\act\in \mathcal A_{+}$,
\begin{align*}
\frac{\fcond(\mathcal A_{+}\mid\state)\scond(\act\mid\state)-\fcond(\act\mid\state)}{\base(\act\mid\state)}
&=\frac{\bigl(\base(\mathcal A_{+}\mid\state)-\val{\base}\bigr)\Q(\state,\act)/\val{\base}-\bigl(1-\Q(\state,\act)\bigr)}{1-\val{\base}}\\
&=\frac{\base(\mathcal A_{+}\mid\state)\Q(\state,\act)/\val{\base}-1}{1-\val{\base}},
\end{align*}
so the lemma's formula reads
\begin{equation}\label{eq:trpo:policy-ratio}
\frac{\pol^\lam(\act\mid\state)}{\base(\act\mid\state)}
=\frac{1+\frac{\base(\mathcal A_{+}\mid\state)}{\lam(1-\val{\base})}}{\val{\base}}\,\Q(\state,\act)
-\frac{1}{\lam(1-\val{\base})}
\quad(\act\in \mathcal A_{+}).
\end{equation}
It is convenient to center $\Q(\state,\cdot)$ at $\val{\base}/\base(\mathcal A_{+}\mid\state)$,
its mean under $\base$ restricted to $\mathcal A_{+}$.  Since
$\frac{1+\frac{\base(\mathcal A_{+}\mid\state)}{\lam(1-\val{\base})}}{\val{\base}}\cdot
\frac{\val{\base}}{\base(\mathcal A_{+}\mid\state)}=\frac{1}{\base(\mathcal A_{+}\mid\state)}+\frac{1}{\lam(1-\val{\base})}$,
subtracting $1$ from~\cref{eq:trpo:policy-ratio} gives
\begin{equation}\label{eq:trpo:centered-ratio}
\frac{\pol^\lam(\act\mid\state)}{\base(\act\mid\state)}-1
=\frac{1+\frac{\base(\mathcal A_{+}\mid\state)}{\lam(1-\val{\base})}}{\val{\base}}
\Bigl(\Q(\state,\act)-\frac{\val{\base}}{\base(\mathcal A_{+}\mid\state)}\Bigr)
+\frac{1-\base(\mathcal A_{+}\mid\state)}{\base(\mathcal A_{+}\mid\state)}
\quad(\act\in \mathcal A_{+}),
\end{equation}
while off $\mathcal A_{+}$, where $\pol^\lam$ vanishes, the left side of \cref{eq:trpo:centered-ratio} is $-1$.

The radius now falls out.  Square \cref{eq:trpo:centered-ratio}, weight by $\base(\act\mid\state)$ and sum
over $\mathcal A_{+}$: the cross term is killed by
$\sum_{\act\in \mathcal A_{+}}\base(\act\mid\state)\bigl(\Q(\state,\act)-\val{\base}/\base(\mathcal A_{+}\mid\state)\bigr)=0$,
the constant term contributes $(1-\base(\mathcal A_{+}\mid\state))^2/\base(\mathcal A_{+}\mid\state)$, and the
actions off $\mathcal A_{+}$ contribute their base mass $1-\base(\mathcal A_{+}\mid\state)$; the last two add
up to $(1-\base(\mathcal A_{+}\mid\state))/\base(\mathcal A_{+}\mid\state)$. 

Thus, defining
\begin{equation}\label{eq:trpo:radius}
R_\lam^2\coloneqq\chi^2(\pol^\lam\Vert\base)
=\frac{1-\base(\mathcal A_{+}\mid\state)}{\base(\mathcal A_{+}\mid\state)}
+\left(\frac{1+\frac{\base(\mathcal A_{+}\mid\state)}{\lam(1-\val{\base})}}{\val{\base}}\right)^{2}
\sum_{\act\in\mathcal A_{+}}\base(\act\mid\state)\Bigl(\Q(\state,\act)-\frac{\val{\base}}{\base(\mathcal A_{+}\mid\state)}\Bigr)^{2},
\end{equation}
this is the radius of the statement.

For optimality no duality is needed: the divergence is a quadratic, so we expand it
around $\pol^\lam$.  For $p\in\Delta_N$, by \cref{eq:trpo:radius},
\[
\chi^2(p\Vert\base)-R_\lam^2
=\sum_{\act}\frac{\bigl(p(\act)-\pol^\lam(\act\mid\state)\bigr)^2}{\base(\act\mid\state)}
+2\sum_{\act}\bigl(p(\act)-\pol^\lam(\act\mid\state)\bigr)
\Bigl(\frac{\pol^\lam(\act\mid\state)}{\base(\act\mid\state)}-1\Bigr).
\]
In the second sum substitute \cref{eq:trpo:centered-ratio} on $\mathcal A_{+}$ and $-1$ off $\mathcal A_{+}$.  Both $p$ and $\pol^\lam$
have total mass one, so
$\sum_{\act\in \mathcal A_{+}}\bigl(p(\act)-\pol^\lam(\act\mid\state)\bigr)=-\sum_{\act\notin \mathcal A_{+}}p(\act)$,
and $\Q(\state,\act)=\pol^\lam(\act\mid\state)=0$ off $\mathcal A_{+}$.  The pieces not involving
$\Q$ therefore collect, with the common factor $-\sum_{\act\notin \mathcal A_{+}}p(\act)$, to
$-2\bigl(\frac{1}{\base(\mathcal A_{+}\mid\state)}+\frac{1}{\lam(1-\val{\base})}\bigr)
+\frac{2(1-\base(\mathcal A_{+}\mid\state))}{\base(\mathcal A_{+}\mid\state)}=-2-\frac{2}{\lam(1-\val{\base})}$;
together with the $-2\sum_{\act\notin \mathcal A_{+}}p(\act)$ coming from off $\mathcal A_{+}$ this leaves
\begin{equation}\label{eq:trpo:divergence-expansion}
\begin{split}
    &\chi^2(p\Vert\base)-R_\lam^2
=\sum_{\act}\frac{\bigl(p(\act)-\pol^\lam(\act\mid\state)\bigr)^2}{\base(\act\mid\state)}\\
+&2\,\frac{1+\frac{\base(\mathcal A_{+}\mid\state)}{\lam(1-\val{\base})}}{\val{\base}}
\sum_{\act}\bigl(p(\act)-\pol^\lam(\act\mid\state)\bigr)\Q(\state,\act)
+\frac{2}{\lam(1-\val{\base})}\sum_{\act\notin \mathcal A_{+}}p(\act).
\end{split}
\end{equation}
This is the whole argument.  If $p$ is feasible the left side is nonpositive; the
first and last terms on the right are nonnegative and the coefficient of the middle
one is positive, so
$\sum_{\act}p(\act)\Q(\state,\act)\le\sum_{\act}\pol^\lam(\act\mid\state)\Q(\state,\act)$,
and strictly so if $p\ne\pol^\lam$, because full support makes the first term
positive.  As $\pol^\lam$ is feasible by \cref{eq:trpo:radius}, it is the unique maximizer.  The last
term in \cref{eq:trpo:divergence-expansion} is where the constraint $p\in\Delta_N$ earns its keep: it is the
contribution of the nonnegativity constraints active off $\mathcal A_{+}$, and without them the
maximizer would carry negative mass there --- which is also why the radius contains
the extra $(1-\base(\mathcal A_{+}\mid\state))/\base(\mathcal A_{+}\mid\state)$.
It remains to look at the dependence on $\lam$.  In $R_\lam^2$ only the squared
coefficient $\bigl(1+\frac{\base(\mathcal A_{+}\mid\state)}{\lam(1-\val{\base})}\bigr)^2/\val{\base}^2$
depends on $\lam$, and it decreases strictly to $1/\val{\base}^2$.  So $R_\lam$ is
non-increasing, and it is strictly decreasing and strictly above its limit at every
finite $\lam$ exactly when the sum in $R_\lam^2$ is positive, i.e.\ when
$\Q(\state,\cdot)$ is not constant on $\mathcal A_{+}$ (the sum vanishes iff
$\Q(\state,\act)=\val{\base}/\base(\mathcal A_{+}\mid\state)$ for every $\act\in \mathcal A_{+}$).  Lastly, by~\cref{eq:trpo:policy-ratio}, $\pol^\lam(\act\mid\state)\to\base(\act\mid\state)\Q(\state,\act)/\val{\base}
=\scond(\act\mid\state)$ as $\lam\to\infty$ for every action, and \cref{eq:trpo:radius} together with
continuity of $\chi^2(\cdot\Vert\base)$ gives $R_\lam^2\to\chi^2(\scond\Vert\base)$.
\end{proof}

\newpage

\subsection{Retrieval Guarantees for Contrastive Learning}

\thmrankings*
\begin{proof}
Throughout, $\base$ has full support and $\val{\base}\in(0,1)$, scores are
log-probabilities for the unweighted loss and log-ratios to $\base$ for the
reweighted one, and a pair $(i,j)$ with $\Q(\state,\act_i)=\Q(\state,\act_j)$ is never
an inversion.  When some action has zero quality the losses are not minimized at
finite scores, and the contrastive policy is the limit of the policies induced by a
minimizing score sequence. The argument given there is written for the \reweighted loss and applies verbatim to the \unweighted loss, whose scores differ from the \reweighted ones by the fixed vector $\log\base(\cdot\mid\state)$. It shows that every action omitted by the limit has zero quality and zero
probability, and that the limit policy minimizes the loss restricted to its support.
An omitted action has the smallest possible quality and the smallest possible
probability, so it belongs to no inversion of the limit policy; it is therefore
enough to prove the containments for a policy that minimizes its loss at finite
scores, restricted if necessary to its support, and we assume this from now on.

We first record the fact used in the text: if $\Inv(\pol_1)\subseteq\Inv(\pol_2)$ and $\pol_1$ does not tie any two actions of distinct quality, then every list sorted by $\pol_1$ scores at least as well as every list sorted by $\pol_2$ under every swap-monotone metric, whatever the tie-breaking. Let $\ell_1$ and $\ell_2$ be two such lists. Take the first action of $\ell_1$ and bring it to the front of $\ell_2$ by adjacent swaps, then the second action of $\ell_1$ to the front of what remains, and so on until $\ell_2$ has become $\ell_1$. Every swap moves an action $x$ being placed above an action $y$ that preceded it in $\ell_2$ but follows it in $\ell_1$. If $x$ and $y$ have equal quality, the quality sequence along the list is unchanged, and so is every swap-monotone metric by $(a)$. If $x$ has the larger quality, the swap moves the higher-quality action up and no swap-monotone metric decreases by $(b)$. The third case does not occur: if $x$ had the smaller quality, then, since $x$ precedes $y$ in $\ell_1$ and $\pol_1$ does not tie actions of distinct quality, $\pol_1(x\mid\state)>\pol_1(y\mid\state)$; so $(y,x)$ is an inversion of $\pol_1$, hence of $\pol_2$, so $\pol_2(y\mid\state)<\pol_2(x\mid\state)$ and $x$ precedes $y$ in every list sorted by $\pol_2$, contradicting that $y$ preceded $x$ in $\ell_2$.

For item~1, the outer containment is one line.  If $(i,j)\in\Inv(\scond)$ then
$\Q(\state,\act_i)>\Q(\state,\act_j)$ and
$\base(\act_i\mid\state)\Q(\state,\act_i)<\base(\act_j\mid\state)\Q(\state,\act_j)$,
whence $\base(\act_i\mid\state)<\base(\act_j\mid\state)\Q(\state,\act_j)/\Q(\state,\act_i)
<\base(\act_j\mid\state)$, that is $(i,j)\in\Inv(\base)$.

Next, $\Inv(\piunw)\subseteq\Inv(\scond)$.  Suppose not: there are $i,j$ with
$\Q(\state,\act_i)>\Q(\state,\act_j)$ and
$\scond(\act_i\mid\state)\ge\scond(\act_j\mid\state)$ but
$\piunw(\act_i\mid\state)<\piunw(\act_j\mid\state)$, so the score of $\act_i$ is
smaller than that of $\act_j$.  Increase the score of $\act_i$ at rate
$1/\scond(\act_i\mid\state)$ and decrease that of $\act_j$ at rate
$1/\scond(\act_j\mid\state)$, keeping the other scores fixed; we show the right
directional derivative of the unweighted loss is negative, contradicting minimality.
Recall that a hinge has slope $1$ at positive arguments and $0$ at negative ones, so
its right slope at a smaller argument never exceeds its left slope at a larger one,
kinks included.

Consider the pairs in which some third action $\act$ is the success action.  The
hinge with failure action $\act_i$ has the smaller argument and it increases; the
one with failure action $\act_j$ has the larger argument and it decreases; after the
rates are absorbed, their weights are
$\scond(\act\mid\state)\fcond(\act_i\mid\state)/\scond(\act_i\mid\state)$ and
$\scond(\act\mid\state)\fcond(\act_j\mid\state)/\scond(\act_j\mid\state)$, and
\[
\frac{\fcond(\act_i\mid\state)}{\scond(\act_i\mid\state)}
=\frac{\val{\base}}{1-\val{\base}}\cdot\frac{1-\Q(\state,\act_i)}{\Q(\state,\act_i)}
<\frac{\val{\base}}{1-\val{\base}}\cdot\frac{1-\Q(\state,\act_j)}{\Q(\state,\act_j)}
=\frac{\fcond(\act_j\mid\state)}{\scond(\act_j\mid\state)} ,
\]
so the smaller weight multiplies the smaller slope and the two contributions add to
something nonpositive.  In the pairs where the third action is the failure action,
the hinge with success action $\act_i$ has the larger argument and it decreases,
the one with success action $\act_j$ has the smaller argument and it increases, and
after absorbing the rates both carry the same weight $\fcond(\act\mid\state)$; again
the sum is nonpositive.  Pairs not involving $\act_i$ or $\act_j$, and the two
self-pairs, are constant.  Of the two pairs formed by $\act_i$ and $\act_j$, the
hinge with success action $\act_i$ has argument $\marg$ plus the score gap, hence is
strictly active with slope $1$ and its argument falls at rate
$1/\scond(\act_i\mid\state)+1/\scond(\act_j\mid\state)$; the reverse hinge rises at
the same rate with slope at most $1$.  Together they contribute at most
\begin{align*}
    &\Bigl(\frac{1}{\scond(\act_i\mid\state)}+\frac{1}{\scond(\act_j\mid\state)}\Bigr)
\bigl(\scond(\act_j\mid\state)\fcond(\act_i\mid\state)
-\scond(\act_i\mid\state)\fcond(\act_j\mid\state)\bigr)\\
=&\bigl(\scond(\act_i\mid\state)+\scond(\act_j\mid\state)\bigr)
\Bigl(\frac{\fcond(\act_i\mid\state)}{\scond(\act_i\mid\state)}
-\frac{\fcond(\act_j\mid\state)}{\scond(\act_j\mid\state)}\Bigr)<0 .
\end{align*}
Finally the SFT term: differentiating
$-\lam\sum_{\act}\scond(\act\mid\state)\log\piunw(\act\mid\state)$ along the
perturbation, the terms linear in the scores cancel and the log-partition
contributes
\[
\lam\Bigl(\frac{\piunw(\act_i\mid\state)}{\scond(\act_i\mid\state)}
-\frac{\piunw(\act_j\mid\state)}{\scond(\act_j\mid\state)}\Bigr)
\le\lam\Bigl(\frac{\piunw(\act_i\mid\state)}{\scond(\act_j\mid\state)}
-\frac{\piunw(\act_j\mid\state)}{\scond(\act_j\mid\state)}\Bigr)<0 ,
\]
using $\scond(\act_i\mid\state)\ge\scond(\act_j\mid\state)$ and then
$\piunw(\act_i\mid\state)<\piunw(\act_j\mid\state)$.  The right directional
derivative is therefore strictly negative, which is impossible at a minimizer.

For the reweighted policy the same perturbation, with rates
$1/\base(\act_i\mid\state)$ and $1/\base(\act_j\mid\state)$, is exactly the
computation in the proof of Theorem~\ref{thm:rw-base}, which gives
$\pirw(\act_i\mid\state)/\base(\act_i\mid\state)\ge\pirw(\act_j\mid\state)/\base(\act_j\mid\state)$
whenever $\Q(\state,\act_i)>\Q(\state,\act_j)$, for every $\lam,\marg\ge0$.  If
$(i,j)\notin\Inv(\base)$, that is $\base(\act_i\mid\state)\ge\base(\act_j\mid\state)$,
multiplying the two inequalities gives
$\pirw(\act_i\mid\state)\ge\pirw(\act_j\mid\state)$, so $(i,j)\notin\Inv(\pirw)$.
This proves item~1.

For item~2 take $\lam^*\coloneqq\bar\lambda(\state)$ and $\marg^*\coloneqq\max\{\bar\delta_{\rw}(\state),\bar\delta_{\unw}(\state)\}$ from Lemma~\ref{lem:uniform-margin}. For $\lam\ge\lam^*$ and $\marg\ge\marg^*$ both contrastive policies equal the policy $\pol^\lam$ of Lemma~\ref{lem:hinge-optimizer} (in the limiting sense of that lemma when $\mathcal A_{+}$ is proper), which vanishes on zero-quality
actions and on the others equals
\[
\pol^\lam(\act\mid\state)
=\scond(\act\mid\state)
\Bigl(1+\frac1\lam\Bigl(\fcond(\mathcal A_{+}\mid\state)-\frac{\fcond(\act\mid\state)}{\scond(\act\mid\state)}\Bigr)\Bigr),
\]
with a positive bracket because $\lam\ge\lam^*$.  Let
$\Q(\state,\act_i)>\Q(\state,\act_j)$ and
$\scond(\act_i\mid\state)\ge\scond(\act_j\mid\state)$.  If $\act_j$ has zero quality
then $\pol^\lam(\act_j\mid\state)=0\le\pol^\lam(\act_i\mid\state)$.  Otherwise both
actions have positive quality, and since
$\fcond(\act\mid\state)/\scond(\act\mid\state)$ decreases with $\Q(\state,\act)$, the
bracket is strictly larger at $\act_i$ than at $\act_j$; multiplied by
$\scond(\act_i\mid\state)\ge\scond(\act_j\mid\state)$ this gives
$\pol^\lam(\act_i\mid\state)>\pol^\lam(\act_j\mid\state)$.  Either way
$(i,j)\notin\Inv(\pol^\lam)$, so $\Inv(\pirw)=\Inv(\pol^\lam)\subseteq\Inv(\scond)$,
and the same holds for $\piunw$.
\end{proof}

\newpage
\input{sections/softplus.tex}

\newpage
\input{sections/experimental_details.tex}

%% file: sections/softplus.tex
\section{The softplus objectives: DPO and the hinge limit}
\label{app:softplus}

The hinge in \cref{eq:loss:unw,eq:loss:rw} is the zero-temperature limit of the softplus $z\mapsto\temp\ln(1+e^{z/\temp})$, which we already used inside the proof of Theorem~\ref{thm:comono} via the bound \cref{eq:comono:v}. The hinge versions are what the generative retrieval community mostly uses, and generative retrieval remains one of the main instantiations of MDPs with a terminal trajectory level binary reward. This motivated us to focus our analysis on contrastive hinge losses. However, it is a natural question to ask if one may use contrastive losses that uses the softplus term at a positive temperature, instead of the hinge limit. 

It should be noted that the softplus losses are, up to a positive factor, the DPO objective of \citet{dpo2023} with a margin in the contrastive and an added success-conditioned likelihood term: the \reweighted\ loss uses the base policy as the DPO reference and the \unweighted\ loss uses the uniform reference. DPO with success conditioning has already been empirically explored in works like \cite{dpop}, however it remains to provide a theoretical analysis of the same, and explore what happens if the contrastive term in the DPO-based objective is also equipped with a margin to enforce further separation between the negatives and positives.

This section proves the four main results from the main text, namely Theorems~\ref{thm:rw-base}, \ref{thm:beat-sft}, \ref{thm:comono} and~\ref{thm:rankings}, for the softplus objectives at every positive temperature and shows that the hinge results are the correctly scaled zero-temperature limit (Appendix~\ref{app:softplus:limit}). Apart from the relation to DPO, we also provide a slightly different theoretical treatment while deriving the results in this section , where we utilize the fact that, while using the softplus version instead of hinge, the objectives become smooth, so stationarity conditions are available and the policies vary continuously in every parameter. 

\subsection{The softplus state losses and their DPO form}
\label{app:softplus:dpo}

Fix a non-degenerate state $\state$ and a temperature $\temp>0$. The \emph{softplus state losses} are the state losses of \cref{eq:loss:unw,eq:loss:rw} with the hinge $\hinge{z}$ replaced by $\temp\ln(1+e^{z/\temp})$:
\begin{equation}\label{eq:sp-loss}
\begin{split}
&\mathcal L^{\temp}_{\unw,\state}(\newpol;\base)\coloneqq-\lam\sum_{\act\in\actset}\scond(\act\mid\state)\ln\newpol(\act\mid\state)\\
&+\sum_{\act_+,\act_-\in\actset}\scond(\act_+\mid\state)\,\fcond(\act_-\mid\state)\;\temp\ln\left(1+\exp\left(\tfrac1\temp\left(\marg+\ln\newpol(\act_-\mid\state)-\ln\newpol(\act_+\mid\state)\right)\right)\right)\\
&\mathcal L^{\temp}_{\rw,\state}(\newpol;\base)\coloneqq-\lam\sum_{\act\in\actset}\scond(\act\mid\state)\ln\newpol(\act\mid\state)\\
&+\sum_{\act_+,\act_-\in\actset}\scond(\act_+\mid\state)\,\fcond(\act_-\mid\state)\;\temp\ln\left(1+\exp\left(\tfrac1\temp\left(\marg+\ln\frac{\newpol(\act_-\mid\state)}{\base(\act_-\mid\state)}-\ln\frac{\newpol(\act_+\mid\state)}{\base(\act_+\mid\state)}\right)\right)\right),
\end{split}
\end{equation}
Everything else is as in Section~\ref{sec:setup}: in an MDP the objective is the sum of the state losses weighted by $\occ^{+}$. Since Lemma~\ref{lem:statewise} applies to any per-step loss; $\piunw^{\temp}(\cdot\mid\state)$ and $\pirw^{\temp}(\cdot\mid\state)$ denote state-wise minimizers. We write $\sigma(z)=(1+e^{-z})^{-1}$, so that $\frac{d}{dz}\,\temp\ln(1+e^{z/\temp})=\sigma(z/\temp)$. By \cref{eq:comono:v}, and because the pair weights $\scond(\act_+\mid\state)\fcond(\act_-\mid\state)$ sum to one (because they come from the product measure of $\scond\times\fcond$), we get:
\begin{equation}\label{eq:sp-approx}
0\ \le\ \mathcal L^{\temp}_{\star,\state}(\newpol;\base)-\mathcal L_{\star,\state}(\newpol;\base)\ \le\ \temp\ln2\,\text{for every policy }\newpol,
\end{equation}
for $\star\in\{\mathrm{rw}, \mathrm{unw}\}$. So the softplus objective uniformly converges to its hinge limit. Lemma~\ref{lem:sp-limit} below turns this into a statement about the policies.

\begin{lemma}[The softplus losses are DPO with a margin and a success-conditioned anchor]\label{lem:sp-dpo}
Fix $\temp>0$ and put $\beta_{\mathrm{DPO}}\coloneqq1/\temp$. For every policy $\newpol$ with full support at $\state$,
\begin{equation}\label{eq:sp-dpo}
\begin{split}
\mathcal L^{\temp}_{\rw,\state}(\newpol;\base)
=\temp\Biggl\{&-\mathop{\E}\limits_{\substack{\act_+\sim\scond(\cdot\mid\state)\\\act_-\sim\fcond(\cdot\mid\state)}}\ln\sigma\left(\beta_{\mathrm{DPO}}\left(\ln\frac{\newpol(\act_+\mid\state)}{\base(\act_+\mid\state)}-\ln\frac{\newpol(\act_-\mid\state)}{\base(\act_-\mid\state)}\right)-\frac{\marg}{\temp}\right)\\
&+\frac{\lam}{\temp}\left(\Ent\left(\scond(\cdot\mid\state)\right)+\KL\left(\scond(\cdot\mid\state)\,\big\|\,\newpol(\cdot\mid\state)\right)\right)\Biggr\},
\end{split}
\end{equation}
and $\mathcal L^{\temp}_{\unw,\state}$ is the same expression with $\base(\cdot\mid\state)$ replaced by the uniform distribution on $\actset$, whose normalizer cancels in the difference of the log-ratios. 
\end{lemma}
\begin{proof}
Since $-\ln\sigma(x)=\ln(1+e^{-x})$, the term inside the expectation equals $\ln\left(1+\exp\left(\tfrac1\temp\left(\marg+\ln\frac{\newpol(\act_-\mid\state)}{\base(\act_-\mid\state)}-\ln\frac{\newpol(\act_+\mid\state)}{\base(\act_+\mid\state)}\right)\right)\right)$, which times $\temp$ is the softplus term of \cref{eq:sp-loss}; and $-\sum_\act\scond(\act\mid\state)\ln\newpol(\act\mid\state)=\Ent(\scond(\cdot\mid\state))+\KL(\scond(\cdot\mid\state)\|\newpol(\cdot\mid\state))$, as in the proof of Lemma~\ref{lem:sucfailnext}.
\end{proof}

\begin{remark}
    The first term inside the braces is the DPO objective with coefficient $\beta_{\mathrm{DPO}}$, reference $\base(\cdot\mid\state)$, preference pairs drawn from $\scond(\cdot\mid\state)\otimes\fcond(\cdot\mid\state)$ with the successful action preferred, and a margin $\marg/\temp$ in the preference logit; the second is the success-conditioned likelihood of Lemma~\ref{lem:sucfailnext}, which anchors $\newpol$ to $\scond$. In particular, at $\lam=\marg=0$ the \reweighted\ softplus loss is $\temp$ times the DPO objective with reference $\base$, and the \unweighted\ one is $\temp$ times the DPO objective with the uniform reference.
\end{remark}

Multiplication by the positive factor $\temp$ does not change minimizers, so every statement below about $\pirw^{\temp}$ is a statement about the population minimizer of DPO with reference $\base$, preference margin $\marg/\temp$ and anchor weight $\lam/\temp$. Two features of \cref{eq:sp-dpo} drive everything that follows. However, we make a note of the fact that, here the pairs are not labeled preferences between two fixed candidates but a successful and a failed action at the same state, so the preference model is determined by $\mathrm{Ber}(\Q)$ and not the usual Bradley Terry model in DPO.

\subsection{What plain DPO learns from success and failure}
\label{app:softplus:bt}

In this section, we discuss how the optimizers of the plain DPO compare to the base policy $\base$ and the success-conditioned policy $\scond$ that comes by optimizing the SFT-loss.

\begin{lemma}[The induced preference model and the DPO endpoint]\label{lem:sp-bt}
Fix a non-degenerate state $\state$ with $0<\Q(\state,\act)<1$ for every $\act$, and $\temp>0$.
\begin{enumerate}
\item At $\lam=\marg=0$, both softplus losses have, up to an additive constant, the unique minimizing log-probability score given by $u(\act)=\temp\ln\frac{\Q(\state,\act)}{1-\Q(\state,\act)}$ in the parametrization \cref{eq:comono:ii}, so that
\begin{equation}\label{eq:sp-endpoint}
\piunw^{\temp}(\act\mid\state)\ \propto\ \left(\frac{\Q(\state,\act)}{1-\Q(\state,\act)}\right)^{\temp},\,
\pirw^{\temp}(\act\mid\state)\ \propto\ \base(\act\mid\state)\left(\frac{\Q(\state,\act)}{1-\Q(\state,\act)}\right)^{\temp}.
\end{equation}
\item $\val{\pirw^{\temp}}(\state)>\val{\base}(\state)$ whenever $\Q(\state,\cdot)$ is non-constant, $\pirw^{\temp}(\cdot\mid\state)\to\base(\cdot\mid\state)$ as $\temp\downarrow0$, and
\begin{equation}\label{eq:sp-endpoint-diff}
\val{\pirw^{\temp}}(\state)-\val{\piunw^{\temp}}(\state)=\frac{\Cov_{\act\sim\piunw^{\temp}(\cdot\mid\state)}\left(\Q(\state,\act),\base(\act\mid\state)\right)}{\sum_\act\piunw^{\temp}(\act\mid\state)\base(\act\mid\state)},
\end{equation}
which is nonnegative if $\base$ is \comon\ with $\Q$ at $\state$ and non positive if \noncomon.
\end{enumerate}
\end{lemma}
\begin{proof}
We start by noticing that $\scond(\cdot\mid\state)$ and $\fcond(\cdot\mid\state)$ have full support and $\mathcal A_{+}=\actset$. So, as noted in the proof of Theorem~\ref{thm:comono}, in the score $u$ of \cref{eq:comono:ii} both losses are, up to additive constants, the objective $C(u)$ of \cref{eq:comono:iii} with the hinge replaced by the softplus. For an unordered pair $\{\act,\act'\}$ of distinct actions, the two terms of $C_\temp$ involving it depend on $u$ only through $d\coloneqq u(\act)-u(\act')$, as
\[
\scond(\act\mid\state)\fcond(\act'\mid\state)\,\temp\ln\left(1+e^{-d/\temp}\right)+\scond(\act'\mid\state)\fcond(\act\mid\state)\,\temp\ln\left(1+e^{d/\temp}\right),
\]
a strictly convex function of $d$ whose derivative, $-\scond(\act\mid\state)\fcond(\act'\mid\state)\,\sigma(-d/\temp)+\scond(\act'\mid\state)\fcond(\act\mid\state)\,\sigma(d/\temp)$, vanishes exactly when $e^{d/\temp}=\scond(\act\mid\state)\fcond(\act'\mid\state)\big/\left(\scond(\act'\mid\state)\fcond(\act\mid\state)\right)$, exactly when $d=\temp\left(\ln\frac{\Q(\state,\act)}{1-\Q(\state,\act)}-\ln\frac{\Q(\state,\act')}{1-\Q(\state,\act')}\right)$. The score $u=\temp\ln\frac{\Q(\state,\cdot)}{1-\Q(\state,\cdot)}$ attains every pairwise minimum simultaneously, hence minimizes $C_\temp$. Note that it is the unique minimizer modulo constants because $C_\temp$ is strictly convex on $\{\sum_\act u(\act)=0\}$, its Hessian being the graph Laplacian $K$ of the proof of Theorem~\ref{thm:comono}, all of whose edge weights are positive when every pair weight is. Converting the score $u$ to the induced policies $\pirw, \piunw$ using \cref{eq:comono:ii} gives \cref{eq:sp-endpoint}.

 By \cref{eq:sp-endpoint}, $\pirw^{\temp}(\act\mid\state)/\base(\act\mid\state)$ is a strictly increasing function of $\Q(\state,\act)$, so Lemma~\ref{lem:cov} gives $\sum_\act\left(\pirw^{\temp}(\act\mid\state)-\base(\act\mid\state)\right)\Q(\state,\act)=\Cov_{\act\sim\base(\cdot\mid\state)}\left(\pirw^{\temp}(\act\mid\state)/\base(\act\mid\state),\Q(\state,\act)\right)>0$ when $\Q(\state,\cdot)$ is non-constant. The exponent $\temp$ in \cref{eq:sp-endpoint} gives the limit as $\temp\downarrow0$. Finally \cref{eq:sp-endpoint-diff} as in the proof of Theorem~\ref{thm:comono}, applied to the log-probability score $u$ and its sign under alignment gives \cref{eq:comono:i}.
\end{proof}

Item~1 from the above result is the closed form of DPO under a Bradley--Terry model \citep{dpo2023}, specialized to the pair law that success--failure sampling induces. Two consequences matter for the comparison with success conditioning. A strictly increasing transformation of $\Q$ preserves the order of the actions but not expected rewards, so plain DPO does not maximize $\val{\pol}(\state)$ over a KL-ball; and the trust region is around $\base$, whereas the comparator of this paper is $\scond$: as $\temp\downarrow0$, i.e.\ as the DPO coefficient grows, $\pirw^{\temp}$ returns to $\base$, while $\val{\scond}(\state)-\val{\base}(\state)=\Var_{\act\sim\base(\cdot\mid\state)}\left[\Q(\state,\act)\right]/\val{\base}(\state)>0$ for non-constant $\Q(\state,\cdot)$, by Lemma~\ref{lem:cov} applied to $\scond(\act\mid\state)/\base(\act\mid\state)=\Q(\state,\act)/\val{\base}(\state)$. Plain DPO can therefore improve the base and still be worse than success conditioning. The following two-action case says exactly when this happens, and shows why including a positive margin is beneficial to mitigate this.

\begin{lemma}[Exact two-action margin threshold, independent of $\lam$]\label{lem:sp-binary}
Consider a non-degenerate state $\state$ with two actions $\act_1,\act_2$ and $1>\Q(\state,\act_1)>\Q(\state,\act_2)>0$, and fix $\temp>0$ and $\lam>0$. We define $A\coloneqq\scond(\act_1\mid\state)\fcond(\act_2\mid\state)$ and $B\coloneqq\scond(\act_2\mid\state)\fcond(\act_1\mid\state)$, and define the two constants $\theta_{\rw}\coloneqq\frac{\Q(\state,\act_1)}{\Q(\state,\act_2)}$ and $\theta_{\unw}\coloneqq\frac{\scond(\act_1\mid\state)}{\scond(\act_2\mid\state)}$. We define $\delta^{\mathrm{crit}}_{\unw}(\tau)\coloneqq\tau\log\max\left\{1,
    \frac{\theta_{\unw}^{2/\temp}-A/B}{(A/B-1)\,\theta_{\unw}^{1/\temp}}\right\}$ and $\delta^{\mathrm{crit}}_{\rw}(\tau)\coloneqq\tau\log\max\left\{1,
    \frac{\theta_{\rw}^{2/\temp}-A/B}{(A/B-1)\,\theta_{\rw}^{1/\temp}}\right\}$. Then, for both \unweighted and \reweighted losses and every $\marg\ge0$, we have:
\begin{equation}\label{eq:sp-binary}
\val{\pi^{\temp}_\star}(\state)\ \ge\ \val{\scond}(\state)
\iff\pi^{\temp}_\star(\act_1\mid\state)\ \ge\ \scond(\act_1\mid\state)
\iff\ \marg\ \ge\ \delta^{\mathrm{crit}}_{\rw}(\tau),
\end{equation}
where $\star\in\{\mathrm{rw},\mathrm{unw}\}$. The threshold on the right tends to $\max\{0, \ln\theta_\star\}$ as $\temp\downarrow0$.

\end{lemma}
\begin{proof}
With two actions, $\val{\pol}(\state)-\val{\scond}(\state)=\left(\pol(\act_1)-\scond(\act_1\mid\state)\right)\left(\Q(\state,\act_1)-\Q(\state,\act_2)\right)$ for every policy $\pol$. We define $u$ to be the log probability $\ln\pol(\cdot\mid\state)$ (normalized with that of $\base$, defined as $\ln\pol(\cdot\mid\state) - \ln\base(\cdot\mid\state)$ in case of the \reweighted objective) of the policy, which we call the ``score'', like usual. We parametrize policies by the score gap $g\coloneqq u(\act_1)-u(\act_2)$, so that $\pol^{u}_0(\act_1\mid\state)=\sigma(g)$ and $\pol^{u}_1(\act_1\mid\state)=\sigma\left(g+\ln\frac{\base(\act_1\mid\state)}{\base(\act_2\mid\state)}\right)$ by \cref{eq:comono:ii}. Up to the constant self-pair terms, the softplus state loss is given by:
\[
F(g)=A\,\temp\ln\left(1+e^{(\marg-g)/\temp}\right)+B\,\temp\ln\left(1+e^{(\marg+g)/\temp}\right)+\lam\,\KL\left(\scond(\cdot\mid\state)\,\|\,\pol^{u}_t(\cdot\mid\state)\right),
\]
with $t=0$ for the \unweighted\ and $t=1$ for the \reweighted\ versions. The first two terms are convex in $g$ and the last is strictly convex, its second derivative being $\pol^u_t(\act_1\mid\state)\pol^u_t(\act_2\mid\state)>0$; so $F$ has a unique minimizer $\hat g$, and for any $g^\circ$, $\hat g\ge g^\circ$ if and only if $F'(g^\circ)\le0$. Take $g^\circ$ to be the gap at which $\pol^u_t(\act_1\mid\state)=\scond(\act_1\mid\state)$. It is $\ln\theta_{\unw}$ for $t=0$, and $\ln\theta_{\rw}$ for $t=1$ because $\ln\frac{\scond(\act_1\mid\state)}{\scond(\act_2\mid\state)}-\ln\frac{\base(\act_1\mid\state)}{\base(\act_2\mid\state)}=\ln\frac{\Q(\state,\act_1)}{\Q(\state,\act_2)}$. Thus $\pi^{\temp}_\star(\act_1\mid\state)\ge\scond(\act_1\mid\state)$ iff $\hat g\ge g^\circ$ iff $F'(g^\circ)\le0$. The derivative of the anchor term is $\lam\left(\pol^u_t(\act_1\mid\state)-\scond(\act_1\mid\state)\right)$ which vanishes at $g^\circ$, so we have:
\[
F'(g^\circ)=B\,\sigma\left(\frac{\marg+g^\circ}{\temp}\right)-A\,\sigma\left(\frac{\marg-g^\circ}{\temp}\right).
\]
Note that this does not involve $\lam$. With $x\coloneqq e^{\marg/\temp}\ge1$ and $w\coloneqq e^{g^\circ/\temp}=\theta_\star^{1/\temp}$ we have $\sigma((\marg+g^\circ)/\temp)=\frac{xw}{1+xw}$ and $\sigma((\marg-g^\circ)/\temp)=\frac{x}{x+w}$, so $F'(g^\circ)\le0$ reads $w(w+x)\le\frac AB(1+xw)$, i.e.\ $w^2-\frac AB\le\left(\frac AB-1\right)xw$. If $w^2\le A/B$ this holds for every $x\ge1$; otherwise it holds iff $x\ge\frac{w^2-A/B}{(A/B-1)w}$, and this right side is at most $1$ exactly when $(w-A/B)(w+1)\le0$, i.e.\ $w\le A/B$. Taking logarithms gives \cref{eq:sp-binary}. As $\temp\downarrow0$, $\temp\ln\frac{w^2-A/B}{(A/B-1)w}=\ln\theta_\star+\temp\ln\left(1-\tfrac AB\theta_\star^{-2/\temp}\right)-\temp\ln(A/B-1)\to\ln\theta_\star$. Finally $\theta_{\rw}<A/B$ because $1-\Q(\state,\act_2)>1-\Q(\state,\act_1)$, and $\theta_{\rw}^{1/\temp}\le A/B$ iff $\temp\ge\ln\theta_{\rw}/\ln(A/B)$.
\end{proof}

\begin{remark}
    Since $\theta_{\rw}<A/B$, the \reweighted\ policy satisfies $\val{\pirw^{\temp}}(\state)\ge\val{\scond}(\state)$ for every $\marg\ge0$ as soon as $\temp\ge\ln\theta_{\rw}/\ln(A/B)$; no temperature satisfies this for the \unweighted\ readout uniformly over base policies, because $\theta_{\unw}=\theta_{\rw}\,\base(\act_1\mid\state)/\base(\act_2\mid\state)$ is unbounded.
\end{remark}

In the instance $\base(\cdot\mid\state)=(\tfrac12,\tfrac12)$ and $\Q(\state,\cdot)=(\tfrac45,\tfrac15)$ one has $\theta_{\rw}=\theta_{\unw}=4$ and $A/B=16$, so at zero margin both DPO policies are worse than $\scond$ exactly when $\temp<\tfrac12$, i.e.\ $\beta_{\mathrm{DPO}}>2$, and by \cref{eq:sp-binary} no anchor weight $\lam$ changes this. However, a margin $\marg\ge\temp\ln\frac{16^{1/\temp}-16}{15\cdot4^{1/\temp}}$ does, and this threshold increases to $\ln4$ as $\temp\downarrow0$. This is basically the version of Lemma~\ref{lem:rw-sft} at positive temperature: the failure of the \reweighted\ loss against success conditioning can be mitigated by a positive margin.

\subsection{Existence, support and stationarity at positive temperature}
\label{app:softplus:foc}

In this section, we analyze the existence of the optimizers of both the \reweighted and \unweighted losses for the softplus version. We also derive the stationarity condition at the optimizer, an expression that we shall repeatedly use to prove its properties.

\begin{lemma}[Existence, support and stationarity]\label{lem:sp-support}
Fix a non-degenerate state $\state$, $\temp>0$, $\lam>0$, $\marg\ge0$. Then, the following holds:
\begin{enumerate}
\item For both the \reweighted and \unweighted objectives, there is a unique policy $\pi^{\temp}_{\mathrm{rw}}(\cdot\mid\state)$ (respectively, $\pi^{\temp}_{\mathrm{unw}}(\cdot\mid\state)$), supported exactly on $\mathcal A_{+}$, such that every minimizing sequence of $\mathcal L^{\temp}_{\mathrm{rw},\state}(\cdot;\base)$ (respectively, that of $\mathcal L^{\temp}_{\mathrm{unw},\state}(\cdot;\base)$) induces policies converging to it, and we have:
\begin{equation}\label{eq:sp-certificate}
\mathcal L^{\temp}_{\mathrm{rw},\state}(\newpol;\base)-\inf_{\hat\pol}\mathcal L^{\temp}_{\mathrm{rw},\state}(\hat\pol;\base)\ \ge\ \lam\,\KL\left(\pi^{\temp}_{\mathrm{rw}}(\cdot\mid\state)\,\big\|\,\newpol(\cdot\mid\state)\right)
\end{equation}
for every policy $\newpol$ with full support, the infimum being over policies with full support. The exact same result holds for the \unweighted loss as well. On $\mathcal A_{+}$, $\pi^{\temp}_{\mathrm{unw}}(\cdot\mid\state)$ (respectively, $\pi^{\temp}_{\mathrm{rw}}(\cdot\mid\state)$) equals $\pol^{u}_t$ with $t=0$ (respectively, $t=1$) in the parametrization \cref{eq:comono:ii}, for a score $u:\mathcal A_{+}\to\mathbb R$ that is unique up to an additive constant. 
We write $\pi^{\temp}_t(\cdot\mid\state)$ for this policy, so that $\pi^{\temp}_0=\pi^{\temp}_{\mathrm{unw}}$ and $\pi^{\temp}_1=\pi^{\temp}_{\mathrm{rw}}$ on $\mathcal A_{+}$; it coincides with the policy $\pol_{t,\temp}$ constructed in the proof of Theorem~\ref{thm:comono}. 
\item Write $\pol_\star\coloneqq\pi^{\temp}_{\mathrm{rw}}(\cdot\mid\state)$ (respectively, $\pi^{\temp}_{\mathrm{unw}}(\cdot\mid\state)$) and, $u$ as the $\base$-normalized log-probability score $\ln\pol(\cdot\mid\state) - \ln\base(\cdot\mid\state)$ (respectively, the raw log-probability $\ln\pol(\cdot\mid\state)$ for \unweighted loss) for the corresponding policies $\pol$. Then, for $\act\in\mathcal A_{+}$, we define the following constants:
\begin{equation}\label{eq:sp-AB}
\begin{split}
    A_\act&\coloneqq\sum_{\act'\in\mathcal A_{+}}\scond(\act'\mid\state)\,\sigma\left(\frac{\marg+u(\act)-u(\act')}{\temp}\right),\\
B_\act&\coloneqq\sum_{\act'\in\mathcal A_{+}}\fcond(\act'\mid\state)\,\sigma\left(\frac{\marg+u(\act')-u(\act)}{\temp}\right).
\end{split}
\end{equation}
Then, for every $\act\in\mathcal A_{+}$, we have:
\begin{equation}\label{eq:sp-foc}
\lam\left(\pol_\star(\act)-\scond(\act\mid\state)\right)=\scond(\act\mid\state)\,B_\act-\fcond(\act\mid\state)\,A_\act .
\end{equation}
\item For every $\act\in\mathcal A_{+}$, we have:
\begin{equation}\label{eq:sp-mass}
\begin{split}
&\scond(\act\mid\state)\left(1-\frac1\lam\left(\frac{\fcond(\act\mid\state)}{\scond(\act\mid\state)}-\fcond(\act\mid\state)\right)\right)\ \le\ \pol_\star(\act)\\
\le& \scond(\act\mid\state)\left(1+\frac{\fcond(\mathcal A_{+}\mid\state)-\fcond(\act\mid\state)}{\lam}\right),
\end{split}
\end{equation}
where $\pol_\star$ is the optimizer of the contrastive losses, and this is true for both the \reweighted and \unweighted cases.
\end{enumerate}
\end{lemma}
\begin{proof}
Fix the state $\state$ and $t\in\{0,1\}$, and write $\mathcal L_\state$ for $\mathcal L^{\temp}_{\mathrm{unw},\state}(\cdot;\base)$ if $t=0$ and for $\mathcal L^{\temp}_{\mathrm{rw},\state}(\cdot;\base)$ if $t=1$. By the standing assumption of the state $\state$ being \emph{non-degenerate}, we have that $\base(\act\mid\state)>0$ for every $\act$ and $0<\val{\base}(\state)<1$. So, $\scond(\cdot\mid\state)$ and $\fcond(\cdot\mid\state)$ are given by \cref{eq:cond}: $\scond(\cdot\mid\state)$ is supported exactly on $\mathcal A_{+}=\{\act:\Q(\state,\act)>0\}$ and is positive there. It should be noted that $\fcond(\cdot\mid\state)$ may vanish on $\mathcal A_{+}$, at the actions with $\Q(\state,\act)=1$, $\fcond(\mathcal A_{+}\mid\state)\le1$; and $\mathcal A_{+}\ne\emptyset$ because $\val{\base}(\state)>0$. 

Two properties of the softplus $z\mapsto\temp\ln(1+e^{z/\temp})$ are used throughout: it is positive, increasing and convex, with derivative $\sigma(z/\temp)\in(0,1)$; and it tends to $0$ as $z\to-\infty$. In both losses the policy enters the argument of the softplus only through the difference of two scores, $\ln\newpol(\act_-\mid\state)-\ln\newpol(\act_+\mid\state)$ for $t=0$ and $\ln\frac{\newpol(\act_-\mid\state)}{\base(\act_-\mid\state)}-\ln\frac{\newpol(\act_+\mid\state)}{\base(\act_+\mid\state)}$ for $t=1$, and this difference does not change when $\ln\newpol(\cdot\mid\state)$ is shifted by the same constant at both actions. As in Section~\ref{sec:setup}, a minimizing sequence of $\mathcal L_\state$ is a sequence of policies with full support, equivalently of finite scores, along which $\mathcal L_\state$ tends to its infimum over policies with full support; $\inf\mathcal L_\state$ denotes that infimum.

We first reduce the problem to actions with positive reward. Let $\mathcal L_{\mathcal A_{+}}$ be $\mathcal L_\state$ with both sums restricted to $\mathcal A_{+}$, the weights $\scond(\act_+\mid\state)\fcond(\act_-\mid\state)$ not re-normalized; it is a function of policies on $\mathcal A_{+}$, and if $\mathcal A_{+}=\actset$ it is $\mathcal L_\state$ itself. Let $\pol$ be a policy with full support on $\actset$ and define $\bar\pol\coloneqq\pol(\cdot\mid\state)/\pol(\mathcal A_{+}\mid\state)$ on $\mathcal A_{+}$. There are no pairs with $\act_+\notin\mathcal A_{+}$, since $\scond(\act_+\mid\state)=0$ there, so $\mathcal L_\state(\pol)$ is the sum of the softplus terms with $\act_+,\act_-\in\mathcal A_{+}$, the softplus terms with $\act_+\in\mathcal A_{+}$ and $\act_-\notin\mathcal A_{+}$, and the success-conditioning term $-\lam\sum_{\act\in\mathcal A_{+}}\scond(\act\mid\state)\ln\pol(\act\mid\state)$. For $\act\in\mathcal A_{+}$, $\ln\pol(\act\mid\state)=\ln\bar\pol(\act)+\ln\pol(\mathcal A_{+}\mid\state)$, a shift by a constant; hence the softplus terms with $\act_+,\act_-\in\mathcal A_{+}$ are the same at $\pol$ and at $\bar\pol$, and, as $\sum_{\act\in\mathcal A_{+}}\scond(\act\mid\state)=1$, the success-conditioning term equals $-\lam\sum_{\act\in\mathcal A_{+}}\scond(\act\mid\state)\ln\bar\pol(\act)-\lam\ln\pol(\mathcal A_{+}\mid\state)$. The remaining softplus terms are positive and $\ln\pol(\mathcal A_{+}\mid\state)\le0$, so we get:
\begin{equation}\label{eq:sp-support:red}
\mathcal L_{\state}(\pol)\ \ge\ \mathcal L_{\mathcal A_{+}}(\bar\pol)-\lam\ln\pol(\mathcal A_{+}\mid\state)\ \ge\ \mathcal L_{\mathcal A_{+}}(\bar\pol).
\end{equation}
Conversely, let $\bar\pol$ be a policy with full support on $\mathcal A_{+}$ and, for $n\ge2$, let $\pol$ be the policy with $\pol(\act\mid\state)=(1-1/n)\,\bar\pol(\act)$ for $\act\in\mathcal A_{+}$ and $\pol(\act\mid\state)=1/(n\,|\actset\setminus\mathcal A_{+}|)$ for $\act\notin\mathcal A_{+}$. Its softplus terms with $\act_+,\act_-\in\mathcal A_{+}$ are those of $\bar\pol$, its success-conditioning term is $-\lam\sum_{\act\in\mathcal A_{+}}\scond(\act\mid\state)\ln\bar\pol(\act)-\lam\ln(1-1/n)$, and for $\act_+\in\mathcal A_{+}$ and $\act_-\notin\mathcal A_{+}$ the argument of the softplus is
\[
\marg-\ln n-\ln|\actset\setminus\mathcal A_{+}|-\ln(1-1/n)-\ln\bar\pol(\act_+)
\]
for $t=0$, and this plus $\ln\base(\act_+\mid\state)-\ln\base(\act_-\mid\state)$ for $t=1$. Both tend to $-\infty$ as $n\to\infty$, and there are finitely many such terms, so $\mathcal L_\state(\pol)\to\mathcal L_{\mathcal A_{+}}(\bar\pol)$. Together with \cref{eq:sp-support:red},
\begin{equation}\label{eq:sp-support:inf}
\inf\mathcal L_\state=\inf\mathcal L_{\mathcal A_{+}},
\end{equation}
the right side being the infimum over policies with full support on $\mathcal A_{+}$.

Now, every policy with full support on $\mathcal A_{+}$ can be written, for the fixed $t$, as
\begin{equation}\label{eq:sp-support:param}
\pol^{u}_{t}(\act\mid\state)=\frac{\base(\act\mid\state)^{t}\,e^{u(\act)}}{\sum_{\act'\in\mathcal A_{+}}\base(\act'\mid\state)^{t}\,e^{u(\act')}},\,\act\in\mathcal A_{+},
\end{equation}
for a score $u:\mathcal A_{+}\to\mathbb R$, unique up to an additive constant, namely $u=\ln\pol^u_t(\cdot\mid\state)-t\ln\base(\cdot\mid\state)+\mathrm{const}$: for $t=0$ the score is the log-probability and for $t=1$ the normalized log-probability, which are the two quantities inside the softplus of the two losses. At $\pol=\pol^u_t$ the argument of the softplus for the pair $(\act_+,\act_-)$ is therefore $\marg+u(\act_-)-u(\act_+)$, the normalizer of \cref{eq:sp-support:param} cancelling, and the success-conditioning term is
\begin{align*}
   & -\lam\sum_{\act\in\mathcal A_{+}}\scond(\act\mid\state)\ln\pol^u_t(\act\mid\state)
=\lam\left[\ln\sum_{\act\in\mathcal A_{+}}\base(\act\mid\state)^{t}e^{u(\act)}-\sum_{\act\in\mathcal A_{+}}\scond(\act\mid\state)\,u(\act)\right]\\
&-\lam t\sum_{\act\in\mathcal A_{+}}\scond(\act\mid\state)\ln\base(\act\mid\state),
\end{align*}
whose last term does not depend on $u$. Hence $\mathcal L_{\mathcal A_{+}}(\pol^u_t)=F_{t,\temp}(u)+\mathrm{const}$, where we define $F_{t,\temp}(u)$ as follows: 
\begin{equation}\label{eq:sp-support:F}
\begin{split}
F_{t,\temp}(u)&\coloneqq C_\temp(u)+\lam\left[\ln\sum_{\act\in\mathcal A_{+}}\base(\act\mid\state)^{t}e^{u(\act)}-\sum_{\act\in\mathcal A_{+}}\scond(\act\mid\state)\,u(\act)\right],\\
C_\temp(u)&\coloneqq\sum_{\act,\act'\in\mathcal A_{+}}\scond(\act\mid\state)\,\fcond(\act'\mid\state)\;\temp\ln\left(1+\exp\left(\frac{\marg+u(\act')-u(\act)}{\temp}\right)\right),
\end{split}
\end{equation}
Thus, minimizing $\mathcal L_{\mathcal A_{+}}$ over policies with full support on $\mathcal A_{+}$ is the same as minimizing $F_{t,\temp}$ over scores.

$F_{t,\temp}$ is unchanged when a constant is added to $u$, so it suffices to minimize it over the zero-sum scores, those with $\sum_{\act\in\mathcal A_{+}}u(\act)=0$. On this set it is coercive: $C_\temp\ge0$, and with $\bar u=\max_\act u(\act)$ and $\underline u=\min_\act u(\act)$, the logarithm in \cref{eq:sp-support:F} is at least $t\ln\base(\act\mid\state)+\bar u$ at an action $\act$ attaining $\bar u$, while $\sum_\act\scond(\act\mid\state)u(\act)\le\bar u-\scond(\act\mid\state)(\bar u-\underline u)$ at an action $\act$ attaining $\underline u$, because every other summand is at most $\scond(\act'\mid\state)\,\bar u$; so
\[
F_{t,\temp}(u)\ \ge\ \lam\,t\min_{\act\in\mathcal A_{+}}\ln\base(\act\mid\state)+\lam\min_{\act\in\mathcal A_{+}}\scond(\act\mid\state)\,(\bar u-\underline u),
\]
and on the zero-sum scores $\bar u-\underline u\ge\max_\act|u(\act)|$, since $\underline u\le0\le\bar u$. As $\min_{\act\in\mathcal A_{+}}\scond(\act\mid\state)>0$ and $\lam>0$, the right side tends to $+\infty$ with $\max_\act|u(\act)|$; the sub level sets of the continuous function $F_{t,\temp}$ on the zero-sum scores are therefore bounded and closed, hence compact, and $F_{t,\temp}$ attains its minimum there. The minimizer is unique. $C_\temp$ is convex, being a nonnegative combination of the convex softplus composed with affine functions of $u$; and the bracket in \cref{eq:sp-support:F} has second derivatives
\[
\frac{\partial^2}{\partial u(\act)\,\partial u(\act')}\ln\sum_{\act''\in\mathcal A_{+}}\base(\act''\mid\state)^{t}e^{u(\act'')}=\pol^u_t(\act\mid\state)\left(\ind\{\act=\act'\}-\pol^u_t(\act'\mid\state)\right),
\]
so that for every vector $v$ on $\mathcal A_{+}$ its quadratic form is $\sum_\act\pol^u_t(\act\mid\state)v(\act)^2-\left(\sum_\act\pol^u_t(\act\mid\state)v(\act)\right)^2$, the variance of $v$ under $\pol^u_t$, which vanishes only if $v$ is constant on $\mathcal A_{+}$, as $\pol^u_t$ has full support. A nonzero zero-sum vector is not constant, so along the zero-sum scores the bracket is strictly convex, hence so is $F_{t,\temp}$ for $\lam>0$, and two distinct minimizers would give a smaller value at their midpoint. Let $u_{t,\temp}$ be the minimizer and $\pol_{t,\temp}\coloneqq\pol^{u_{t,\temp}}_t$, a policy with full support on $\mathcal A_{+}$; by the preceding paragraph it is the unique minimizer of $\mathcal L_{\mathcal A_{+}}$ over policies with full support on $\mathcal A_{+}$, and by \cref{eq:sp-support:inf}
\begin{equation}\label{eq:sp-support:min}
\inf\mathcal L_\state=\inf\mathcal L_{\mathcal A_{+}}=\mathcal L_{\mathcal A_{+}}(\pol_{t,\temp}).
\end{equation}

Now we derive the stationarity condition at the infimum. The partial derivatives of $F_{t,\temp}$ are
\begin{equation}\label{eq:sp-support:grad}
\begin{split}
\frac{\partial F_{t,\temp}}{\partial u(\act)}(u)&=\sum_{\act'\in\mathcal A_{+}}\scond(\act'\mid\state)\fcond(\act\mid\state)\,\sigma\left(\frac{\marg+u(\act)-u(\act')}{\temp}\right)\\
&-\sum_{\act'\in\mathcal A_{+}}\scond(\act\mid\state)\fcond(\act'\mid\state)\,\sigma\left(\frac{\marg+u(\act')-u(\act)}{\temp}\right) +\lam\left(\pol^u_t(\act\mid\state)-\scond(\act\mid\state)\right),
\end{split}
\end{equation}
the first sum collecting the pairs in which $\act$ is the failed action, whose softplus argument increases with $u(\act)$, the second those in which $\act$ is the successful action, whose argument decreases with $u(\act)$, and the last term being the derivative of the bracket, since $\partial_{u(\act)}\ln\sum_{\act'}\base(\act'\mid\state)^te^{u(\act')}=\pol^u_t(\act\mid\state)$. Summed over $\act\in\mathcal A_{+}$, the two double sums cancel after exchanging the names of $\act$ and $\act'$ in one of them, and $\sum_\act\pol^u_t(\act\mid\state)=\sum_\act\scond(\act\mid\state)=1$; so the gradient of $F_{t,\temp}$ at every $u$ is a zero-sum vector. Since $u_{t,\temp}$ minimizes $F_{t,\temp}$ over the zero-sum scores, for every zero-sum $v$ the function $r\mapsto F_{t,\temp}(u_{t,\temp}+rv)$ is minimal at $r=0$, so the gradient at $u_{t,\temp}$ is orthogonal to every zero-sum $v$; being itself zero-sum, it is zero:
\begin{equation}\label{eq:sp-support:foc}
\frac{\partial F_{t,\temp}}{\partial u(\act)}(u_{t,\temp})=0\,\text{for every }\act\in\mathcal A_{+}.
\end{equation}

\emph{Item 1: the certificate.} Let $u$ be any score on $\mathcal A_{+}$. Convexity of $C_\temp$ gives
\[
C_\temp(u)\ \ge\ C_\temp(u_{t,\temp})+\sum_{\act\in\mathcal A_{+}}\frac{\partial C_\temp}{\partial u(\act)}(u_{t,\temp})\,\left(u(\act)-u_{t,\temp}(\act)\right),
\]
and the bracket satisfies an exact identity: by \cref{eq:sp-support:param}, for every $\act\in\mathcal A_{+}$,
\[
\ln\pol_{t,\temp}(\act\mid\state)-\ln\pol^u_t(\act\mid\state)=u_{t,\temp}(\act)-u(\act)+\ln\sum_{\act'}\base(\act'\mid\state)^te^{u(\act')}-\ln\sum_{\act'}\base(\act'\mid\state)^te^{u_{t,\temp}(\act')},
\]
and averaging this under $\pol_{t,\temp}$ gives
\[
\ln\sum_{\act}\base(\act\mid\state)^{t}e^{u(\act)}-\ln\sum_{\act}\base(\act\mid\state)^{t}e^{u_{t,\temp}(\act)}-\sum_\act\pol_{t,\temp}(\act\mid\state)\left(u(\act)-u_{t,\temp}(\act)\right)=\KL\left(\pol_{t,\temp}\,\|\,\pol^u_t\right).
\]
Multiplying the identity by $\lam$, adding it to the convexity inequality and using \cref{eq:sp-support:grad,eq:sp-support:foc},
\begin{equation}\label{eq:sp-support:cert}
F_{t,\temp}(u)-F_{t,\temp}(u_{t,\temp})\ \ge\ \sum_{\act\in\mathcal A_{+}}\frac{\partial F_{t,\temp}}{\partial u(\act)}(u_{t,\temp})\,\left(u(\act)-u_{t,\temp}(\act)\right)+\lam\,\KL\left(\pol_{t,\temp}\,\|\,\pol^u_t\right)=\lam\,\KL\left(\pol_{t,\temp}\,\|\,\pol^u_t\right).
\end{equation}
In terms of policies, $\mathcal L_{\mathcal A_{+}}(\bar\pol)-\mathcal L_{\mathcal A_{+}}(\pol_{t,\temp})\ge\lam\KL(\pol_{t,\temp}\|\bar\pol)$ for every $\bar\pol$ with full support on $\mathcal A_{+}$. Now let $\pol$ be any policy with full support on $\actset$ and $\bar\pol$ its conditional law on $\mathcal A_{+}$ as above. Since $\pol_{t,\temp}$ is supported on $\mathcal A_{+}$, where $\pol(\act\mid\state)=\pol(\mathcal A_{+}\mid\state)\,\bar\pol(\act)$,
\[
\KL(\pol_{t,\temp}\|\pol)=\sum_{\act\in\mathcal A_{+}}\pol_{t,\temp}(\act\mid\state)\ln\frac{\pol_{t,\temp}(\act\mid\state)}{\pol(\mathcal A_{+}\mid\state)\,\bar\pol(\act)}=\KL(\pol_{t,\temp}\|\bar\pol)-\ln\pol(\mathcal A_{+}\mid\state),
\]
and therefore, by \cref{eq:sp-support:red,eq:sp-support:min} and the policy form of \cref{eq:sp-support:cert}, we get:
\begin{align*}
\mathcal L_{\state}(\pol)-\inf\mathcal L_{\state}\ &\ge\ \mathcal L_{\mathcal A_{+}}(\bar\pol)-\lam\ln\pol(\mathcal A_{+}\mid\state)-\mathcal L_{\mathcal A_{+}}(\pol_{t,\temp})\\ &\ge\ \lam\KL(\pol_{t,\temp}\|\bar\pol)-\lam\ln\pol(\mathcal A_{+}\mid\state)=\lam\,\KL(\pol_{t,\temp}\|\pol).
\end{align*}
Define $\pi^{\temp}_{\mathrm{unw}}(\cdot\mid\state)$ if $t=0$, and $\pi^{\temp}_{\mathrm{rw}}(\cdot\mid\state)$ if $t=1$, as the policy equal to $\pol_{t,\temp}$ on $\mathcal A_{+}$ and to $0$ off $\mathcal A_{+}$. It is supported exactly on $\mathcal A_{+}$, the last display is \cref{eq:sp-certificate} for it, and on $\mathcal A_{+}$ it is $\pi^{\temp}_t(\cdot\mid\state)=\pol^{u_{t,\temp}}_t$, with the score $u_{t,\temp}$ unique up to an additive constant by \cref{eq:sp-support:param}.

Now we proceed to show the convergence of minimizing sequences. For any two probability vectors $p,q$ on a finite set, $\sum_\act|p(\act)-q(\act)|\le\sqrt{2\,\KL(p\|q)}$; this is Pinsker's inequality \cite{coverinformationtheory}. If $\pol_n$ is a minimizing sequence of $\mathcal L_\state$, then by \cref{eq:sp-certificate} $\lam\KL(\pol_{t,\temp}\|\pol_n)\le\mathcal L_\state(\pol_n)-\inf\mathcal L_\state\to0$, hence $\sum_\act|\pi^{\temp}_t(\act\mid\state)-\pol_n(\act\mid\state)|\to0$, where $\pi^{\temp}_t(\cdot\mid\state)$ is read as $0$ off $\mathcal A_{+}$. Minimizing sequences exist, for instance the policies constructed in the proof of \cref{eq:sp-support:inf} from $\bar\pol=\pol_{t,\temp}$, and a convergent sequence has one limit, so the policy just defined is the unique one with the property of item~1. Finally, $F_{t,\temp}$ is the smoothed objective of the proof of Theorem~\ref{thm:comono} at smoothing parameter $\temp$, and $\pol_{t,\temp}$ is its minimizer, which is the last claim of item~1.

 The policy $\pol_\star$ is as defined before, for $t=1$ in the \reweighted\ and $t=0$ in the \unweighted\ case, and its score is $u=\ln\pol_\star-t\ln\base(\cdot\mid\state)=u_{t,\temp}+\mathrm{const}$ on $\mathcal A_{+}$ by \cref{eq:sp-support:param}; the constant cancels in the differences $u(\act)-u(\act')$ of \cref{eq:sp-AB}. So $A_\act$ and $B_\act$ are the two sums in \cref{eq:sp-support:grad} evaluated at $u_{t,\temp}$, with the common factors taken out: the first sum in \cref{eq:sp-support:grad} is $\fcond(\act\mid\state)A_\act$ and the second is $\scond(\act\mid\state)B_\act$. Since $\pol^{u_{t,\temp}}_t(\act\mid\state)=\pol_\star(\act)$ for $\act\in\mathcal A_{+}$, \cref{eq:sp-support:foc} reads
\[
\fcond(\act\mid\state)A_\act-\scond(\act\mid\state)B_\act+\lam\left(\pol_\star(\act)-\scond(\act\mid\state)\right)=0,
\]
which is \cref{eq:sp-foc}.

Now, every value of $\sigma$ lies in $(0,1)$ and the weights in \cref{eq:sp-AB} are nonnegative. Keeping the term $\act'=\act$ of each sum, which equals $\scond(\act\mid\state)\sigma(\marg/\temp)$ in $A_\act$ and $\fcond(\act\mid\state)\sigma(\marg/\temp)$ in $B_\act$, and bounding the remaining terms below by $0$ and above by their weights, whose totals are $1-\scond(\act\mid\state)$ and $\fcond(\mathcal A_{+}\mid\state)-\fcond(\act\mid\state)$,
\begin{align*}
\scond(\act\mid\state)\,\sigma(\marg/\temp)&\le A_\act\le\scond(\act\mid\state)\,\sigma(\marg/\temp)+1-\scond(\act\mid\state),\\
\fcond(\act\mid\state)\,\sigma(\marg/\temp)&\le B_\act\le\fcond(\act\mid\state)\,\sigma(\marg/\temp)+\fcond(\mathcal A_{+}\mid\state)-\fcond(\act\mid\state).
\end{align*}
In the right side of \cref{eq:sp-foc}, use the upper bound on $B_\act$ and the lower bound on $A_\act$ for an upper bound, and the lower bound on $B_\act$ and the upper bound on $A_\act$ for a lower bound; in both the terms containing $\sigma(\marg/\temp)$, namely $\pm\scond(\act\mid\state)\fcond(\act\mid\state)\sigma(\marg/\temp)$, cancel, and
\[
-\fcond(\act\mid\state)\left(1-\scond(\act\mid\state)\right)\ \le\ \lam\left(\pol_\star(\act)-\scond(\act\mid\state)\right)\ \le\ \scond(\act\mid\state)\left(\fcond(\mathcal A_{+}\mid\state)-\fcond(\act\mid\state)\right).
\]
Dividing by $\lam$ and adding $\scond(\act\mid\state)$ gives \cref{eq:sp-mass}, because $\fcond(\act\mid\state)\left(1-\scond(\act\mid\state)\right)=\scond(\act\mid\state)\left(\fcond(\act\mid\state)/\scond(\act\mid\state)-\fcond(\act\mid\state)\right)$.
\end{proof}

\subsection{The hinge as the zero-temperature limit}
\label{app:softplus:limit}
Before proceeding to prove the results on the value function of the softplus version of the two contrastive losses, we first show what we have been claiming so far in this section, that the $\temp\downarrow0$ limit of the contrastive losses is the contrastive hinge loss.

\begin{lemma}[Temperature stability]\label{lem:sp-limit}
Fix a non-degenerate state $\state$, $\lam>0$ and $\marg\ge0$, and let $\pi^{0}_{\mathrm{rw}}(\cdot\mid\state)\coloneqq\pirw(\cdot\mid\state)$ and $\pi^{0}_{\mathrm{unw}}(\cdot\mid\state)\coloneqq\piunw(\cdot\mid\state)$ denote the hinge policies of Section~\ref{sec:setup}. For $0\le\temp'<\temp$,
\begin{equation}\label{eq:sp-stability}
\max\left\{\KL\left(\pi^{\temp'}_{\mathrm{rw}}(\cdot\mid\state)\,\big\|\,\pi^{\temp}_{\mathrm{rw}}(\cdot\mid\state)\right),\,\KL\left(\pi^{\temp}_{\mathrm{rw}}(\cdot\mid\state)\,\big\|\,\pi^{\temp'}_{\mathrm{rw}}(\cdot\mid\state)\right)\right\} \le\ \frac{(\temp-\temp')\ln2}{\lam}.
\end{equation}
Consequently, we have:
\[
\left\|\pi^{\temp}_{\mathrm{rw}}(\cdot\mid\state)-\pi^{\temp'}_{\mathrm{rw}}(\cdot\mid\state)\right\|_1\le\sqrt{\frac{2(\temp-\temp')\ln2}{\lam}},\,
\left|\val{\pi^{\temp}_{\mathrm{rw}}}(\state)-\val{\pi^{\temp'}_{\mathrm{rw}}}(\state)\right|\le\osc\left(\Q(\state,\cdot)\right)\sqrt{\frac{(\temp-\temp')\ln2}{2\lam}}.
\] In a finite-horizon MDP with the same $\lam>0$ at every state, $|\val{\pi^{\temp}_{\mathrm{rw}}}(\initlaw)-\val{\pirw}(\initlaw)|\le\horizon\sqrt{\temp\ln2/(2\lam)}$. The same bounds hold for the \unweighted\ policies.
\end{lemma}
\begin{proof}
Fix the state $\state$ and $t\in\{0,1\}$, with $t=1$ for the \reweighted\ and $t=0$ for the \unweighted\ policies; the argument is the same for both. By Lemma~\ref{lem:sp-support}, for every $\temp>0$ the policy $\pi^{\temp}_{\mathrm{rw}}(\cdot\mid\state)$ (respectively, $\pi^{\temp}_{\mathrm{unw}}(\cdot\mid\state)$) is supported exactly on $\mathcal A_{+}$ and equals there $\pol_{t,\temp}=\pol^{u_{t,\temp}}_t$, where $u_{t,\temp}$ is the unique minimizer over the zero-sum scores of the function $F_{t,\temp}$ of \cref{eq:sp-support:F}. By the proof of Theorem~\ref{thm:comono}, the hinge policy $\pirw(\cdot\mid\state)$ (respectively, $\piunw(\cdot\mid\state)$) is supported exactly on $\mathcal A_{+}$ and equals there $\pol_t=\pol^{u_t}_t$, where $u_t$ is the unique minimizer over the zero-sum scores of the function $F_t$ of \cref{eq:comono:iv}, which is $F_{t,\temp}$ with every softplus replaced by the hinge. We write $F_{t,0}\coloneqq F_t$, $u_{t,0}\coloneqq u_t$ and $\pol_{t,0}\coloneqq\pol_t$, so that $\pi^{\temp}_{\mathrm{rw}}(\cdot\mid\state)=\pol_{1,\temp}$ and $\pi^{\temp}_{\mathrm{unw}}(\cdot\mid\state)=\pol_{0,\temp}$ on $\mathcal A_{+}$ for every $\temp\ge0$. All the policies compared below are therefore positive on $\mathcal A_{+}$ and zero off it, so their $\KL$ divergences are finite and, like their $\ell^1$ distances, are sums over $\mathcal A_{+}$.

Two facts are used. The first is that the objectives at two temperatures are uniformly close: for every score $u$ on $\mathcal A_{+}$ and $0\le\temp'<\temp$,
\begin{equation}\label{eq:sp-limit:close}
0\ \le\ F_{t,\temp}(u)-F_{t,\temp'}(u)\ \le\ (\temp-\temp')\ln2 .
\end{equation}
The bracket in \cref{eq:sp-support:F} does not depend on the temperature, so $F_{t,\temp}(u)-F_{t,\temp'}(u)=C_\temp(u)-C_{\temp'}(u)$ is the sum over the pairs $\act,\act'\in\mathcal A_{+}$, with the weights $\scond(\act\mid\state)\fcond(\act'\mid\state)$, of the differences $\temp\ln(1+e^{z/\temp})-\temp'\ln(1+e^{z/\temp'})$ at $z=\marg+u(\act')-u(\act)$, where $\temp'\ln(1+e^{z/\temp'})$ is read as $\hinge{z}$ when $\temp'=0$. The weights are nonnegative and sum to $\fcond(\mathcal A_{+}\mid\state)\le1$, so it suffices to show that each difference lies in $[0,(\temp-\temp')\ln2]$. For fixed $z$ and $r>0$,
\[
\frac{d}{dr}\left[r\ln\left(1+e^{z/r}\right)\right]=\ln\left(1+e^{z/r}\right)-\frac zr\,\sigma(z/r)=-p\ln p-(1-p)\ln(1-p)\,\text{at }p=\sigma(z/r),
\]
because $1-\sigma(z/r)=(1+e^{z/r})^{-1}$ and $z/r=\ln\frac{p}{1-p}$. The right side is nonnegative, as $0<p<1$, and it is at most $\ln2$: by the concavity of the logarithm, $p\ln\frac1p+(1-p)\ln\frac1{1-p}\le\ln\left(p\cdot\frac1p+(1-p)\cdot\frac1{1-p}\right)=\ln2$. Integrating over $r\in[\temp',\temp]$ gives the claim when $\temp'>0$. When $\temp'=0$, $\temp\ln(1+e^{z/\temp})-\hinge{z}=\temp\ln(1+e^{-|z|/\temp})$, as one checks separately for $z\ge0$ and $z<0$, and $\ln(1+e^{-|z|/\temp})\in(0,\ln2]$; this is \cref{eq:comono:v}.

The second fact is the following lower bound. For every $\temp\ge0$ and every score $u$ on $\mathcal A_{+}$, we have:
\begin{equation}\label{eq:sp-limit:cert}
F_{t,\temp}(u)-F_{t,\temp}(u_{t,\temp})\ \ge\ \lam\,\KL\left(\pol_{t,\temp}\,\|\,\pol^u_t\right),
\end{equation}
which is \cref{eq:sp-support:cert} for $\temp>0$ and \cref{eq:comono:cert} for $\temp=0$.

Now fix $0\le\temp'<\temp$. Since $F_{t,\temp'}\le F_{t,\temp}$ at $u_{t,\temp}$, since $u_{t,\temp}$ minimizes $F_{t,\temp}$, and by the upper bound of \cref{eq:sp-limit:close} at $u_{t,\temp'}$, we have:
\[
F_{t,\temp'}(u_{t,\temp})-F_{t,\temp'}(u_{t,\temp'})\ \le\ F_{t,\temp}(u_{t,\temp})-F_{t,\temp'}(u_{t,\temp'})\ \le\ F_{t,\temp}(u_{t,\temp'})-F_{t,\temp'}(u_{t,\temp'})\ \le\ (\temp-\temp')\ln2.
\]
Applying \cref{eq:sp-limit:cert} at temperature $\temp'$, for $u=u_{t,\temp}$, lower bounds the left side by $\lam\KL(\pol_{t,\temp'}\|\pol_{t,\temp})$. This is the first inequality of \cref{eq:sp-stability}. In the other direction, $u_{t,\temp'}$ minimizes $F_{t,\temp'}$ and $F_{t,\temp'}\le F_{t,\temp}$ at $u_{t,\temp}$, so $F_{t,\temp'}(u_{t,\temp'})\le F_{t,\temp'}(u_{t,\temp})\le F_{t,\temp}(u_{t,\temp})$, and therefore, by the upper bound of \cref{eq:sp-limit:close} at $u_{t,\temp'}$, we have:
\[
F_{t,\temp}(u_{t,\temp'})-F_{t,\temp}(u_{t,\temp})\ =\ \left(F_{t,\temp}(u_{t,\temp'})-F_{t,\temp'}(u_{t,\temp'})\right)+\left(F_{t,\temp'}(u_{t,\temp'})-F_{t,\temp}(u_{t,\temp})\right)\ \le\ (\temp-\temp')\ln2 ;
\]
\cref{eq:sp-limit:cert} at temperature $\temp$, applied to $u=u_{t,\temp'}$, bounds the left side below by $\lam\KL(\pol_{t,\temp}\|\pol_{t,\temp'})$, which is the second inequality of \cref{eq:sp-stability}.

For probability vectors $p,q$ on a finite set, we have $\sum_\act|p(\act)-q(\act)|\le\sqrt{2\,\KL(p\|q)}$ with $\KL$ from Pinsker's inequality \cite{coverinformationtheory}. Applied to the first inequality of \cref{eq:sp-stability}, it gives the $\ell^1$ bound. For the state-wise value functions, since $\sum_\act p(\act)=\sum_\act q(\act)=1$, the same constant may be subtracted from $\Q(\state,\cdot)$ at every action without changing $\sum_\act(p(\act)-q(\act))\Q(\state,\act)$. Subtracting $\tfrac12\left(\max_{\act'}\Q(\state,\act')+\min_{\act'}\Q(\state,\act')\right)$ leaves a function of absolute value at most $\tfrac12\osc(\Q(\state,\cdot))$, so we have:
\[
\left|\val{p}(\state)-\val{q}(\state)\right|\ \le\ \tfrac12\,\osc\left(\Q(\state,\cdot)\right)\sum_\act|p(\act)-q(\act)|.
\]
Inserting the $\ell^1$ bound gives $\osc(\Q(\state,\cdot))\sqrt{(\temp-\temp')\ln2/(2\lam)}$.

In a finite-horizon MDP every decision state is non-degenerate under $\base$ by the standing assumption, and the state losses at a decision state $\state$ are the contextual-bandit losses with $\Qb(\state,\cdot)$ in place of $\Q(\state,\cdot)$, so the $\ell^1$ bound holds at every decision state. Lemma~\ref{lem:pdl}, with $\pol=\pi^{\temp}_{\mathrm{rw}}$ and $\pol'=\pirw$, gives
\begin{align*}
    \left|\val{\pi^{\temp}_{\mathrm{rw}}}(\initlaw)-\val{\pirw}(\initlaw)\right|\ &\le\ \tfrac12\sum_{\state}\occ_{\pi^{\temp}_{\mathrm{rw}},\initlaw}(\state)\,\left\|\pi^{\temp}_{\mathrm{rw}}(\cdot\mid\state)-\pirw(\cdot\mid\state)\right\|_1\\
    &\le\ \tfrac12\,\horizon\sqrt{2\temp\ln2/\lam}=\horizon\sqrt{\temp\ln2/(2\lam)},
\end{align*}
by the $\ell^1$ bound with $\temp'=0$ and $\sum_\state\occ_{\pi^{\temp}_{\mathrm{rw}},\initlaw}(\state)\le\horizon$ from the same lemma.
\end{proof}

In Lemma~\ref{lem:sp-dpo}, \cref{eq:sp-approx} says that the hinge objectives are the limit, as $\beta_{\mathrm{DPO}}\to\infty$ with $\lam$ and $\marg$ fixed, of $\beta_{\mathrm{DPO}}^{-1}$ times the DPO objective with coefficient $\beta_{\mathrm{DPO}}$, preference margin $\beta_{\mathrm{DPO}}\marg$ and anchor weight $\beta_{\mathrm{DPO}}\lam$, and Lemma~\ref{lem:sp-limit} says that the DPO policies converge to the hinge policies along this path, at rate $\sqrt{2\ln2/(\lam\beta_{\mathrm{DPO}})}$ in $\ell^1$ . Letting $\beta_{\mathrm{DPO}}$ grow with the preference margin and the anchor weight held fixed is a different limit: it sends $\marg$ and $\lam$ to zero, and plain DPO with reference $\base$ then returns to $\base$ (Lemma~\ref{lem:sp-bt}). 

In lemmas~\ref{lem:sp-bt} and~\ref{lem:sp-binary}, we show that without the margin plain DPO improves $\base$ but need not improve $\scond$, and that the SFT anchor weight $\lam$ alone not change this. However, in the next section, we shall show that the margin-augmented, success-anchored DPO objective satisfies the four guarantees of Theorems~\ref{thm:rw-base}, \ref{thm:beat-sft}, \ref{thm:comono} and~\ref{thm:rankings} at every temperature, with margin thresholds \cref{eq:sp-sft-margin,eq:sp-rank-margin} that grow linearly in the temperature, and that the hinge is the natural endpoint of this family. This will imply that, having a margin and SFT term gives provable leverage to the vanilla DPO objective.

\subsection{The four main results at positive temperature}
\label{app:softplus:main}

We now prove the counterparts of Theorems~\ref{thm:rw-base}, \ref{thm:beat-sft}, \ref{thm:comono} and~\ref{thm:rankings} for the softplus losses at a fixed temperature $\temp>0$. Throughout this subsection, for $\lam>0$, $\pol_\star$ with $\star\in\{\mathrm{rw},\mathrm{unw}\}$ is the optimizer of Lemma~\ref{lem:sp-support}, $u$ is its score as defined in item~2 of that lemma, and $A_\act$, $B_\act$ are the sums \cref{eq:sp-AB} at that score, so that \cref{eq:sp-foc} and \cref{eq:sp-mass} hold at every $\act\in\mathcal A_{+}$; the case $\lam=0$, to which the lemma does not apply, is treated where it occurs, in Theorems~\ref{thm:sp-base} and~\ref{thm:sp-comono}. All the proofs are organized around the stationarity condition \cref{eq:sp-foc}. Comparing it at two actions gives the order in which the optimizer ranks them (Theorems~\ref{thm:sp-base} and~\ref{thm:sp-rankings}); bounding $A_\act$ and $B_\act$ at a large margin gives the distance between the optimizer and the hinge policy $\pol^{\lam}$ of Lemma~\ref{lem:hinge-optimizer} (Lemma~\ref{lem:sp-saturation} and Theorem~\ref{thm:sp-sft}); and differentiating it along the path from the \unweighted\ to the \reweighted\ loss gives the comparison of their values (Theorem~\ref{thm:sp-comono}).

\paragraph{Softplus \reweighted loss always improves over rollouts:} In \cref{thm:rw-base}, we have proved that, given any rollout policy $\base$, the optimizer of the \reweighted loss always improves over it in value function for the hinge-based contrastive loss. Here we shall show that, the same holds for the softplus-based contrastive losses too, at any given temperature.

\begin{theorem}[\reweighted\ softplus always improves over the base]\label{thm:sp-base}
Fix $\temp>0$ and $\marg\ge0$, and a finite-horizon MDP with base policy $\base$ whose decision states are non-degenerate.
\begin{enumerate}
\item Let $\lam>0$, and let $\pirw^{\temp}$ be the state-wise minimizer of the \reweighted\ softplus objective. Then $\val{\pirw^{\temp}}(\state)\ge\val{\base}(\state)$ for every state $\state$.
\item Let $\lam=0$, and assume that $0<\Qb(\state,\act)<1$ for every decision state $\state$ and every action $\act$. Then the \reweighted\ softplus state loss has a unique minimizer at every decision state, and it has full support; writing $\pirw^{\temp}$ for the state-wise minimizer, $\val{\pirw^{\temp}}(\state)\ge\val{\base}(\state)$ for every state $\state$.
\end{enumerate}
In both cases, at every decision state $\state$, $\act\mapsto\pirw^{\temp}(\act\mid\state)/\base(\act\mid\state)$ is a strictly increasing function of $\Qb(\state,\act)$, and $\sum_\act\pirw^{\temp}(\act\mid\state)\Qb(\state,\act)\ge\val{\base}(\state)$, strictly whenever $\Qb(\state,\cdot)$ is non-constant.
\end{theorem}

\begin{proof}
Fix a decision state $\state$. The state loss at $\state$ is the contextual-bandit loss \cref{eq:sp-loss} with $\Qb(\state,\cdot)$ in place of $\Q(\state,\cdot)$, so throughout the proof $\Q=\Qb(\state,\cdot)$, $\mathcal A_{+}=\{\act:\Qb(\state,\act)>0\}$, and $\scond$, $\fcond$ are given by \cref{eq:cond} with this $\Q$. For $\lam>0$, Lemma~\ref{lem:sp-support} gives: $\pirw^{\temp}(\cdot\mid\state)$ is supported exactly on $\mathcal A_{+}$, its score $u=\ln\pirw^\temp-\ln\base(\cdot\mid\state)$ on $\mathcal A_{+}$ is unique up to an additive constant, and the stationarity condition in \cref{eq:sp-foc} holds at every $\act\in\mathcal A_{+}$. For $\lam=0$ under the assumption of item~2, $\mathcal A_{+}=\actset$ and both $\scond(\cdot\mid\state)$ and $\fcond(\cdot\mid\state)$ are positive at every action. Using the notations from \cref{eq:sp-support:param}, we can say that every policy with full support is $\pol^u_1$ for a score $u$ on $\actset$, unique up to an additive constant, by \cref{eq:sp-support:param} with $\mathcal A_{+}=\actset$, and the state loss at $\pol^u_1$ is exactly $C_\temp(u)$ of \cref{eq:sp-support:F}: the success-conditioning term is absent and the softplus arguments are $\marg+u(\act')-u(\act)$. A policy without full support has infinite loss, because the pair with that action as the success and any action of positive mass as the failure has an infinite softplus argument and positive weight; so the minimization is over scores. 

We note that, $C_\temp$ is unchanged when a constant is added to $u$, so we minimize over the zero-sum scores. There it tends to $+\infty$ as $\max_\act|u(\act)|\to\infty$. To see this, we make the following observation. Suppose $\bar u=\max_\act u(\act)$ and $\underline u=\min_\act u(\act)$. The pair formed by an action attaining $\underline u$ as the success and an action attaining $\bar u$ as the failure has weight at least $\min_\act\scond(\act\mid\state)\min_\act\fcond(\act\mid\state)>0$. Its softplus argument is $\marg+\bar u-\underline u$, and the softplus output is at least its argument, so $C_\temp(u)\ge\min_\act\scond(\act\mid\state)\min_\act\fcond(\act\mid\state)\,(\marg+\bar u-\underline u)$, while $\bar u-\underline u\ge\max_\act|u(\act)|$ on the zero-sum scores. Thus, the the contrastive term diverges wrt $\max_\act|u(\act)|$.

As in the proof of Lemma~\ref{lem:sp-support}, the sublevel sets of the continuous function $C_\temp$ on the zero-sum scores are compact and the minimum is attained. We now argue that it is attained once.The matrix of second derivatives of $C_\temp$ is the graph Laplacian $K$ computed in the proof of Theorem~\ref{thm:comono} for the smoothed objective, with the smoothing parameter equal to $\temp$. Its edge weights $k_{\act\act'}$ are positive when every pair weight $\scond(\act\mid\state)\fcond(\act'\mid\state)$ is, so that $v^\top Kv=\sum_{\{\act,\act'\}}k_{\act\act'}(v(\act)-v(\act'))^2>0$ for every nonconstant $v$. Hence $C_\temp$ is strictly convex on the zero-sum scores and two distinct minimizers would give a smaller value at their midpoint. 

Let $u$ be the minimizer and $\pirw^\temp\coloneqq\pol^u_1$ is a policy with full support. The gradient of $C_\temp$ is the first two sums of \cref{eq:sp-support:grad}; it is a zero-sum vector, and at the minimizer over zero-sum scores it is orthogonal to every zero-sum vector, hence zero. Using the notation from \cref{eq:sp-AB} this can be written as $\fcond(\act\mid\state)A_\act-\scond(\act\mid\state)B_\act=0$ for every $\act$, which is \cref{eq:sp-foc} with $\lam=0$. This proves the first two claims of item~2, and in both cases we now have, for every $\act\in\mathcal A_{+}$, after dividing \cref{eq:sp-foc} by $\scond(\act\mid\state)>0$, we get:
\begin{equation}\label{eq:sp-base:ratio}
\lam\left(\frac{\pirw^\temp(\act)}{\scond(\act\mid\state)}-1\right)=B_\act-\frac{\fcond(\act\mid\state)}{\scond(\act\mid\state)}\,A_\act ,
\end{equation}
where the left side is $0$ when $\lam=0$.

Now, we hall show that $\act\mapsto\pirw^\temp(\act)/\base(\act\mid\state)$ is a strictly increasing function of $\Q(\state,\act)$ on $\actset$. Let $\act,\act'\in\mathcal A_{+}$ with $u(\act)\le u(\act')$. Since $\sigma$ is increasing, every term of $A_\act$ is at most the corresponding term of $A_{\act'}$, and every term of $B_\act$ is at least the corresponding term of $B_{\act'}$; moreover $A_\act>0$, as $\sigma>0$ and $\scond>0$ on $\mathcal A_{+}$. So
\begin{equation}\label{eq:sp-base:monotone}
0<A_\act\le A_{\act'}\,\text{and}\, B_\act\ge B_{\act'},\,\text{with }A_\act<A_{\act'}\text{ if }u(\act)<u(\act'),
\end{equation}
the strictness because $\sigma$ is strictly increasing and every weight in $A_\act$ is positive. By \cref{eq:cond}, we have:
\begin{equation}\label{eq:sp-base:kappa}
\frac{\fcond(\act\mid\state)}{\scond(\act\mid\state)}=\frac{\val{\base}(\state)}{1-\val{\base}(\state)}\cdot\frac{1-\Q(\state,\act)}{\Q(\state,\act)},
\end{equation}
which is a strictly decreasing function of $\Q(\state,\act)$. Suppose now that $\Q(\state,\act)>\Q(\state,\act')$ and $u(\act)\le u(\act')$. Then \cref{eq:sp-base:kappa} is smaller at $\act$ than at $\act'$, and with \cref{eq:sp-base:monotone},
\[
\frac{\fcond(\act\mid\state)}{\scond(\act\mid\state)}A_\act\ \le\ \frac{\fcond(\act\mid\state)}{\scond(\act\mid\state)}A_{\act'}\ <\ \frac{\fcond(\act'\mid\state)}{\scond(\act'\mid\state)}A_{\act'},
\]
the last inequality being strict because $A_{\act'}>0$; together with $B_\act\ge B_{\act'}$, the right side of \cref{eq:sp-base:ratio} is strictly larger at $\act$ than at $\act'$. If $\lam=0$ both right sides are $0$, a contradiction. If $\lam>0$ we get
\begin{equation}\label{eq:sp-base:key}
\frac{\pirw^\temp(\act)}{\scond(\act\mid\state)}\ >\ \frac{\pirw^\temp(\act')}{\scond(\act'\mid\state)}\,\text{whenever }\Q(\state,\act)>\Q(\state,\act')\text{ and }u(\act)\le u(\act').
\end{equation}
We make a note here that was derived from \cref{eq:sp-foc} and the monotonicity of $\sigma$ alone, so it holds for the \unweighted\ loss as well, with $u$ the score of that loss; we shall infact use it in this form in the proof of Theorem~\ref{thm:sp-rankings}. 

For the optimizer of the \reweighted\ loss, we have: $\frac{\pirw^\temp(\act)}{\scond(\act\mid\state)}=\frac{\pirw^\temp(\act)}{\base(\act\mid\state)}\cdot\frac{\val{\base}(\state)}{\Q(\state,\act)}$ by \cref{eq:cond}, and $u(\act)-u(\act')=\ln\frac{\pirw^\temp(\act)/\base(\act\mid\state)}{\pirw^\temp(\act')/\base(\act'\mid\state)}$. From \cref{eq:sp-base:key} we can conclude that $u(\act)-u(\act')>\ln\frac{\Q(\state,\act)}{\Q(\state,\act')}>0$, contradicting our assumption that $u(\act)\le u(\act')$. Hence, in both cases, $\Q(\state,\act)>\Q(\state,\act')$ forces $u(\act)>u(\act')$, that is, $\frac{\pirw^\temp(\act)}{\base(\act\mid\state)}>\frac{\pirw^\temp(\act')}{\base(\act'\mid\state)}$. If instead $\Q(\state,\act)=\Q(\state,\act')$, suppose $u(\act)<u(\act')$. Then \cref{eq:sp-base:kappa} takes the same value at $\act$ and at $\act'$, and by \cref{eq:sp-base:monotone} the right side of \cref{eq:sp-base:ratio} is at least as large at $\act$ as at $\act'$. For $\lam=0$ it is strictly larger, because the common value of \cref{eq:sp-base:kappa} is then positive, as $\Q<1$, and $A_\act<A_{\act'}$; this contradicts that both right sides are $0$. For $\lam>0$ it gives $\frac{\pirw^\temp(\act)}{\scond(\act\mid\state)}\ge\frac{\pirw^\temp(\act')}{\scond(\act'\mid\state)}$, hence $u(\act)-u(\act')\ge\ln\frac{\Q(\state,\act)}{\Q(\state,\act')}=0$, again a contradiction. By symmetry $u(\act)=u(\act')$, and the two ratios are equal. Finally, off $\mathcal A_{+}$ the ratio $\pirw^\temp(\act)/\base(\act\mid\state)$ is $0$ and so is $\Q(\state,\act)$, while on $\mathcal A_{+}$ the ratio is positive. This proves our claim.

We finally compute the difference in the value functions. Since $\sum_\act\base(\act\mid\state)\frac{\pirw^\temp(\act)}{\base(\act\mid\state)}=1=\sum_\act\base(\act\mid\state)$, we have:
\begin{align*}
    \sum_{\act}\left(\pirw^\temp(\act)-\base(\act\mid\state)\right)\Q(\state,\act)&=\sum_\act\base(\act\mid\state)\left(\frac{\pirw^\temp(\act)}{\base(\act\mid\state)}-1\right)\Q(\state,\act)\\
    &=\Cov_{\act\sim\base(\cdot\mid\state)}\left(\frac{\pirw^\temp(\act)}{\base(\act\mid\state)},\Q(\state,\act)\right),
\end{align*}
The above expression and Lemma~\ref{lem:cov} implies that this covariance is nonnegative, and positive when $\Q(\state,\cdot)$ is non-constant. For the difference in the value functions, we apply Performance difference Lemma(\cref{lem:pdl}) with the initial law concentrated at $\state$, $\pol=\pirw^{\temp}$ and $\pol'=\base$ to get:
\[
\val{\pirw^{\temp}}(\state)-\val{\base}(\state)=\sum_{\state'}\occ_{\pirw^{\temp},\state}(\state')\sum_{\act}\left(\pirw^{\temp}(\act\mid\state')-\base(\act\mid\state')\right)\Qb(\state',\act)\ \ge\ 0,
\]
because each inner sum is the covariance (whose non-negativity we just proved) at the decision state $\state'$ and the occupancies $\occ_{\pirw^{\temp},\state}(\state')$ are nonnegative.
\end{proof}

\paragraph{Large-margin improvement of softplus-contrastive losses over success conditioning.}
The next three results concern the large-margin region, and they use the quantities of Lemmas~\ref{lem:hinge-optimizer} and~\ref{lem:uniform-margin} throughout; we recall them here once again. For the fixed non-degenerate state $\state$ and with $\mathcal A_{+}=\{\act:\Q(\state,\act)>0\}$. The SFT anchor threshold from \cref{eq:lampos} is
\[
\lam^*=\max_{\act\in\mathcal A_{+}}\left(\frac{\fcond(\act\mid\state)}{\scond(\act\mid\state)}-\fcond(\mathcal A_{+}\mid\state)\right)\ \ge\ 0,
\]
and for $\lam>\lam^*$, we define the policy $\pol^\lam$ from \cref{eq:explicit-policy} as
\[
\pol^{\lam}(\act\mid\state)=\begin{cases}
    \scond(\act\mid\state)+\frac{\fcond(\mathcal A_{+}\mid\state)\,\scond(\act\mid\state)-\fcond(\act\mid\state)}{\lam}\quad&\text{for }\act\in\mathcal A_{+},\\
    0\quad&\text{for }\act\notin\mathcal A_{+},
\end{cases},
\]
which is a probability vector, positive on $\mathcal A_{+}$, which does not depend on the margin. The thresholds from \cref{eq:thresholds} are defined as:
\begin{align*}
\bar\lambda(\state)&=\max\{2\lam^*,2\},\,
\bar\delta_{\mathrm{rw}}(\state)=\ln\frac{\max_{\act\in\mathcal A_{+}}\Q(\state,\act)}{\min_{\act\in\mathcal A_{+}}\Q(\state,\act)}+\ln3,\\
\bar\delta_{\mathrm{unw}}(\state)&=\ln\frac{\max_{\act\in\mathcal A_{+}}\base(\act\mid\state)\,\Q(\state,\act)}{\min_{\act\in\mathcal A_{+}}\base(\act\mid\state)\,\Q(\state,\act)}+\ln3.
\end{align*}
Lemma~\ref{lem:uniform-margin} says that, for $\lam\ge\bar\lambda(\state)$, the hinge policies are $\pirw(\cdot\mid\state)=\pol^{\lam}(\cdot\mid\state)$ when $\marg\ge\bar\delta_{\mathrm{rw}}(\state)$ and $\piunw(\cdot\mid\state)=\pol^{\lam}(\cdot\mid\state)$ when $\marg\ge\bar\delta_{\mathrm{unw}}(\state)$. The softplus results use the same margin thresholds, but the condition on $\lam$ is strengthened to
\begin{equation}\label{eq:sp-threshold}
\lam\ \ge\ \max\left\{2\max_{\act\in\mathcal A_{+}}\left(\frac{\fcond(\act\mid\state)}{\scond(\act\mid\state)}-\fcond(\act\mid\state)\right),\ 2\right\}.
\end{equation}
Since $\fcond(\act\mid\state)\le\fcond(\mathcal A_{+}\mid\state)$, the inner maximum in \cref{eq:sp-threshold} is at least $\lam^*$, so \cref{eq:sp-threshold} implies $\lam\ge\bar\lambda(\state)$; and since it gives $\lam\ge2\lam^*$ and $\lam\ge2>0$, it implies $\lam>\lam^*$, so that $\pol^{\lam}$ is defined. The reason for the stronger condition is that at positive temperature the exact form \cref{eq:explicit-policy} of the hinge optimizer is replaced by the two-sided bounds \cref{eq:sp-mass}, and \cref{eq:sp-threshold} is what makes the two sides of \cref{eq:sp-mass} equal to $\tfrac12\scond(\act\mid\state)$ and $\tfrac32\scond(\act\mid\state)$ at worst.

Now we shall show that even for the softplus loss, at large $\lam, \marg$, the learnt policies $\pirw, \piunw$ gets very close to $\pol^\lam$.

\begin{lemma}[Saturation at large margin]\label{lem:sp-saturation}
Fix a non-degenerate state $\state$, $\temp>0$, $\marg\ge0$ and a sufficiently large $\lam$ satisfying the lower bound from \cref{eq:sp-threshold}. Let $\pol^{\lam}$ be the policy of Lemma~\ref{lem:hinge-optimizer} which is described above. Defining $\bar\delta_{\mathrm{rw}}(\state)$ as in \cref{eq:thresholds}, we have:
\begin{equation}\label{eq:sp-saturation}
\begin{split}
&\left\|\pirw^\temp(\cdot\mid\state)-\pol^{\lam}(\cdot\mid\state)\right\|_1\ \le\ \frac{2\,\fcond(\mathcal A_{+}\mid\state)}{\lam}\,e^{-(\marg-\bar\delta_{\mathrm{rw}}(\state))/\temp}\\
\text{and}\,\,&\left|\val{\pirw^\temp}(\state)-\val{\pol^{\lam}}(\state)\right|\ \le\ \frac1\lam\,e^{-(\marg-\bar\delta_{\mathrm{rw}}(\state))/\temp}.
\end{split}
\end{equation}
The exact same result holds for $\piunw^\temp$ as well, with the margin threshold $\bar\delta_{\mathrm{unw}}$ instead of $\bar\delta_{\mathrm{rw}}$.
\end{lemma}
\begin{proof}
By \cref{eq:sp-threshold}, the lower bound in \cref{eq:sp-mass} is at least $\scond(\act\mid\state)(1-\tfrac12)$, as $\frac{\fcond(\act\mid\state)}{\scond(\act\mid\state)}-\fcond(\act\mid\state)\le\lam/2$, and the upper bound is at most $\scond(\act\mid\state)(1+\tfrac12)$, as $\fcond(\mathcal A_{+}\mid\state)-\fcond(\act\mid\state)\le1\le\lam/2$. Hence
\begin{equation}\label{eq:sp-saturation:half}
\tfrac12\ \le\ \frac{\pirw^\temp(\act)}{\scond(\act\mid\state)}\ \le\ \tfrac32\,\text{for every }\act\in\mathcal A_{+}.
\end{equation}
We use this to bound the differences of the score $u$ of $\pirw^\temp$ from below. For the \reweighted\ loss, $u=\ln\pirw^\temp-\ln\base(\cdot\mid\state)$ on $\mathcal A_{+}$, and by \cref{eq:cond} $\ln\scond(\act\mid\state)-\ln\base(\act\mid\state)=\ln\Q(\state,\act)-\ln\val{\base}(\state)$, so for $\act,\act'\in\mathcal A_{+}$,
\[
u(\act)-u(\act')=\ln\frac{\Q(\state,\act)}{\Q(\state,\act')}+\ln\frac{\pirw^\temp(\act)/\scond(\act\mid\state)}{\pirw^\temp(\act')/\scond(\act'\mid\state)}\ \ge\ -\ln\frac{\max_{\act''\in\mathcal A_{+}}\Q(\state,\act'')}{\min_{\act''\in\mathcal A_{+}}\Q(\state,\act'')}-\ln3=-\bar\delta_{\mathrm{rw}}(\state),
\]
by \cref{eq:sp-saturation:half} and \cref{eq:thresholds}. For the \unweighted\ loss, $u=\ln\piunw^\temp$ on $\mathcal A_{+}$ and \cref{eq:sp-saturation:half} holds for $\piunw^\temp$ as well, since \cref{eq:sp-mass} does. Also, we have $\ln\scond(\act\mid\state)=\ln\base(\act\mid\state)+\ln\Q(\state,\act)-\ln\val{\base}(\state)$, so in the same way $u(\act)-u(\act')\ge-\ln\frac{\max_{\act''}\base(\act''\mid\state)\Q(\state,\act'')}{\min_{\act''}\base(\act''\mid\state)\Q(\state,\act'')}-\ln3=-\bar\delta_{\mathrm{unw}}(\state)$.

In both cases every sigmoid in \cref{eq:sp-AB} has argument at least $(\marg-\bar\delta_{\mathrm{rw}}(\state))/\temp$ (respectively, $(\marg-\bar\delta_{\mathrm{unw}}(\state))/\temp$ for the \unweighted case). Since $1-\sigma(z)=e^{-z}/(1+e^{-z})\le e^{-z}$ for every real $z$, and since the weights in \cref{eq:sp-AB} sum to $1$ in $A_\act$ and to $\fcond(\mathcal A_{+}\mid\state)$ in $B_\act$, we have:
\begin{equation}\label{eq:sp-saturation:AB}
1-e^{-(\marg-\bar\delta_{\mathrm{rw}}(\state))/\temp}\ \le\ A_\act\ \le\ 1,\,
\fcond(\mathcal A_{+}\mid\state)\left(1-e^{-(\marg-\bar\delta_{\mathrm{rw}}(\state))/\temp}\right)\ \le\ B_\act\ \le\ \fcond(\mathcal A_{+}\mid\state).
\end{equation}
The exact same expression as above holds when $\bar\delta_{\mathrm{rw}}$ is replaced by $\bar\delta_{\mathrm{unw}}$ as well. By \cref{eq:explicit-policy}, $\lam\left(\pol^{\lam}(\act\mid\state)-\scond(\act\mid\state)\right)=\scond(\act\mid\state)\fcond(\mathcal A_{+}\mid\state)-\fcond(\act\mid\state)$ on $\mathcal A_{+}$, and subtracting this from \cref{eq:sp-foc} gives
\[
\pirw^\temp(\act)-\pol^{\lam}(\act\mid\state)=\frac{\scond(\act\mid\state)\left(B_\act-\fcond(\mathcal A_{+}\mid\state)\right)-\fcond(\act\mid\state)\left(A_\act-1\right)}{\lam},
\]
whose absolute value is at most $\left(\scond(\act\mid\state)\fcond(\mathcal A_{+}\mid\state)+\fcond(\act\mid\state)\right)e^{-(\marg-\bar\delta_{\mathrm{rw}}(\state))/\temp}/\lam$ by \cref{eq:sp-saturation:AB}. Both policies are supported on $\mathcal A_{+}$, so summing over $\act\in\mathcal A_{+}$, where $\sum_\act\scond(\act\mid\state)=1$ and $\sum_\act\fcond(\act\mid\state)=\fcond(\mathcal A_{+}\mid\state)$, gives the first bound in \cref{eq:sp-saturation}. 

   The second bound follows from $\left|\val{p}(\state)-\val{q}(\state)\right|\le\tfrac12\osc\left(\Q(\state,\cdot)\right)\sum_\act|p(\act)-q(\act)|$ for probability vectors $p,q$ on $\actset$, shown in the proof of Lemma~\ref{lem:sp-limit} by subtracting the midpoint of $\max_\act\Q(\state,\act)$ and $\min_\act\Q(\state,\act)$ from $\Q(\state,\cdot)$, together with $\osc(\Q(\state,\cdot))\le1$ and $\fcond(\mathcal A_{+}\mid\state)\le1$.

      For the \unweighted\ loss the argument is the same with $\piunw^{\temp}$ in place of $\pirw^{\temp}$ and $\bar\delta_{\mathrm{unw}}(\state)$ in place of $\bar\delta_{\mathrm{rw}}(\state)$: the score bound was derived above, and the remaining steps use only \cref{eq:sp-foc} and \cref{eq:sp-saturation:AB}, which hold for either loss.
\end{proof}

\begin{theorem}[Softplus counterpart of Theorem~\ref{thm:beat-sft}]\label{thm:sp-sft}
Fix a non-degenerate context $\state$ of a contextual bandit, $\temp>0$ and $\lam$ as in \cref{eq:sp-threshold}. Then, for every $\marg\ge0$, we have:
\begin{equation}\label{eq:sp-sft}
\val{\pirw^\temp}(\state)-\val{\scond}(\state)\ \ge\ \frac1\lam\left(\frac{\val{\base}(\state)}{1-\val{\base}(\state)}\Var_{\act\sim\scond(\cdot\mid\state)}\left[\Q(\state,\act)\right]-e^{-(\marg-\bar\delta_{\mathrm{rw}}(\state))/\temp}\right).
\end{equation}
In particular, if $\Q(\state,\cdot)$ is non-constant on $\mathcal A_{+}$ and the margin $\marg$ satisfies the lower bound
\begin{equation}\label{eq:sp-sft-margin}
\marg\ \ge\ \bar\delta_{\mathrm{rw}}(\state)+\temp\ln\frac{2\left(1-\val{\base}(\state)\right)}{\val{\base}(\state)\,\Var_{\act\sim\scond(\cdot\mid\state)}\left[\Q(\state,\act)\right]},
\end{equation}
then $\val{\pirw^\temp}(\state)-\val{\scond}(\state)\ge\frac{\val{\base}(\state)}{2\lam\left(1-\val{\base}(\state)\right)}\Var_{\act\sim\scond(\cdot\mid\state)}\left[\Q(\state,\act)\right]>0$. The exact same conclusions hold for the optimer of the \unweighted loss, $\piunw^\temp$ as well, with $\bar\delta_{\mathrm{unw}}$ in place of $\bar\delta_{\mathrm{rw}}$.
\end{theorem}
\begin{proof}
By \cref{eq:sp-threshold}, $\lam\ge\bar\lambda(\state)$, where $\bar\lambda(\state)$ is as defined in the preceding discussion. For any $\marg'\ge\bar\delta_{\mathrm{rw}}(\state)$, Lemma~\ref{lem:uniform-margin} gives that the hinge policy $\pirw(\cdot\mid\state)$ at the parameters $(\lam,\marg')$ equals $\pol^{\lam}(\cdot\mid\state)$, and Theorem~\ref{thm:beat-sft} gives $\val{\pirw}(\state)-\val{\scond}(\state)\ge\frac{\val{\base}(\state)}{\lam\left(1-\val{\base}(\state)\right)}\Var_{\act\sim\scond(\cdot\mid\state)}\left[\Q(\state,\act)\right]$. Since $\pol^{\lam}$ does not depend on the margin, we have, for every $\lam$ as in \cref{eq:sp-threshold},
\[
\val{\pol^{\lam}}(\state)-\val{\scond}(\state)\ \ge\ \frac{\val{\base}(\state)}{\lam\left(1-\val{\base}(\state)\right)}\Var_{\act\sim\scond(\cdot\mid\state)}\left[\Q(\state,\act)\right].
\]
Subtracting the second bound of \cref{eq:sp-saturation} gives \cref{eq:sp-sft}. 

If $\marg$ satisfies the bound in \cref{eq:sp-sft-margin}, we have that:
$$e^{-(\marg-\bar\delta_{\mathrm{rw}}(\state))/\temp}\le\frac{\val{\base}(\state)}{2(1-\val{\base}(\state))}\Var_{\act\sim\scond(\cdot\mid\state)}[\Q(\state,\act)],$$ 
which gives the second claim. The variance is positive because $\scond(\cdot\mid\state)$ is positive on $\mathcal A_{+}$, where $\Q(\state,\cdot)$ is non-constant.  The same argument, with the \unweighted\ clauses of Lemma~\ref{lem:uniform-margin} and Theorem~\ref{thm:beat-sft}, the \unweighted\ case of Lemma~\ref{lem:sp-saturation}, and $\bar\delta_{\mathrm{unw}}(\state)$ in place of $\bar\delta_{\mathrm{rw}}(\state)$, gives the conclusions for $\piunw^\temp$.
\end{proof}

We now make three remarks on the statement. First, unlike the hinge policies, the two softplus policies need not coincide at any finite margin; both are within an exponentially small distance of the policy $\pol^{\lam}$ of Lemma~\ref{lem:hinge-optimizer}, to which both hinge policies reduce by Lemma~\ref{lem:uniform-margin}, and the gain of Theorem~\ref{thm:beat-sft} is recovered up to the same exponential error, uniformly in $\lam$. Second, as $\temp\downarrow0$ at a fixed $\marg>\bar\delta_{\mathrm{rw}}(\state)$ (respectively, $\bar\delta_{\mathrm{unw}}(\state)$) the error vanishes and \cref{eq:sp-sft} becomes the bound of Theorem~\ref{thm:beat-sft}. Third, when $\Q(\state,\cdot)$ takes only the values $0$ and $1$, as for a deterministic verifier of complete responses, $\fcond$ vanishes on $\mathcal A_{+}$, so $B_\act=0$ and $\fcond(\act\mid\state)=0$ for every $\act\in\mathcal A_{+}$, and \cref{eq:sp-foc} reads $\lam\left(\pirw^\temp(\act)-\scond(\act\mid\state)\right)=0$, which means both softplus policies equal $\scond$ at every $\lam>0$, every $\marg\ge0$ and every temperature, exactly as the hinge policies do.

Now, we focus on the retrieval properties of the optimizers of the softplus contrastive losses, which is similar to the statement of \cref{thm:rankings} for the hinge loss.

\begin{theorem}[Retrieval guarantees of softplus-contrastive losses]\label{thm:sp-rankings}
Consider a contextual bandit with a non-degenerate context $\state$, and fix $\temp>0$ and $\marg\ge0$.
\begin{enumerate}
\item Let $\lam>0$. Then, whenever $\Q(\state,\act)>\Q(\state,\act')$, we have:
\begin{align}
\scond(\act\mid\state)\ge\scond(\act'\mid\state)&\ \Longrightarrow\ \piunw^{\temp}(\act\mid\state)>\piunw^{\temp}(\act'\mid\state),\label{eq:sp-rank:unw}\\
\base(\act\mid\state)\ge\base(\act'\mid\state)&\ \Longrightarrow\ \pirw^{\temp}(\act\mid\state)>\pirw^{\temp}(\act'\mid\state).\label{eq:sp-rank:rw}
\end{align}
In particular, $\Inv(\piunw^{\temp})\subseteq\Inv(\scond)\subseteq\Inv(\base)$ and $\Inv(\pirw^{\temp})\subseteq\Inv(\base)$.
\item Let $\lam$ satisfy \cref{eq:sp-threshold}, and let the margin $\marg$ satisfy the lower bound
\begin{equation}\label{eq:sp-rank-margin}
\marg\ \ge\ \bar\delta_{\mathrm{rw}}(\state)+\temp\ln\frac{4\left(1-\val{\base}(\state)\right)}{\min\left\{\base(\act'\mid\state)\left(\Q(\state,\act)-\Q(\state,\act')\right)\right\}},
\end{equation}
where the minimum runs over the pairs $\act,\act'\in\mathcal A_{+}$ with $\Q(\state,\act)>\Q(\state,\act')$ and $\scond(\act\mid\state)\ge\scond(\act'\mid\state)$ (no condition if there is no such pair). Then, whenever $\Q(\state,\act)>\Q(\state,\act')$, we have:
\begin{equation}\label{eq:sp-rank:sft}
\scond(\act\mid\state)\ge\scond(\act'\mid\state)\ \Longrightarrow\ \pirw^{\temp}(\act\mid\state)>\pirw^{\temp}(\act'\mid\state).
\end{equation}
In particular, $\Inv(\pirw^{\temp})\subseteq\Inv(\scond)$. The exact same conclusions hold for $\piunw^{\temp}$ as well, with $\bar\delta_{\mathrm{unw}}(\state)$ in place of $\bar\delta_{\mathrm{rw}}(\state)$ in \cref{eq:sp-rank-margin}.
\end{enumerate}
Consequently, for every swap-monotone metric and for arbitrary tie-breaking in both lists, every list sorted by $\piunw^{\temp}$ scores at least as well as every list sorted by $\scond$, and every list sorted by $\pirw^{\temp}$ scores at least as well as every list sorted by $\base$ (item~1); under the conditions of item~2, every list sorted by $\pirw^{\temp}$ (respectively, by $\piunw^{\temp}$) scores at least as well as every list sorted by $\scond$.
\end{theorem}
\begin{proof}
In this proof we abbreviate $\pirw^{\temp}(\act\mid\state)$ to $\pirw^{\temp}(\act)$ and $\piunw^{\temp}(\act\mid\state)$ to $\piunw^{\temp}(\act)$. If $\act\in\mathcal A_{+}$ and $\act'\notin\mathcal A_{+}$, then $\pirw^{\temp}(\act)>0=\pirw^{\temp}(\act')$ and $\piunw^{\temp}(\act)>0=\piunw^{\temp}(\act')$ by Lemma~\ref{lem:sp-support}, so every implication in the theorem holds for such pairs; and a pair with $\Q(\state,\act)>\Q(\state,\act')$ always has $\act\in\mathcal A_{+}$. It therefore remains to treat $\act,\act'\in\mathcal A_{+}$.

We start by proving the first part. Let $\act,\act'\in\mathcal A_{+}$ with $\Q(\state,\act)>\Q(\state,\act')$. For the \reweighted\ loss, Theorem~\ref{thm:sp-base} gives $\frac{\pirw^{\temp}(\act)}{\base(\act\mid\state)}>\frac{\pirw^{\temp}(\act')}{\base(\act'\mid\state)}$, and multiplying by $\base(\act\mid\state)\ge\base(\act'\mid\state)$ gives \cref{eq:sp-rank:rw}. For the \unweighted\ loss, the score is $u=\ln\piunw^{\temp}$ on $\mathcal A_{+}$. Suppose that $\scond(\act\mid\state)\ge\scond(\act'\mid\state)$ but $\piunw^{\temp}(\act)\le\piunw^{\temp}(\act')$, that is, $u(\act)\le u(\act')$. Then \cref{eq:sp-base:key}, which holds for the \unweighted\ loss with $\piunw^{\temp}$ in place of $\pirw^{\temp}$ and $u=\ln\piunw^{\temp}$ gives $\frac{\piunw^{\temp}(\act)}{\scond(\act\mid\state)}>\frac{\piunw^{\temp}(\act')}{\scond(\act'\mid\state)}$, hence $\piunw^{\temp}(\act)>\piunw^{\temp}(\act')\frac{\scond(\act\mid\state)}{\scond(\act'\mid\state)}\ge\piunw^{\temp}(\act')$, a contradiction. This proves \cref{eq:sp-rank:unw}.

For the containments of the inversion sets, let $(\act,\act')$ be an inversion of $\piunw^{\temp}$, that is, $\Q(\state,\act)>\Q(\state,\act')$ and $\piunw^{\temp}(\act)<\piunw^{\temp}(\act')$. By \cref{eq:sp-rank:unw}, $\scond(\act\mid\state)<\scond(\act'\mid\state)$, so $(\act,\act')$ is an inversion of $\scond$. In the same way, by \cref{eq:sp-rank:rw}, an inversion of $\pirw^{\temp}$ is an inversion of $\base$. Finally, if $(\act,\act')$ is an inversion of $\scond$, then $\base(\act\mid\state)\Q(\state,\act)<\base(\act'\mid\state)\Q(\state,\act')$ by \cref{eq:cond}, so $\base(\act\mid\state)<\base(\act'\mid\state)\frac{\Q(\state,\act')}{\Q(\state,\act)}<\base(\act'\mid\state)$, and $(\act,\act')$ is an inversion of $\base$.

Now we proceed to the proof of the second part of the theorem. Let $\act,\act'\in\mathcal A_{+}$ with $\Q(\state,\act)>\Q(\state,\act')$ and $\scond(\act\mid\state)\ge\scond(\act'\mid\state)$. We first compare the two coordinates of the policy $\pol^{\lam}$, which does not depend on the loss. By \cref{eq:explicit-policy}, we have:
\[
\pol^{\lam}(\act\mid\state)=\scond(\act\mid\state)\left(1+\frac1\lam\left(\fcond(\mathcal A_{+}\mid\state)-\frac{\fcond(\act\mid\state)}{\scond(\act\mid\state)}\right)\right)\quad\text{for }\act\in\mathcal A_{+},
\]
In the above expression, the bracket is positive, because $\frac{\fcond(\act\mid\state)}{\scond(\act\mid\state)}-\fcond(\mathcal A_{+}\mid\state)\le\lam^*<\lam$ by \cref{eq:lampos}. Subtracting the two values and using \cref{eq:sp-base:kappa} for the ratios $\fcond/\scond$, we have:
\begin{align*}
\pol^{\lam}(\act\mid\state)-\pol^{\lam}(\act'\mid\state)
&=\left(\scond(\act\mid\state)-\scond(\act'\mid\state)\right)\left(1+\frac1\lam\left(\fcond(\mathcal A_{+}\mid\state)-\frac{\fcond(\act\mid\state)}{\scond(\act\mid\state)}\right)\right)\\
&\quad+\frac{\scond(\act'\mid\state)}{\lam}\left(\frac{\fcond(\act'\mid\state)}{\scond(\act'\mid\state)}-\frac{\fcond(\act\mid\state)}{\scond(\act\mid\state)}\right)\\
&\ge\frac{\scond(\act'\mid\state)}{\lam}\cdot\frac{\val{\base}(\state)}{1-\val{\base}(\state)}\left(\frac1{\Q(\state,\act')}-\frac1{\Q(\state,\act)}\right)\\
&=\frac{\base(\act'\mid\state)\left(\Q(\state,\act)-\Q(\state,\act')\right)}{\lam\left(1-\val{\base}(\state)\right)\Q(\state,\act)}
\ \ge\ \frac{\base(\act'\mid\state)\left(\Q(\state,\act)-\Q(\state,\act')\right)}{\lam\left(1-\val{\base}(\state)\right)},
\end{align*}
where the first term was dropped because it is nonnegative, $\scond(\act'\mid\state)=\base(\act'\mid\state)\Q(\state,\act')/\val{\base}(\state)$ by \cref{eq:cond}, and $\Q(\state,\act)\le1$.

Now, if $\marg$ satisfies the bound in \cref{eq:sp-rank-margin}, we have that:
\[
e^{-(\marg-\bar\delta_{\mathrm{rw}}(\state))/\temp}\ \le\ \frac{\base(\act'\mid\state)\left(\Q(\state,\act)-\Q(\state,\act')\right)}{4\left(1-\val{\base}(\state)\right)},
\]
so the first bound of \cref{eq:sp-saturation}, with $\fcond(\mathcal A_{+}\mid\state)\le1$, gives
\begin{align*}
    &\left|\pirw^{\temp}(\act)-\pol^{\lam}(\act\mid\state)\right|+\left|\pirw^{\temp}(\act')-\pol^{\lam}(\act'\mid\state)\right|\\
    \le&\ \left\|\pirw^{\temp}(\cdot\mid\state)-\pol^{\lam}(\cdot\mid\state)\right\|_1\le\ \frac{\base(\act'\mid\state)\left(\Q(\state,\act)-\Q(\state,\act')\right)}{2\lam\left(1-\val{\base}(\state)\right)}.
\end{align*}
Therefore, we have:
\begin{align*}
    \pirw^{\temp}(\act)-\pirw^{\temp}(\act')\ &\ge\ \pol^{\lam}(\act\mid\state)-\pol^{\lam}(\act'\mid\state)-\frac{\base(\act'\mid\state)\left(\Q(\state,\act)-\Q(\state,\act')\right)}{2\lam\left(1-\val{\base}(\state)\right)}\\
    &\ge\ \frac{\base(\act'\mid\state)\left(\Q(\state,\act)-\Q(\state,\act')\right)}{2\lam\left(1-\val{\base}(\state)\right)}\ >\ 0,
\end{align*}
which is \cref{eq:sp-rank:sft}. The containment $\Inv(\pirw^{\temp})\subseteq\Inv(\scond)$ follows from \cref{eq:sp-rank:sft} exactly as in item~1. For the \unweighted\ loss the argument is the same with $\piunw^{\temp}$ in place of $\pirw^{\temp}$ and $\bar\delta_{\mathrm{unw}}(\state)$ in place of $\bar\delta_{\mathrm{rw}}(\state)$: the comparison of the two coordinates of $\pol^{\lam}$ does not involve the loss, and the \unweighted\ case of Lemma~\ref{lem:sp-saturation} supplies the saturation bound.

 For the argument about the swap monotone metrics, we recall from Section~\ref{sec:retrieval} that a list sorted by a policy orders the actions by non-increasing probability under that policy, ties broken arbitrarily, and that a swap-monotone metric is a function of the sequence of qualities read along the list which does not decrease when two adjacent entries are swapped so that the one of larger quality moves up; by the first property it is unchanged when two entries of equal quality are swapped. 
 
 We shall show the following: if $\pol_1$ and $\pol_2$ are two policies such that, whenever $\Q(\state,x)>\Q(\state,y)$, $\pol_2(x\mid\state)\ge\pol_2(y\mid\state)$ implies $\pol_1(x\mid\state)>\pol_1(y\mid\state)$, then every list sorted by $\pol_1$ scores at least as well as every list sorted by $\pol_2$ under every swap-monotone metric. This hypothesis is exactly \cref{eq:sp-rank:unw} for $(\pol_1,\pol_2)=(\piunw^{\temp},\scond)$, \cref{eq:sp-rank:rw} for $(\pol_1,\pol_2)=(\pirw^{\temp},\base)$, and \cref{eq:sp-rank:sft} and its \unweighted\ version for $(\pol_1,\pol_2)=(\pirw^{\temp},\scond)$ and $(\piunw^{\temp},\scond)$, so this proves the last claim of the theorem.

Let $\ell_1$ be a list sorted by $\pol_1$ and $\ell_2$ a list sorted by $\pol_2$. Take the first action of $\ell_1$ and bring it to the front of $\ell_2$ by adjacent swaps, then the second action of $\ell_1$ to the front of what remains, and so on, until $\ell_2$ has become $\ell_1$; the actions not yet moved keep their relative order from $\ell_2$. Every swap moves an action $x$ up past an action $y$ such that $x$ precedes $y$ in $\ell_1$ and $y$ preceded $x$ in $\ell_2$. If $\Q(\state,x)\ge\Q(\state,y)$, the metric does not decrease. The case $\Q(\state,x)<\Q(\state,y)$ does not occur: $y$ preceding $x$ in $\ell_2$ gives $\pol_2(y\mid\state)\ge\pol_2(x\mid\state)$, so the hypothesis with the pair $(y,x)$ gives $\pol_1(y\mid\state)>\pol_1(x\mid\state)$, which contradicts that $x$ precedes $y$ in $\ell_1$. Hence the metric does not decrease along the sequence of swaps, and its value at $\ell_1$ is at least its value at $\ell_2$. This proves our claim.
\end{proof}

We finally give the comparison between \unweighted and \reweighted contrastive losses in the softplus case. For hinge losses, we stated and proved in \cref{thm:comono} that whether $\val{\pirw}$ or $\val{\piunw}$ is higher depends on how aligned the rewards $\Q(\cdot\mid\state)$ are with the rollout policy $\base(\cdot\mid\state)$. We show here that an exactly similar statement holds for the softplus case as well.

\begin{theorem}[\unweighted versus \reweighted for softplus losses]\label{thm:sp-comono}
Consider a contextual bandit with a non-degenerate context $\state$, and fix $\temp>0$ and $\marg\ge0$.
\begin{enumerate}
\item Let $\lam>0$. If additionally $\base$ is \comon\ with $\Q$ at $\state$, then $\val{\pirw^{\temp}}(\state)\ge\val{\piunw^{\temp}}(\state)$; if $\base$ is \noncomon\ with $\Q$ at $\state$, then $\val{\pirw^{\temp}}(\state)\le\val{\piunw^{\temp}}(\state)$. In either case the inequality is strict as soon as some $\act,\act'\in\mathcal A_{+}$ have $\left(\Q(\state,\act)-\Q(\state,\act')\right)\left(\base(\act\mid\state)-\base(\act'\mid\state)\right)\ne0$.
\item Let $\lam=0$, and assume that $0<\Q(\state,\act)<1$ for every $\act$. Then both softplus state losses have a unique minimizer, $\piunw^{\temp}(\cdot\mid\state)$ and $\pirw^{\temp}(\cdot\mid\state)$, both with full support, and the conclusions of item~1 hold for them.
\end{enumerate}
\end{theorem}
\begin{proof}
Fix the context $\state$. Comonotonicity of $\base$ and $\Q$ at $\state$ says that
\begin{equation}\label{eq:sp-comono:align}
\left(\Q(\state,\act)-\Q(\state,\act')\right)\left(\base(\act\mid\state)-\base(\act'\mid\state)\right)\ \ge\ 0\quad\text{for all }\act,\act'\in\actset,
\end{equation}
and the \noncomon\ hypothesis reverses this inequality. We treat the comonotone case; reversing every sign below gives the other one.

We start with the $\lam=0$ case. Under its assumption, $\mathcal A_{+}=\actset$ and $\scond(\cdot\mid\state)$, $\fcond(\cdot\mid\state)$ are positive at every action. As shown in the proof of Theorem~\ref{thm:sp-base} for the \reweighted\ loss (the same argument applies to the \unweighted\ loss), a policy without full support has infinite loss, every policy with full support is $\pol^u_t$ for a score $u$ unique up to an additive constant, with $t=0$ for the \unweighted\ and $t=1$ for the \reweighted\ loss by \cref{eq:sp-support:param}, and at $\pol^u_t$ the softplus arguments are $\marg+u(\act')-u(\act)$ for both $t$, because $\ln\pol^u_0(\act'\mid\state)-\ln\pol^u_0(\act\mid\state)$ and $\ln\frac{\pol^u_1(\act'\mid\state)}{\base(\act'\mid\state)}-\ln\frac{\pol^u_1(\act\mid\state)}{\base(\act\mid\state)}$ both equal $u(\act')-u(\act)$, the normalizer of \cref{eq:sp-support:param} cancelling; so both losses equal $C_\temp(u)$, and the same proof shows that $C_\temp$ has a unique minimizer $u$ over the zero-sum scores. 

Hence each loss has a unique minimizer, with full support, namely, by \cref{eq:sp-support:param} with $t=0$ and $t=1$, $\piunw^{\temp}(\act\mid\state)=e^{u(\act)}/\sum_{\act'}e^{u(\act')}$ and $\pirw^{\temp}(\act\mid\state)=\base(\act\mid\state)e^{u(\act)}/\sum_{\act'}\base(\act'\mid\state)e^{u(\act')}$; that is,
\[
\pirw^{\temp}(\act\mid\state)=\frac{\piunw^{\temp}(\act\mid\state)\,\base(\act\mid\state)}{\sum_{\act'}\piunw^{\temp}(\act'\mid\state)\,\base(\act'\mid\state)},
\]
so that
\[
\val{\pirw^{\temp}}(\state)-\val{\piunw^{\temp}}(\state)=\frac{\Cov_{\act\sim\piunw^{\temp}(\cdot\mid\state)}\left(\base(\act\mid\state),\Q(\state,\act)\right)}{\sum_{\act'}\piunw^{\temp}(\act'\mid\state)\,\base(\act'\mid\state)},
\]
the numerator being
\[
\sum_\act\piunw^{\temp}(\act\mid\state)\base(\act\mid\state)\Q(\state,\act)-\sum_{\act'}\piunw^{\temp}(\act'\mid\state)\base(\act'\mid\state)\sum_\act\piunw^{\temp}(\act\mid\state)\Q(\state,\act).
\]

By Lemma~\ref{lem:cov}, the covariance equals 
$$\tfrac12\sum_{\act,\act'}\piunw^{\temp}(\act\mid\state)\piunw^{\temp}(\act'\mid\state)\left(\base(\act\mid\state)-\base(\act'\mid\state)\right)\left(\Q(\state,\act)-\Q(\state,\act')\right),$$ 
whose terms are nonnegative by \cref{eq:sp-comono:align}, and positive for the pairs in the strictness condition, as $\piunw^{\temp}(\cdot\mid\state)$ has full support. This proves the $\lam=0$ case.

Now, we deal with the $\lam > 0$ case. The proof follows that of Theorem~\ref{thm:comono}: the two losses are the endpoints $t=0$ and $t=1$ of the family $F_{t,\temp}$ of \cref{eq:sp-support:F}, and we follow the value of its minimizer as $t$ moves from $0$ to $1$. The softplus is smooth, so there is no smoothing to do and no limit to take at the end.

We first define the ``path'' connecting $\pirw^\temp$ and $\piunw^\temp$. By Lemma~\ref{lem:sp-support}, $\piunw^{\temp}(\cdot\mid\state)=\pol_{0,\temp}$ and $\pirw^{\temp}(\cdot\mid\state)=\pol_{1,\temp}$ on $\mathcal A_{+}$, and both vanish outside $\mathcal A_{+}$; here, as in the proof of that lemma, $\pol_{t,\temp}=\pol^{u_{t,\temp}}_t$ and $u_{t,\temp}$ is the unique minimizer of $F_{t,\temp}$ over the zero-sum scores on $\mathcal A_{+}$.Since $\Q(\state,\cdot)$ vanishes off $\mathcal A_{+}$, $\val{\pol}(\state)=\sum_{\act\in\mathcal A_{+}}\pol(\act\mid\state)\Q(\state,\act)$ for both policies, so we work on $\mathcal A_{+}$ from now on. All vectors, matrices and sums are indexed by $\mathcal A_{+}$, and if $\mathcal A_{+}$ is a singleton both policies are the point mass on it and there is nothing to prove. In the proof of Lemma~\ref{lem:sp-support}, the existence and uniqueness of the minimizer of $F_{t,\temp}$ over the zero-sum scores and the stationarity condition \cref{eq:sp-support:foc} were derived for $t\in\{0,1\}$, but the derivation used $t$ only through the parametrization \cref{eq:sp-support:param}, which stays valid for every real $t$, with $\pol^u_t$ eing policy with full support on $\mathcal A_{+}$ and $u=\ln\pol^u_t-t\ln\base(\cdot\mid\state)+\mathrm{const}$. The only change is that the lower bound $t\ln\base(\act\mid\state)+\bar u$ on the logarithm in \cref{eq:sp-support:F}, used there to show that $F_{t,\temp}$ tends to $+\infty$ with $\max_\act|u(\act)|$, becomes $-|t|\max_\act|\ln\base(\act\mid\state)|+\bar u$. So for every real $t$, $F_{t,\temp}$ has a unique minimizer $u_{t,\temp}$ over the zero-sum scores, $\pol_{t,\temp}\coloneqq\pol^{u_{t,\temp}}_t$ has full support on $\mathcal A_{+}$, and \cref{eq:sp-support:foc} holds. We write $\val{\pol_{t,\temp}}(\state)=\sum_\act\pol_{t,\temp}(\act\mid\state)\Q(\state,\act)$ for $t$ in an open interval containing $[0,1]$.

We now prove differentiability of the path. Write the stationarity condition \cref{eq:sp-support:foc} as the equation $\Phi(u,t)=0$ between zero-sum vectors, where $\Phi(u,t)$ is the gradient \cref{eq:sp-support:grad} of $F_{t,\temp}$ at $u$, a smooth function of the zero-sum score $u$ and of $t$. Its derivative in $u$ is $K+\lam\left(\operatorname{diag}(\pol^u_t(\cdot\mid\state))-\pol^u_t(\cdot\mid\state)\,\pol^u_t(\cdot\mid\state)^{\top}\right)$, where $K$ is the matrix of second derivatives of $C_\temp$ at $u$. Each term of \cref{eq:sp-support:F} depends on $u$ only through $u(\act')-u(\act)$, so differentiating \cref{eq:sp-support:grad} once more and grouping the two orientations of each pair, $K$ is the graph Laplacian
\begin{equation}\label{eq:sp-comono:k}
\begin{split}
(Kv)(\act)&=\sum_{\act'\ne\act}k_{\act\act'}\left(v(\act)-v(\act')\right),\\
k_{\act\act'}&=\frac{1}{\temp}\left[\scond(\act\mid\state)\fcond(\act'\mid\state)\,\sigma'\left(\tfrac{\marg+u(\act')-u(\act)}{\temp}\right)+\scond(\act'\mid\state)\fcond(\act\mid\state)\,\sigma'\left(\tfrac{\marg+u(\act)-u(\act')}{\temp}\right)\right]\\
&\ge\ 0,
\end{split}
\end{equation}
with $\sigma'=\sigma(1-\sigma)>0$. So $K$ is symmetric, $K\mathbf 1=0$, and $v^\top Kv=\sum_{\{\act,\act'\}}k_{\act\act'}\left(v(\act)-v(\act')\right)^2\ge0$. The second summand of the derivative is the covariance matrix of $\pol^u_t(\cdot\mid\state)$, whose quadratic form at $v$ is the variance of $v$ under $\pol^u_t(\cdot\mid\state)$, positive for every nonconstant $v$, in particular for every nonzero zero-sum $v$. The derivative of $\Phi$ in $u$ is therefore invertible on the zero-sum vectors. We now use the implicit function theorem on $\Phi$: if $\Phi$ is continuously differentiable near a point $(u_0,t_0)$ with $\Phi(u_0,t_0)=0$ and the derivative of $\Phi$ in $u$ is invertible there, then on a neighbourhood of $t_0$ the solutions of $\Phi(u,t)=0$ near $u_0$ form the graph of a continuously differentiable function of $t$. Since $u_{t,\temp}$ is the unique zero-sum solution of $\Phi(u,t)=0$, by strict convexity, this function is $t\mapsto u_{t,\temp}$; hence $t\mapsto u_{t,\temp}$ and $t\mapsto\pol_{t,\temp}$ are continuously differentiable. Write $\dot u$ and $\dot\pol$ for their derivatives at a fixed $t$, and abbreviate $\pol_{t,\temp}(\cdot\mid\state)$ to $\pol_{t,\temp}$ and $u_{t,\temp}$ to $u$ in the rest of the proof.

Since $\pol_{t,\temp}$ is the softmax of $u+t\ln\base(\cdot\mid\state)$ by \cref{eq:sp-support:param}, the chain rule gives
\[
\dot\pol=\left(\operatorname{diag}(\pol_{t,\temp})-\pol_{t,\temp}\pol_{t,\temp}^\top\right)\left(\dot u+\ln\base(\cdot\mid\state)\right),
\]
that is,
\[
\dot\pol=\operatorname{diag}(\pol_{t,\temp})\,y,\quad y\coloneqq\dot u+\ln\base(\cdot\mid\state)-\left(\sum_\act\pol_{t,\temp}(\act)\left(\dot u(\act)+\ln\base(\act\mid\state)\right)\right)\mathbf 1 ,
\]
where the vector $y$ satisfies $\sum_\act\pol_{t,\temp}(\act)\,y(\act)=0$. Differentiating $\Phi(u_{t,\temp},t)=0$ in $t$ gives $K\dot u+\lam\dot\pol=0$, and $K\dot u=K\left(y-\ln\base(\cdot\mid\state)\right)$ because $K\mathbf 1=0$. Thus, we have:
\[
M y=K\ln\base(\cdot\mid\state),\quad M\coloneqq K+\lam\operatorname{diag}\left(\pol_{t,\temp}\right).
\]

$M$ is symmetric positive definite, as $v^\top Mv=v^\top Kv+\lam\sum_\act\pol_{t,\temp}(\act)\,v(\act)^2>0$ for $v\ne0$; let $G\coloneqq M^{-1}$. Then $y=GK\ln\base(\cdot\mid\state)$ and we have:
\begin{align*}
&\dot\pol=\operatorname{diag}(\pol_{t,\temp})\,G\,\left(M-\lam\operatorname{diag}(\pol_{t,\temp})\right)\ln\base(\cdot\mid\state)=\Omega\,\ln\base(\cdot\mid\state),\\
&\Omega\coloneqq\operatorname{diag}(\pol_{t,\temp})-\lam\operatorname{diag}(\pol_{t,\temp})\,G\operatorname{diag}(\pol_{t,\temp}).
\end{align*}
Now, we note three properties of $\Omega$. (1) It is symmetric. (2) Its rows sum to zero: $M\mathbf 1=\lam\,\pol_{t,\temp}$, since $K\mathbf 1=0$. Hence $G\,\pol_{t,\temp}=\mathbf 1/\lam$ and $\Omega\mathbf 1=\pol_{t,\temp}-\operatorname{diag}(\pol_{t,\temp})\mathbf 1=0$. (3) And its off-diagonal entries $\Omega_{\act\act'}=-\lam\,\pol_{t,\temp}(\act)\,\pol_{t,\temp}(\act')\,G_{\act\act'}$ are nonpositive, because $G$ is entrywise nonnegative. If $x$ is a column of $G$, then $Mx$ is a unit vector, and if $x$ had a negative minimum at a coordinate $\act$, then $(Mx)(\act)=\lam\pol_{t,\temp}(\act)\,x(\act)+\sum_{\act'\ne\act}k_{\act\act'}\left(x(\act)-x(\act')\right)<0$, which is impossible. 

For a symmetric matrix with zero row sums, we note the following identity for all vectors $f,g$: $f^\top\Omega g=-\sum_{\{\act,\act'\}}\Omega_{\act\act'}\left(f(\act)-f(\act')\right)\left(g(\act)-g(\act')\right)$. Thus, we have:
\begin{equation}\label{eq:sp-comono:deriv}
\begin{split}
\frac{d}{dt}\val{\pol_{t,\temp}}(\state)&=\sum_\act\Q(\state,\act)\,\dot\pol(\act)=\Q(\state,\cdot)^\top\,\Omega\,\ln\base(\cdot\mid\state)\\
&=\sum_{\{\act,\act'\}}\omega_{\act\act'}(t)\,\left(\Q(\state,\act)-\Q(\state,\act')\right)\left(\ln\base(\act\mid\state)-\ln\base(\act'\mid\state)\right),
\end{split}
\end{equation}
The sum runs over unordered pairs of distinct actions of $\mathcal A_{+}$, with $\omega_{\act\act'}(t)=\lam\,\pol_{t,\temp}(\act)\,\pol_{t,\temp}(\act')\,G_{\act\act'}\ge0$. Since $\ln$ is increasing, \cref{eq:sp-comono:align} makes every term of \cref{eq:sp-comono:deriv} nonnegative, and integrating over $t\in[0,1]$ gives
\[
\val{\pirw^{\temp}}(\state)-\val{\piunw^{\temp}}(\state)=\int_0^1\frac{d}{dt}\val{\pol_{t,\temp}}(\state)\,dt\ \ge\ 0 .
\]

Finally, we prove the strictness of the inequality under the stated condition. Suppose that $\act_1,\act_2\in\mathcal A_{+}$ satisfy $\left(\Q(\state,\act_1)-\Q(\state,\act_2)\right)\left(\base(\act_1\mid\state)-\base(\act_2\mid\state)\right)>0$. Then $\Q(\state,\cdot)$ is not constant on $\mathcal A_{+}$, so some $\act''\in\mathcal A_{+}$ has $\Q(\state,\act'')<1$, that is, $\fcond(\act''\mid\state)>0$, and \cref{eq:sp-comono:k} gives $k_{\act\act''}\ge\frac1\temp\scond(\act\mid\state)\fcond(\act''\mid\state)\,\sigma'\left(\tfrac{\marg+u(\act'')-u(\act)}{\temp}\right)>0$ for every $\act\ne\act''$, because $\scond>0$ on $\mathcal A_{+}$. Then $G$ is entry-wise positive. Indeed, let $x$ be a column of $G$, so that $x\ge0$ and $Mx$ is a unit vector. If $x(\act)=0$ for some $\act$, then we have:
$$(Mx)(\act)=\lam\pol_{t,\temp}(\act)\,x(\act)+\sum_{\act'\ne\act}k_{\act\act'}\left(x(\act)-x(\act')\right)=-\sum_{\act'\ne\act}k_{\act\act'}\,x(\act')\le0,$$ 
so $(Mx)(\act)=0$ and $x(\act')=0$ for every $\act'$ with $k_{\act\act'}>0$. Since $k_{\act\act''}>0$ for every $\act\ne\act''$, this gives $x(\act'')=0$ regardless of $\act$, and then, applied at $\act''$, $x(\act')=0$ for every $\act'$, thus $x=0$, contradicting $Mx\ne0$. Hence $G_{\act\act'}>0$ and $\omega_{\act\act'}(t)>0$ for every pair and every $t\in[0,1]$. In \cref{eq:sp-comono:deriv} the term of the pair $\{\act_1,\act_2\}$ is then strictly positive at every $t$, both of its factors being nonzero of the same sign, the remaining terms are nonnegative, and the derivative is continuous in $t$; so the integral above is strictly positive. In the \noncomon\ case every sign reverses.
\end{proof}

%% file: sections/experimental_details.tex
\section{Additional Experiments and Details}

\label{app:expt}

Here we first validate our \cref{thm:rw-base,thm:beat-sft,thm:comono} via the following synthetic setup.

 \subsection{Empirical analysis in semi-synthetic bandits}
 \label{subsec:expt:semisynthetic}
We use a semi-synthetic contextual bandit to separate the effects of
reference--reward alignment, finite feedback, and SFT regularization.
The contexts are representations of MS MARCO queries; the actions,
reward mechanism, and reference policies are constructed.

\paragraph{Experimental setup}
We want to design a contextual bandit setup where each context $\ctx$ is given by text queries and each context has a space of $|\actset|=32$ actions. 
We obtain query texts from the MS MARCO dataset~\cite{msmarco} where each query is represented using its feature extracted by a frozen T5-base encoder~\cite{raffel2023exploringlimitstransferlearning}. In this synthetic setup, the policy selects an action rather than generating unrestricted text. Both the base $\base$ and the learner $\pi_\theta$ are softmax policies of the form $\pi_\theta(\cdot\mid q)=\operatorname{softmax}(f_\theta(x_q))$. \add{We consider two classes of policies under shared parametrization: (1) where $f_\theta$ is an MLP ($\base$ is constructed at frozen weights of $f_\theta$) and (2) where $f_\theta$ is linear (we call the resulting class log-linear).}

We specifically control the \emph{alignment} between $\base$ and $\Q$, by assigning rewards to its action ranks using permutations
with prescribed inversion counts. This gives exact per-query Kendall
correlations $\tau\in\{-1,-\tfrac12,0,\tfrac12,1\}$ between the base and the reward. Positive and negative actions are sampled independently from $\base$ and we train $f_\theta$ using the contrastive losses $\lunw,\,\lrw$ on its induced policy $\pi_\theta$.

\paragraph{Convex combination loss:}Let $\lctr, \lsft$ be the contrastive (\unweighted or \reweighted) and SFT terms respectively. We optimize the convex combination $(1 - \omega)\lctr + \omega\lsft$. It should be noted that the optimizer $\pol_{\omega}$ of this loss is equivalent to the optimizers of $\lrw, \lunw$ at $\lambda=\omega/(1-\omega)$. Taking the convex combination was an experimental choice to avoid numerical instabilities. Since $\lam\to\infty$ as $\omega\to 1$, it implies that the pure SFT trained $\pi_{\mathrm{SFT}}$ corresponds to the optimizer for the above loss when $\omega = 1$.

\begin{figure}[H]
    \centering
    \includegraphics[width=\linewidth]{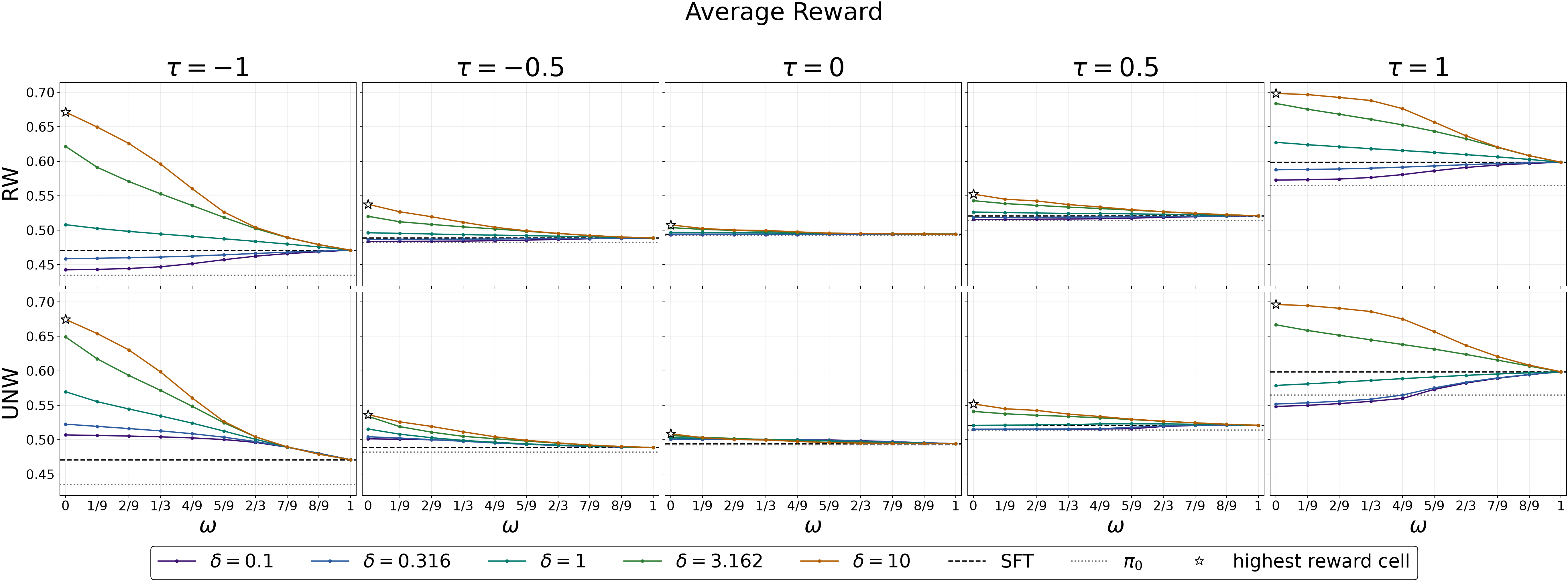}
    \caption{MLP-parametrized contrastive policy optimization on a semi-synthetic bandit: Expected reward of the trained policy as a function of $\omega$ for the \reweighted (top row) and \unweighted (bottom row) losses, one column per Kendall correlation $\tau$ between $\base$ and $\Q$; each curve is one margin $\delta$, the dashed line is the trained SFT endpoint, the dotted line corresponds to $\pi_0$, and the star marks the highest-reward cell.}
    \label{fig:mlp-plot-synthetic}
\end{figure}

\paragraph{Reward-alignment study}
We evaluate ten values of $\omega\in\{0,1/9,\ldots,1\}$ and all five margins on a log scale
$\delta\in[0.1,10]$ for both $\pirw, \piunw$ sharing one trained SFT endpoint at $\omega=1$, across all 5 reward functions with the correlations $\{-1,-\tfrac12,0,\tfrac12,1\}$. For the class of MLP-parametrized policies, we show $\val{\pirw}, \val{\piunw}$ and $\val{\pirw} - \val{\piunw}$ across the grid in \cref{fig:mlp-plot-synthetic}.

We make the following observations. \textbf{(1)} In every column the highest-reward cell is the pure-hinge endpoint $\omega=0$ at the largest margin $\delta=10$, and for $\delta\ge1$ the value decreases monotonically in $\omega$ toward the SFT endpoint; for $\delta\le0.316$ the order is reversed and the value rises toward SFT. This is the population behaviour of Theorem~\ref{thm:beat-sft} and Theorem~\ref{thm:trpo} (the gain is $\Gamma/\lam$ and the trust region shrinks with $\lam$) on one side, and of Lemma~\ref{lem:rw-sft} on the other; there is no interior optimum here. \textbf{(2)} The SFT endpoint is above the base in every column.

\begin{figure}[H]
    \centering
    \includegraphics[width=\linewidth]{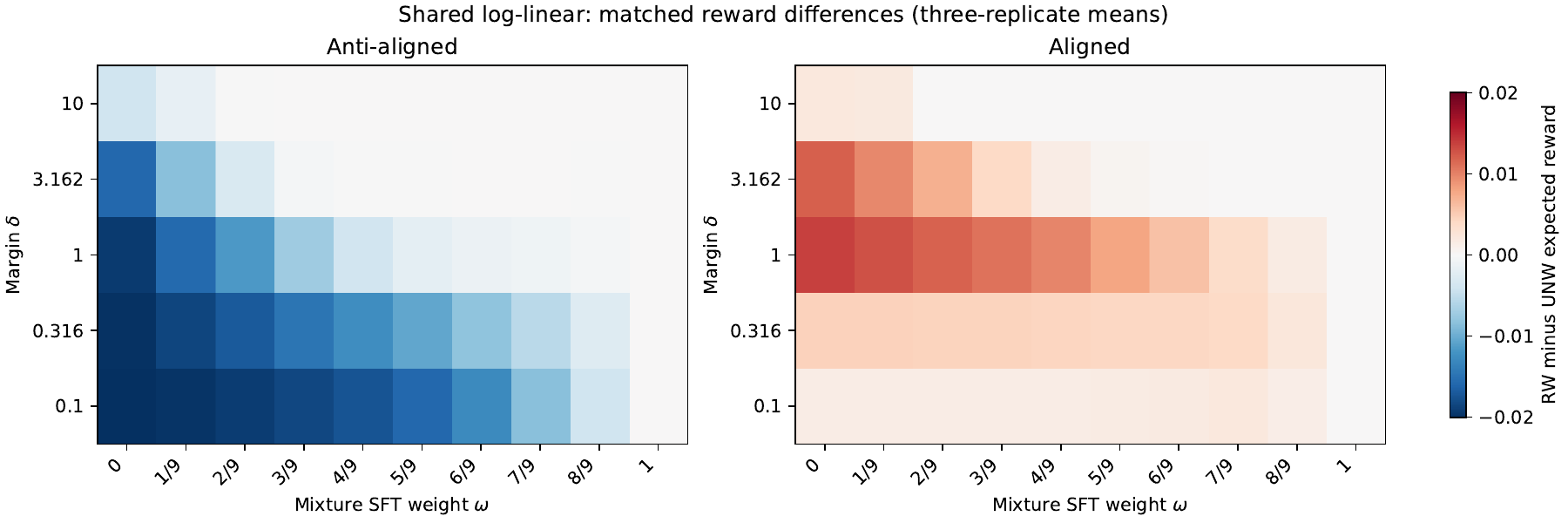}
    \caption{\add{$\val{\pirw}-\val{\piunw}$  for a shared log-linear policy across the grid. The left
    panel is anti-aligned and the right panel aligned. Cells average
    three replicates. Positive values favor \reweighted{} and negative
    values favor \unweighted{}. The horizontal axis is the convex-mixture
    weight $\omega$, with additive coefficient $\lam=\omega/(1-\omega)$;
    the shared SFT endpoint at $\omega=1$ has zero difference.}}
    \label{fig:loglinear-plot-synthetic}
\end{figure}

\add{For the log-linear policy parametrization, we report the difference between $V(\pirw)$ and $V(\piunw)$ across two levels of reward alignment $\tau\in\{+1, -1\}$ in the following \cref{fig:loglinear-plot-synthetic} for the full grid across different SFT anchor weights and margins across three independent runs.}

\add{We make the following observations: \textbf{(1)} $\pirw$ dominates $\piunw$ for aligned policies and it flips for anti-aligned ones, validating the prediction of Theorem~\ref{thm:comono}. \textbf{(2)} The dominance is more profound for smaller $\lambda$ since for larger $\lambda$, both converge to $\pi_{\mathrm{SFT}}$. The above results can be summarized in the following table:}

\begin{table}[H]
    \centering
    \small
    \setlength{\tabcolsep}{5pt}
    \add{\begin{tabular}{lllrrr}
        \toprule
        Policy class & Alignment & Expected order & Strict & Ties & Reversals \\
        \midrule
        Log-linear & Aligned & RW above UNW & 34/40 & 6/40 & 0/40 \\
        Log-linear & Anti-aligned & UNW above RW & 32/40 & 8/40 & 0/40 \\
        MLP & Aligned & RW above UNW & 37/40 & 3/40 & 0/40 \\
        MLP & Anti-aligned & UNW above RW & 37/40 & 3/40 & 0/40 \\
        \bottomrule
    \end{tabular}}
    \caption{\add{Alignment study: counts of strict differences,
    exact ties, and reversals in reward ordering.}}
    \label{tab:synthetic:alignment}
\end{table}

\paragraph{\add{Controlling and measuring reward variance}}
\add{For a separate experiment to validate \cref{thm:beat-sft} we measure reward variance under the corresponding $\scond$. For each context, we draw
$\act_1,\ldots,\act_{128}$ independently from $\scond(\cdot\mid\ctx)$ and
compute the empirical reward variance. We evaluate
$\lam\in\{8,16,32,64\}$, holding $\delta=10$ for log-linear and $\delta=2.56$ for MLP fixed throughout the variance studies.
This setup is to validate \cref{thm:rw-base,thm:beat-sft} where we have theoretically proved that $\pirw$ always improves over $\base$ and that both $\pirw,\piunw$ improve over $\scond$ for large $\lambda$ and margin with the expected reward gain lower bounded by $\Var_{a\sim\scond}Q(s,a)$.}

\paragraph{\add{Improvement of $\pirw$ over $\base$}}
\add{We first validate \cref{thm:rw-base} via the following tables.}
\begin{table}[H]
    \centering
    \small
    \add{\begin{tabular}{lrr}
        \toprule
        Alignment & RW fits above base & Mean reward improvement \\
        \midrule
        Fully anti-aligned, $-1$ & 135/135 & $+0.080247$ \\
        Partially anti-aligned, $-1/2$ & 135/135 & $+0.082075$ \\
        Zero rank correlation, $0$ & 135/135 & $+0.072485$ \\
        Partially aligned, $+1/2$ & 135/135 & $+0.060472$ \\
        Fully aligned, $+1$ & 135/135 & $+0.065925$ \\
        \midrule
        \textbf{Total} & \textbf{675/675} & $\mathbf{+0.072241}$ \\
        \bottomrule
    \end{tabular}}
    \caption{\add{Shared MLP: improvement of $\pirw$ over its matched $\base$.}}
    \label{tab:synthetic:base-mlp}
\end{table}

\begin{table}[H]
    \centering
    \small
    \add{\begin{tabular}{lrr}
        \toprule
        Alignment & RW fits above base & Mean reward improvement \\
        \midrule
        Fully anti-aligned, $-1$ & 135/135 & $+0.011269$ \\
        Fully aligned, $+1$ & 135/135 & $+0.010545$ \\
        \midrule
        \textbf{Total} & \textbf{270/270} & $\mathbf{+0.010907}$ \\
        \bottomrule
    \end{tabular}}
    \caption{\add{Shared log-linear: improvement of $\pirw$ over its matched
    $\base$.}}
    \label{tab:synthetic:base-loglinear}
\end{table}

\add{Thus, for both policy classes $\pirw$ improves over $\base$, validating \cref{thm:rw-base}.}

\noindent\add{\textbf{Variance-dependent improvement over SFT.}
We compare $\val{\pirw}-\val{\widehat{\scond}}$ and
$\val{\piunw}-\val{\widehat{\scond}}$ at each SFT weight, targeting
the variance-dependent prediction of Theorem~\ref{thm:beat-sft}.
For compactness, \cref{tab:synthetic:variance} averages aligned and
anti-aligned results equally at each variance level. Each difference is
computed against its own learned SFT baseline before averaging.
The measured variance is also averaged across alignments; it is not
the variance of a merged action sample.}

\begin{table}[H]
    \centering
    \small
    \setlength{\tabcolsep}{4pt}
    \resizebox{\linewidth}{!}{%
    \add{\begin{tabular}{llrrrrr}
        \toprule
        Policy class & Level & Mean $\widehat\sigma_+^2$ & $\lam=8$ & $\lam=16$ & $\lam=32$ & $\lam=64$ \\
        \midrule
        Log-linear & Low & 0.0000718 & 0.0427 / 0.0427 & 0.0213 / 0.0213 & 0.0107 / 0.0107 & 0.0053 / 0.0053 \\
        Log-linear & Medium & 0.0006434 & 0.3350 / 0.3350 & 0.1676 / 0.1676 & 0.0838 / 0.0838 & 0.0419 / 0.0419 \\
        Log-linear & High & 0.0017764 & 0.9100 / 0.9100 & 0.4538 / 0.4538 & 0.2270 / 0.2270 & 0.1135 / 0.1135 \\
        \midrule
        MLP & Low & 0.0006086 & 0.3871 / 0.2889 & 0.2071 / 0.1633 & 0.1838 / 0.1611 & 0.1578 / 0.1463 \\
        MLP & Medium & 0.0053766 & 3.4111 / 2.8471 & 1.8844 / 1.6767 & 0.6802 / 0.5945 & 0.5299 / 0.4796 \\
        MLP & High & 0.0143392 & 8.6731 / 7.7626 & 4.5929 / 4.1606 & 1.9270 / 1.7225 & 1.0185 / 0.8736 \\
        \bottomrule
    \end{tabular}}
    }
    \caption{\add{Variance study: each entry reports \reweighted{} / \unweighted{}
    improvement over the matched learned SFT policy, multiplied by
    $10^3$. Measured variance is unscaled. Rows average the two alignments
    equally, with three replicates per alignment. Low, medium, and high
    denote increasing reward spreads within each class, not identical
    variances across classes.}}
    \label{tab:synthetic:variance}
\end{table}

\add{Thus, we empirically validate that the reward gain over SFT tracks with $\Var_{\scond}(Q)/\lambda$, as predicted by the proof of \cref{thm:beat-sft}.}

\subsection{Generative Retrieval Experimental Details}
\label{app:expt:GR}
\paragraph{Data and reference policies}

We use the MS MARCO passage collection~\citep{msmarco} of 8,841,823 passages. The training
source contains 808,731 queries and 532,751 qrel rows. Feedback construction
uses a fixed subset of 10,000 training queries with at least one positive
qrel. A fixed 500-query selection holdout is chosen from this subset, which we used for validation.
Final reporting uses all 6,980 dev-small queries and their 7,437 qrel rows.

The generator and tokenizer use Facebook BART-large~\citep{bart2020}. A fixed compact
trie and terminal-state mapping cover the complete passage collection. Each
terminal pseudo-query identifier maps to one or more passage IDs and contains
at most 31 content tokens followed by exactly one EOS token. All references
come from one vanilla MINDER~\citep{minder2023} training trajectory. Its upstream supervised
query-to-pseudo-query recipe uses batch size 128, learning rate
$5\times10^{-5}$, 1,000 warmup updates, and checkpoints every 1,000 updates.

We evaluate three reference blocks: \emph{Early} at update 1,000,
\emph{Midpoint} at update 3,000, and \emph{Final} at the final vanilla
checkpoint. Midpoint is the saved checkpoint whose native beam-slot reward
is closest to the arithmetic mean of Early and Final, among checkpoints
strictly between these anchors; ties favor the earlier update. Each block
uses its own frozen reference, candidate support, feedback pairs, and
independent model selection.

\paragraph{Candidate support and reward construction}

For each reference and feedback query, we 
run trie-constrained beam search with width 100 and maximum identifier
length 32, including EOS. The output contains exactly 100 terminal slots.
The constrained beam score sums token log-probabilities, including EOS.

For every retained identifier, we store its full sequence log-probability
\begin{equation}
\ell_0(q,z)=\sum_{t=1}^{T(z)}\log\pi_0(z_t\mid q,z_{<t}).
\label{eq:minder-reference-score}
\end{equation}
EOS contributes to this sum; padding does not. These scores are neither
length-normalized nor interchangeable with the constrained beam scores.

For training support, duplicate terminal states are collapsed, retaining the
occurrence with the highest constrained score. For reward feedback, we have used Qwen3-Reranker-0.6B~\citep{qwen3embedding2025}.
For passage $d$,
the soft reward is
\begin{equation}
\label{eq:minder-passage-reward}
\widehat Q(q,d)=\sigma\!\left(a_{\mathrm{yes}}(q,d)-a_{\mathrm{no}}(q,d)\right),
\end{equation}
where $a_{\mathrm{yes}}$ and $a_{\mathrm{no}}$ are the corresponding output
logits of the Qwen reranker. If $M(z)$ is the set of unique passages associated with identifier
$z$, its reward is the average:
\begin{equation}
\bar Q(q,z)=\frac{1}{|M(z)|}\sum_{d\in M(z)}\widehat Q(q,d).
\label{eq:minder-identifier-reward}
\end{equation}

\paragraph{Relevance label scoring}
Let $\mathcal C_q$ denote the finite set of unique
candidate identifiers for query $q$. Define
\begin{align}
b_{\mathcal C}(z\mid q)&=\frac{\exp\ell_0(q,z)}{\sum_{z'\in\mathcal C_q}\exp\ell_0(q,z')},\quad
V_{\mathcal C}(q)=\sum_{z\in\mathcal C_q}b_{\mathcal C}(z\mid q)\bar Q(q,z),
\\
p_+(z\mid q)&=\frac{b_{\mathcal C}(z\mid q)\bar Q(q,z)}{V_{\mathcal C}(q)},\quad p_-(z\mid q)=\frac{b_{\mathcal C}(z\mid q)(1-\bar Q(q,z))}{1-V_{\mathcal C}(q)}.
\label{eq:minder-routing}
\end{align}
We draw eight positives and eight negatives independently, with replacement,
and pair them by draw index. Every positive draw is
also an SFT target.

The trainable score $\ell_\theta(q,z)$ is the full-vocabulary sequence sum
defined as in~\eqref{eq:minder-reference-score}. The readouts are
\begin{equation}
s_{\mathrm{UNW}}(q,z)=\ell_\theta(q,z),\qquad
s_{\mathrm{RW}}(q,z)=\ell_\theta(q,z)-\ell_0(q,z).
\end{equation}
For a batch of $B$ routed pairs and margin $\delta$, we minimize
\begin{align}
\mathcal L_{\mathrm{ctr}}
&=\frac1B\sum_{i=1}^B
  [\delta-s(q_i,z_i^+)+s(q_i,z_i^-)]_+,\\
\mathcal L_{\mathrm{SFT}}
&=-\frac1B\sum_{i=1}^B\ell_\theta(q_i,z_i^+),\\
\mathcal L_\omega
&=(1-\omega)\mathcal L_{\mathrm{ctr}}+\omega\mathcal L_{\mathrm{SFT}}.
\label{eq:minder-mixture}
\end{align}
Neither term is normalized by identifier length. We use $\omega$ for the
experimental mixture weight to distinguish it from the additive SFT
coefficient in the theory. For $\omega<1$, the objectives have the same
minimizers under $\lambda=\omega/(1-\omega)$. This rescaling does not imply
identical finite-training trajectories with a fixed optimizer configuration.

For each readout and reference, the grid is
\begin{equation}
\omega\in\{0,1/9,\ldots,8/9\},\quad
\delta\in\{0.1,2.575,5.05,7.525,10\}.
\end{equation}
Thus each block has 45 coordinates per readout, including the pure hinge
endpoint at $\omega=0$. One shared SFT model represents $\omega=1$.

\paragraph{Comparison methods}
We include the frozen reference, Reward-SFT, DPO~\citep{dpo2023},
RankNet~\citep{ranknet2005}, SimPO~\citep{simpo}, and DPO+SFT
as baselines.
Let $\ell_\theta^\pm=\log\pi_\theta(z^\pm\mid q)$ and
$\ell_0^\pm=\log\pi_0(z^\pm\mid q)$ denote the full-vocabulary
sequence log-probabilities under the trained and frozen
reference policies, respectively.
Let $T^\pm$ denote the numbers of scored, non-padding target tokens,
and define $\operatorname{sp}(x)=\log(1+e^x)$.
With $\operatorname{mean}$ denoting the empirical batch average, the respective losses are

\begin{align}
\label{eq:dpo_losses}
\mathcal L_{\mathrm{SFT}}
&=-\operatorname{mean}\ell_\theta^+,
\\
\mathcal L_{\mathrm{DPO}}(\beta)
&=\operatorname{mean}\operatorname{sp}
 \left(-\beta
 \left[(\ell_\theta^+-\ell_\theta^-)
       -(\ell_0^+-\ell_0^-)\right]\right),
\\
\mathcal L_{\mathrm{RankNet}}
&=\operatorname{mean}\operatorname{sp}
 \left[-\beta_{\mathrm{RankNet}}
       (\ell_\theta^+-\ell_\theta^-)\right],
\\
\mathcal L_{\mathrm{SimPO}}
&=\operatorname{mean}\operatorname{sp}
 \left(-\left[
 \beta_{\mathrm{SimPO}}
 \left(\frac{\ell_\theta^+}{T^+}
       -\frac{\ell_\theta^-}{T^-}\right)
 -\delta_{\mathrm{SimPO}}\right]\right),
\\
\mathcal L_{\mathrm{DPO+SFT}}
&=(1-\eta)\mathcal L_{\mathrm{DPO}}
       (\beta_{\mathrm{DPO+SFT}})
  +\eta\mathcal L_{\mathrm{SFT}}.
\end{align}

\add{
For each reference checkpoint, the pairwise baselines share the same
frozen reward-routed positive--negative pair dataset.
On the fixed candidate support, positive and negative identifiers are
sampled independently with replacement, with weights proportional to
$\exp(\ell_0(q,z))\bar Q(q,z)$ and
$\exp(\ell_0(q,z))(1-\bar Q(q,z))$, respectively.
These are the implemented finite-candidate-support counterparts of
the success- and failure-conditioned reference policies
$\scond$ and $\fcond$.
Reward-SFT uses only the stored positive occurrences, retaining their
multiplicity.
Thus, the feedback source is shared, but per-epoch data exposure is
not identical across all baselines.
}

\add{All the pairwise baselines use exactly the same positive and negative actions sampled from the success- and failure-conditioned base policies $\scond$ and $\fcond$ respectively. Across individual tasks, we ran a hyperparameter sweep over $\beta_{DPO}, \beta_{SimPO}, \delta_{SimPO}, \beta_{RankNet}$ over a fixed grid to select the best hyperparameters on a held-out validation set.}

\paragraph{\add{Tuning Budget and Training details:}}
\add{
We tune each trainable baseline separately for the Early, Midpoint,
and Final reference checkpoints.
For DPO, we search
$\beta_{\mathrm{DPO}}\in\{0.01,0.1,0.3,1\}$;
for RankNet,
$\beta_{\mathrm{RankNet}}\in\{0.1,0.3,1,3\}$;
and for SimPO,
$\beta_{\mathrm{SimPO}}\in\{2,5,7\}$ and
$r=\delta_{\mathrm{SimPO}}/\beta_{\mathrm{SimPO}}
\in\{0,0.5\}$, setting
$\delta_{\mathrm{SimPO}}=\beta_{\mathrm{SimPO}}r$.
For DPO+SFT, we search
$\beta_{\mathrm{DPO+SFT}}\in\{0.01,0.1,0.3,1\}$ and
$\lambda\in\{0.01,0.1, 1, 2\}$, where $\lambda = \eta/(1-\eta)$.
Each of these objective configurations is tuned over learning rates
$\{5\times10^{-7},2\times10^{-6}, 2\times10^{-5}\}$.
For Reward-SFT, we searched over additional learning rates
$\{10^{-6},3\times10^{-6}, 5\times10^{-6}, 7\times10^{-6}\}$.
} Each trainable model starts from its block's reference. Optimization uses
AdamW, batch size 16, one accumulation step, gradient
clipping at 1.0, and a linear schedule with 250 warmup updates.

\paragraph{Native decoding and passage ranking}

Evaluation uses the same query prefix and constrained beam width 100.
Generation reward retains all 100 raw terminal slots. Passage ranking first
deduplicates decoded identifier text, retaining the occurrence with
the largest constrained sequence score. For retained identifier
$z$, let $c(z,d)$ be its mapping multiplicity for passage $d$. Native passage
mass is
\begin{equation}
A(q,d)=\sum_z c(z,d)\exp\ell_{\mathrm{beam}}(q,z).
\end{equation}
Passages are ordered by decreasing mass; ties follow first appearance in
the resolved identifier stream. The first 100 unique passage IDs form the
native output. Beam search approximates sequence search, and the final
passage ranking also aggregates across identifiers.

\paragraph{Reported Retrieval metrics}

All metrics are macro-averaged over reporting queries. Let $R_q$ be the set
of positively judged passages, $d_i$ the passage at native rank $i$, and
$I_i=\mathbf1\{d_i\in R_q\}$. Missing positions contribute zero. We report
the following at $k\in\{10,100\}$:
\begin{align}
\operatorname{Precision}@k(q)&=\frac{\sum_{i=1}^k I_i}{k},\,\operatorname{Recall}@k(q)=\frac{\sum_{i=1}^k I_i}{|R_q|},\\
\operatorname{AP}@k(q)&=\frac1{|R_q|}\sum_{i=1}^k
 I_i\operatorname{Precision}@i(q).
\end{align}
MAP is the mean AP; its denominator remains $|R_q|$, even when
$k<|R_q|$. Reciprocal rank is $1/j$ when the first relevant passage occurs
at rank $j\leq k$, and zero otherwise; MRR is its query mean. NDCG uses
linear qrel gain,
\begin{equation}
\operatorname{DCG}@k(q)=\sum_{i=1}^k\frac{g_q(d_i)}{\log_2(i+1)},
\quad
\operatorname{NDCG}@k(q)=\frac{\operatorname{DCG}@k(q)}{\operatorname{IDCG}@k(q)},
\end{equation}
where unjudged or nonrelevant passages have grade zero, and IDCG sorts the
positive grades. This uses $g$, rather than $2^g-1$, as gain.

\paragraph{\add{Reported Reward Metrics.}} \add{We report two kinds of rewards: mean beam slot reward over trie-constrained beam search (without replacement) of width 100 and Monte Carlo reward (with replacement) by sampling 100 times.}

\add{For raw terminal slots $z_{q,1},\ldots,z_{q,100}$, the Beam reward is
\begin{equation}
G(q)=\frac1{100}\sum_{j=1}^{100}\bar Q(q,z_{q,j}).
\label{eq:minder-beam-reward-appendix}
\end{equation}
 Each raw slot has equal weight; repeated
slots retain multiplicity. Within each slot, passage IDs are deduplicated
before applying~\eqref{eq:minder-identifier-reward}.}

\add{On the other hand, for a given query $q$ we sample trajectories $\tau_1,\cdots,\tau_{100}\sim\pol(q)$ (iid sampling) and we compute the Monte Carlo Expected Reward as follows:
\begin{equation}
\hat V (q)=\frac1{100}\sum_{j=1}^{100}\bar Q(q,\tau_j).
\label{eq:minder-mc-reward-appendix}
\end{equation}
 Thus, once the actions (in the case of Generative retrieval, pseudo queries) are obtained, the reward computation is essentially the same. The key distinction lies in how the pseudoqueries are obtained: by beam search or just by sampling.}

\paragraph{\add{Beam Slot and Monte Carlo Rewards}}
\add{Here, we report both Average Beam search reward for a width of 100 and a Monte Carlo expected reward obtained by sampling from the learnt policy with replacement. We examine how $\val{\hat\pol}$ varies with the SFT anchor weight $\omega$ and the margin $\delta$ for the contrastive losses, and how they compare to the SFT reward, for both the reward measurements. All plots are averaged across 3 seeds.}

\begin{figure}[H]
    \centering
    \includegraphics[width=\linewidth]{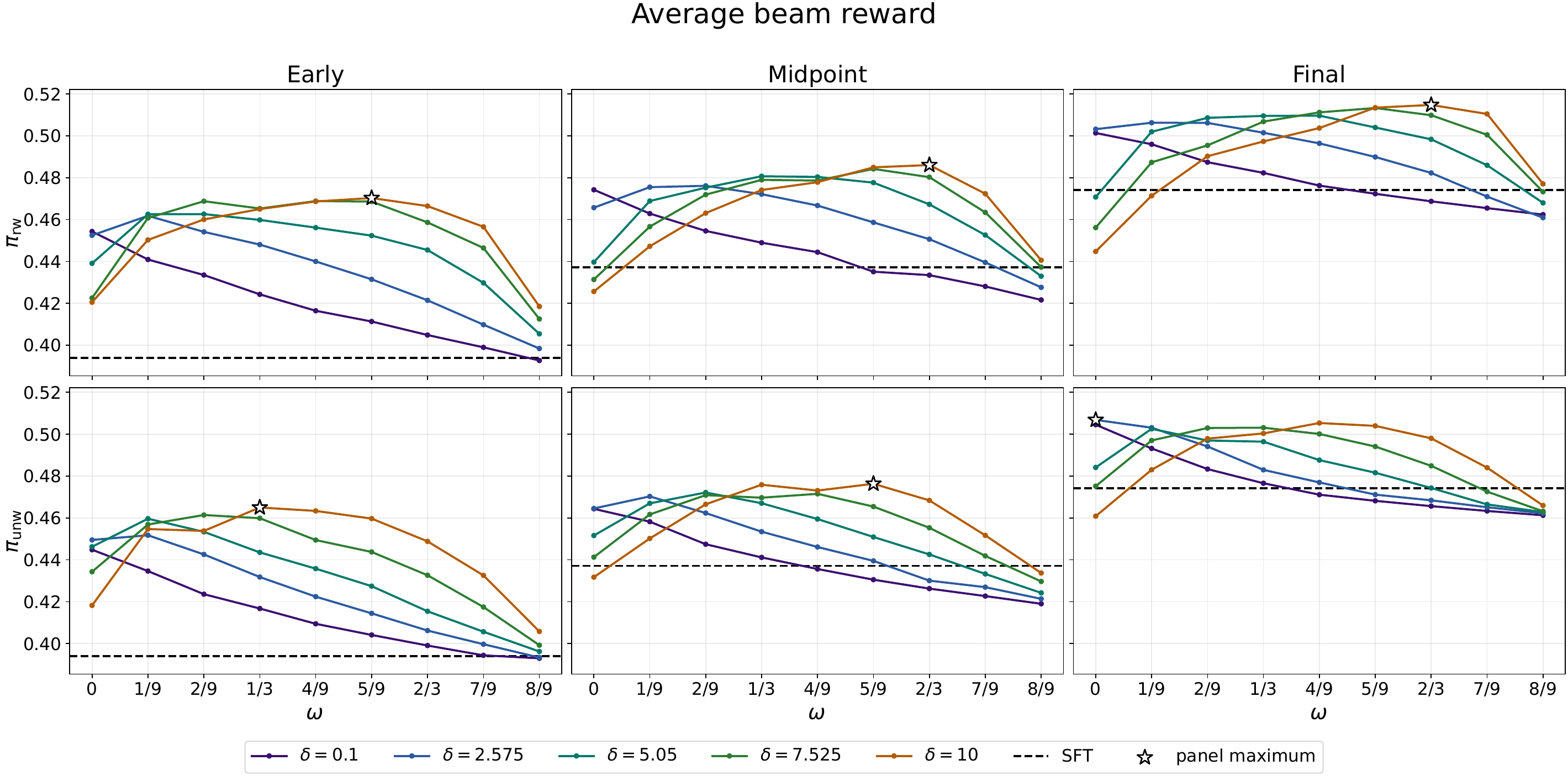}
    \caption{\add{Variation of Mean Beam Reward of the learnt policy across different initial $\base$ across 3 seeds. Each curve is one margin $\delta$. Dashed line is Reward SFT.}}
    \label{fig:beam_reward_app}
\end{figure}

\begin{figure}[H]
    \centering
    \includegraphics[width=\linewidth]{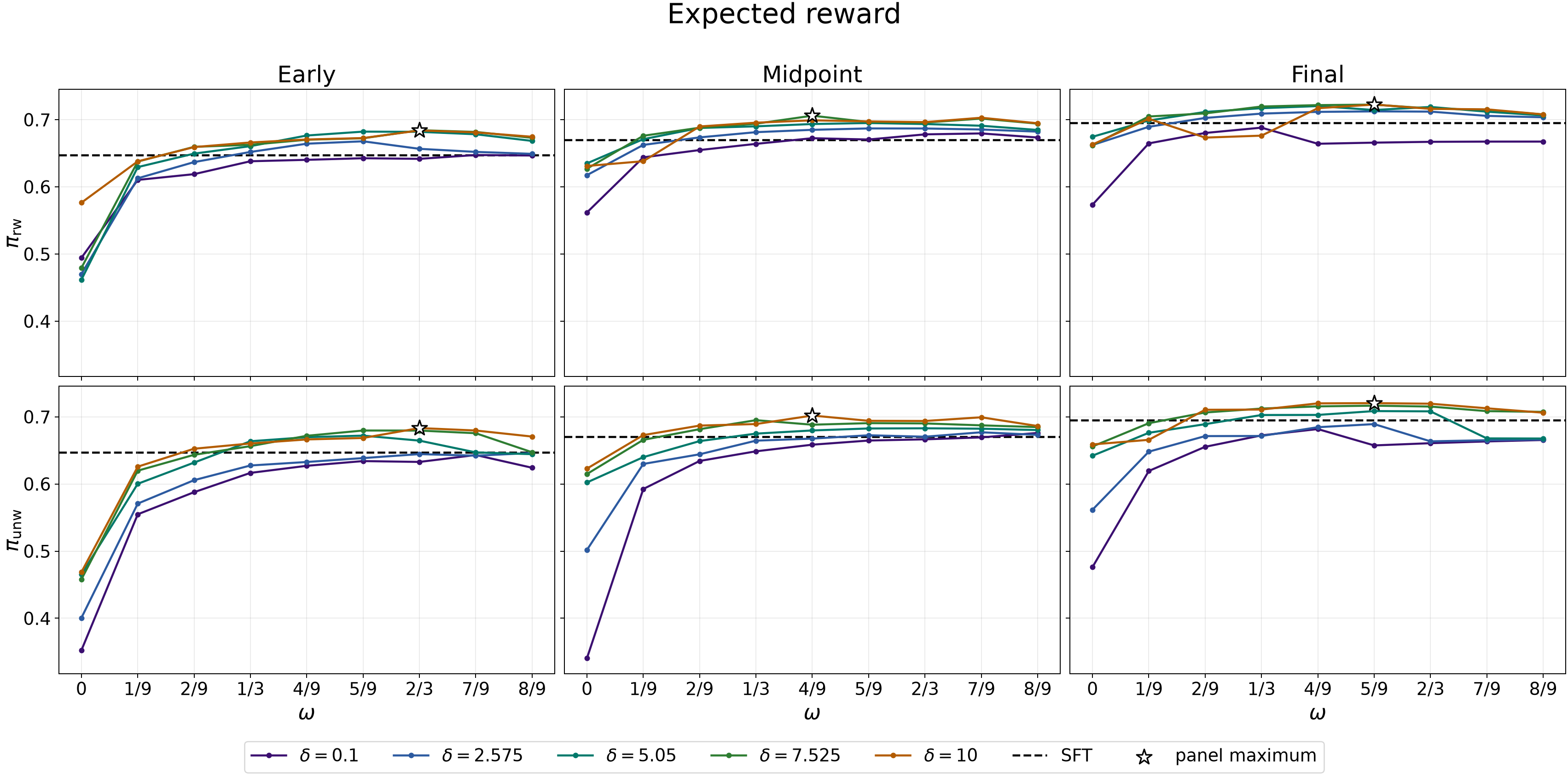}
    \caption{\add{Variation of expected reward $\val{\newpol}$ across different initial $\base$ across 3 seeds. Each curve is one margin $\delta$.Dashed line is Reward SFT.}}
    \label{fig:mc_reward_app}
\end{figure}

\paragraph{Additional ranking numbers}Our goal is to have a detailed understanding of how good the trained policies $\pirw, \piunw$ are across different ranking metrics and across the $(\lambda, \marg)$ configurations of contrastive losses and how they vary across the checkpoints. Basically we want to check on 3 axes: alignment, contrastive loss and hyperparameters. We plot the variation of ranking performance for contrastive losses across their hyperparam grids for each of the checkpoints in the following figures.

\begin{figure}[H]
    \centering
    \includegraphics[width=\linewidth]{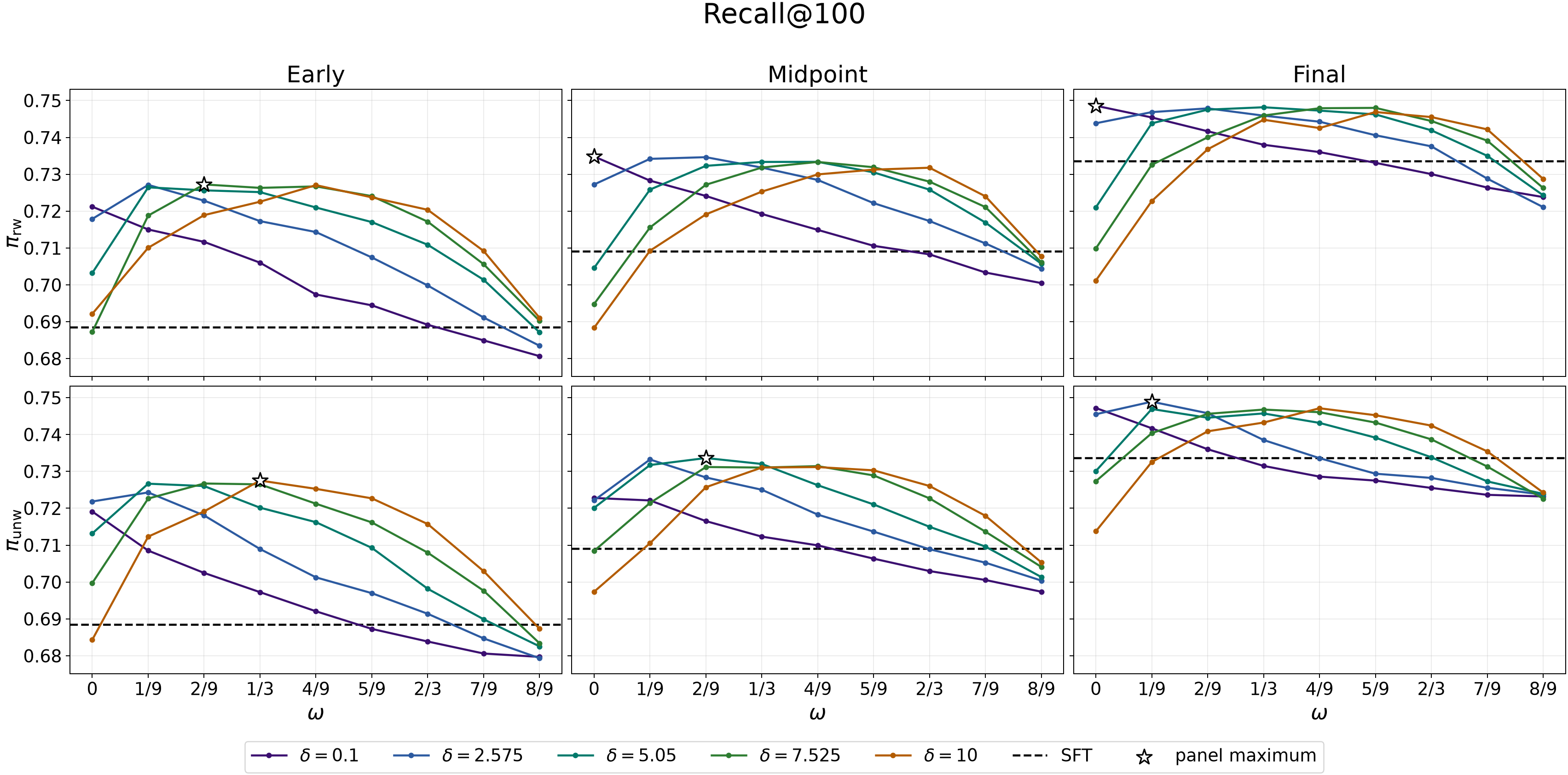}
    \caption{\add{Variation of Recall with $\omega$ across different initial $\base$ across 3 seeds. Each curve is one margin $\delta$. Dashed line is Reward SFT.}}
    \label{fig:recall_plot_minder_app}
\end{figure}

\begin{figure}[H]
    \centering
    \includegraphics[width=\linewidth]{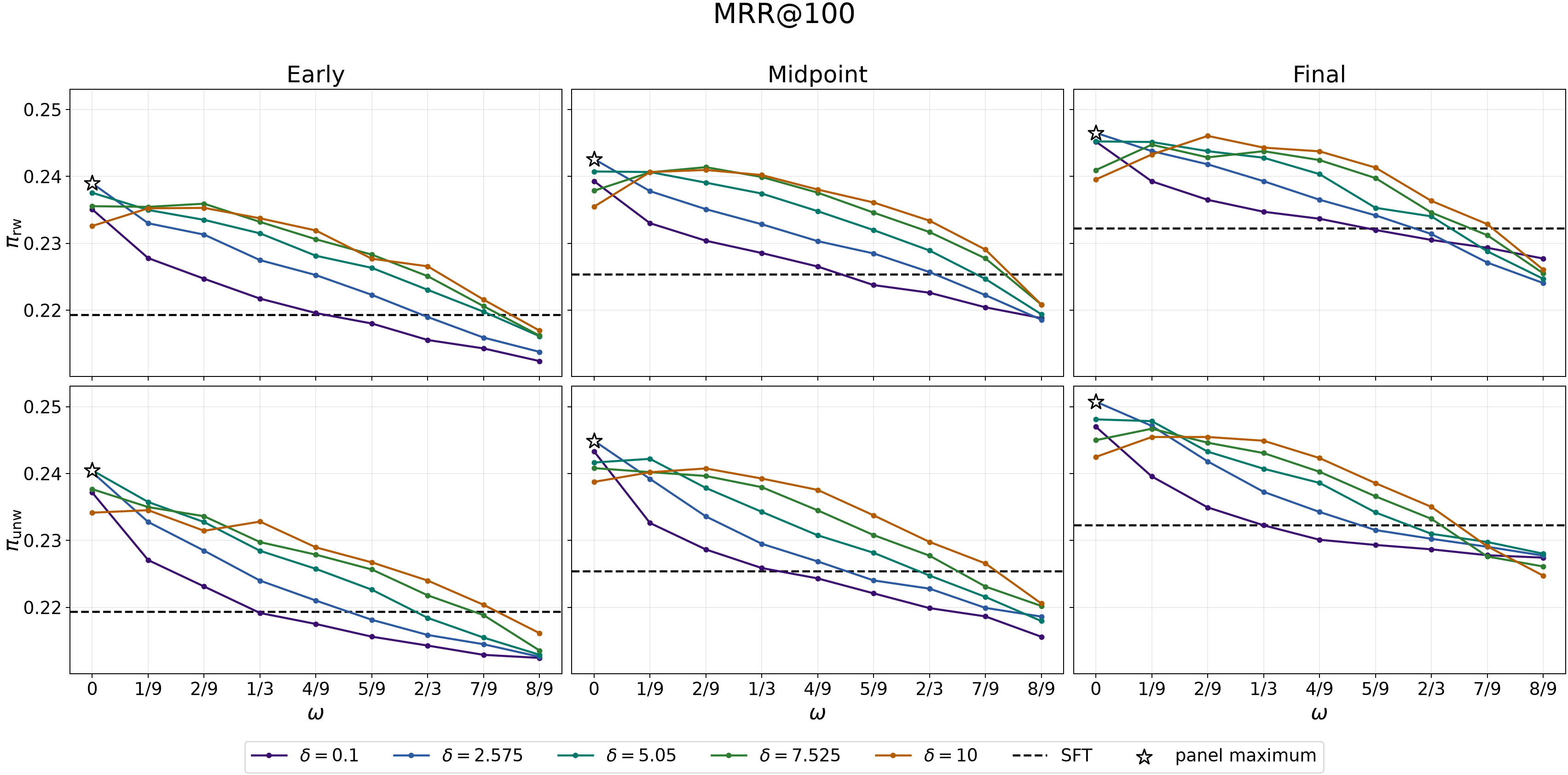}
    \caption{\add{Variation of MRR with $\omega$ across different initial $\base$ across 3 seeds. Each curve is one margin $\delta$. Dashed line is Reward SFT.}}
    \label{fig:mrr_plot_minder_app}
\end{figure}

\begin{figure}[H]
    \centering
    \includegraphics[width=\linewidth]{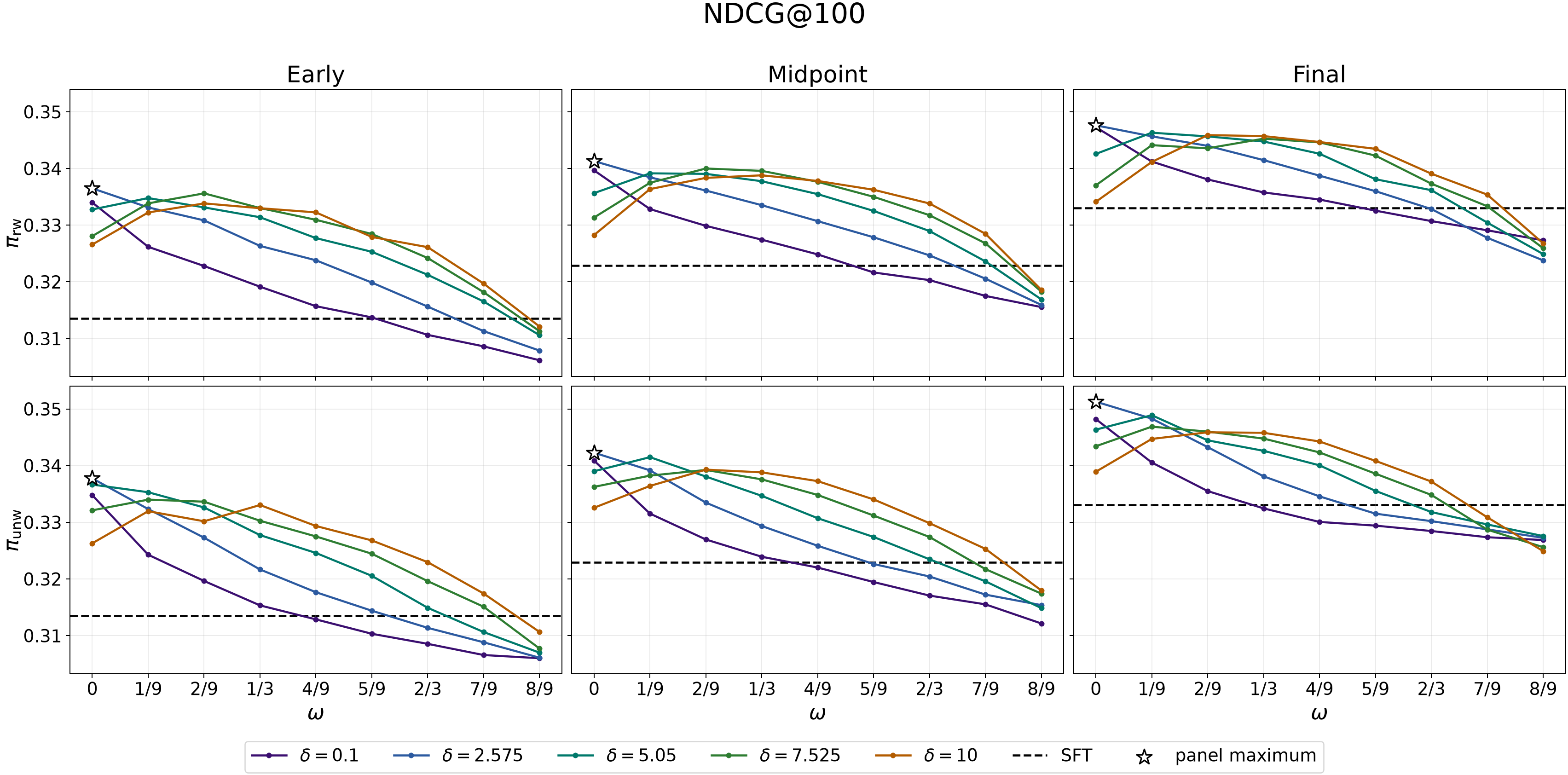}
    \caption{\add{Variation of NDCG with $\omega$ across different initial $\base$ across 3 seeds. Each curve is one margin $\delta$. Dashed line is Reward SFT.}}
    \label{fig:ndcg_plot_minder_app}
\end{figure}

\begin{figure}[H]
    \centering
    \includegraphics[width=\linewidth]{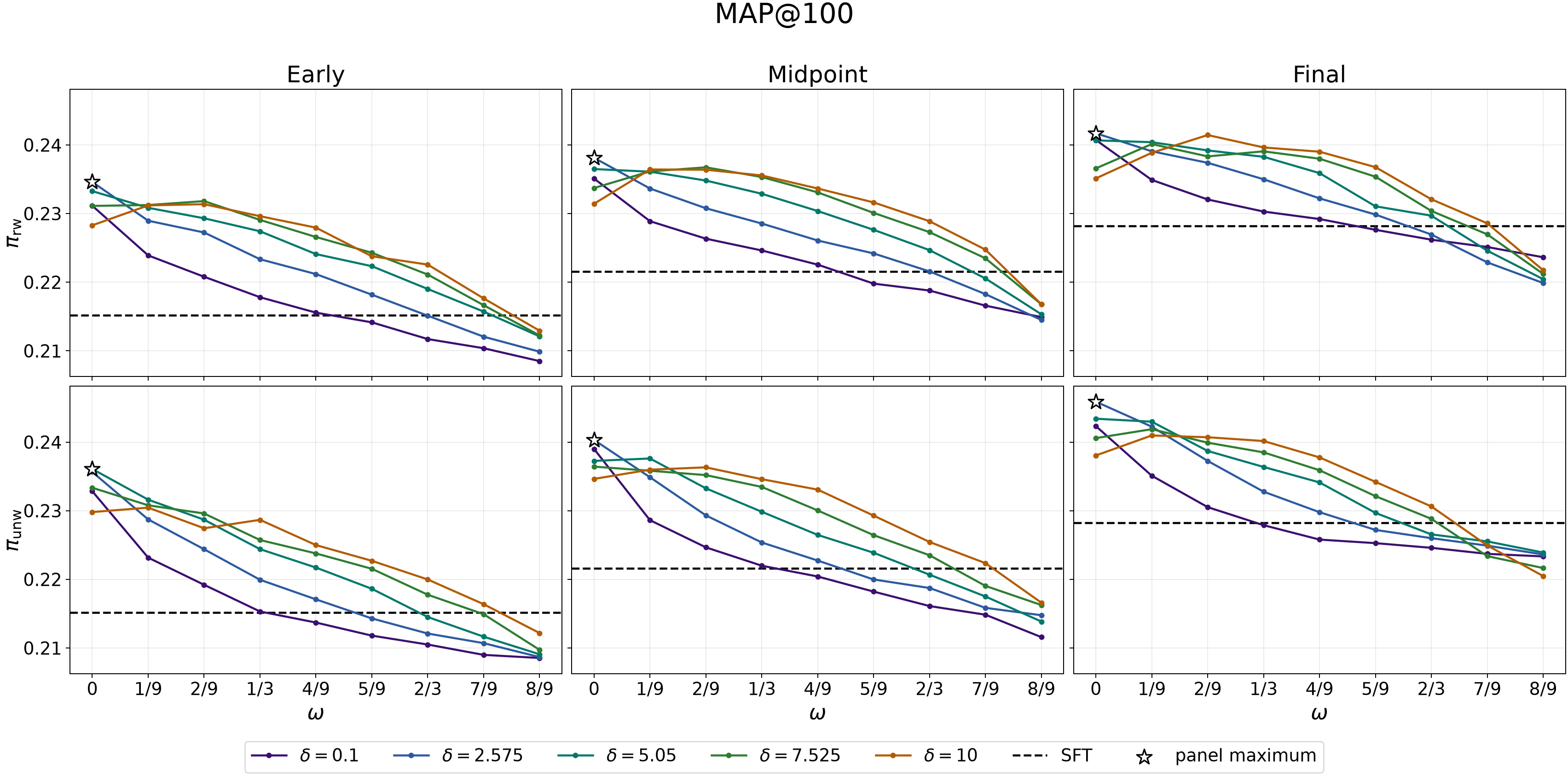}
    \caption{\add{Variation of MAP with $\omega$ across different initial $\base$ across 3 seeds. Each curve is one margin $\delta$. Dashed line is Reward SFT.}}
    \label{fig:map_plot_minder_app}
\end{figure}

\begin{figure}[H]
    \centering
    \includegraphics[width=\linewidth]{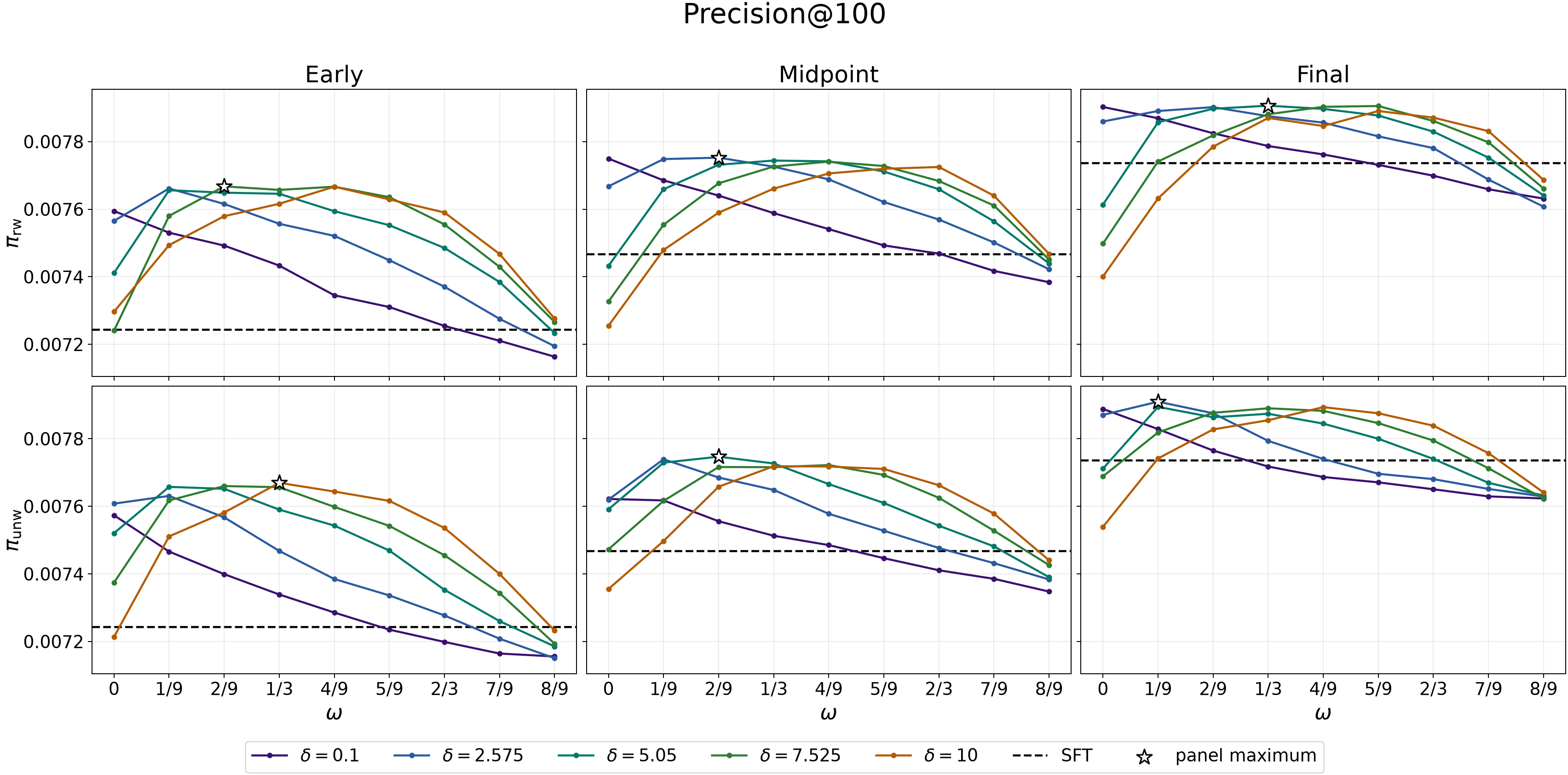}
    \caption{\add{Variation of Precision with $\omega$ across different initial $\base$ across 3 seeds. Each curve is one margin $\delta$. Dashed line is Reward SFT.}}
    \label{fig:precision_plot_minder_app}
\end{figure}

Across all ranking metrics in \cref{fig:mrr_plot_minder_app,fig:recall_plot_minder_app,fig:ndcg_plot_minder_app,fig:map_plot_minder_app,fig:precision_plot_minder_app}  we observe that \textbf{(1)} Low-margin curves degrade below $\scond$ fast, highlighting the importance of a large margin as prescribed by our theorems. \textbf{(2)} Both the contrastive losses are better than pure SFT for a significant number of hyperparameters.

For several of the ranking metrics we see that $\piunw$ is the winner, but $\pirw$ trails close, despite $\pirw$ having higher reward for midpoint and final checkpoints. This is consistent with our theoretical intuition of $\pirw$ being optimal for reward, but $\piunw$ being designed more for ranking.

\begin{figure}[H]
    \centering
    \includegraphics[width=\linewidth]{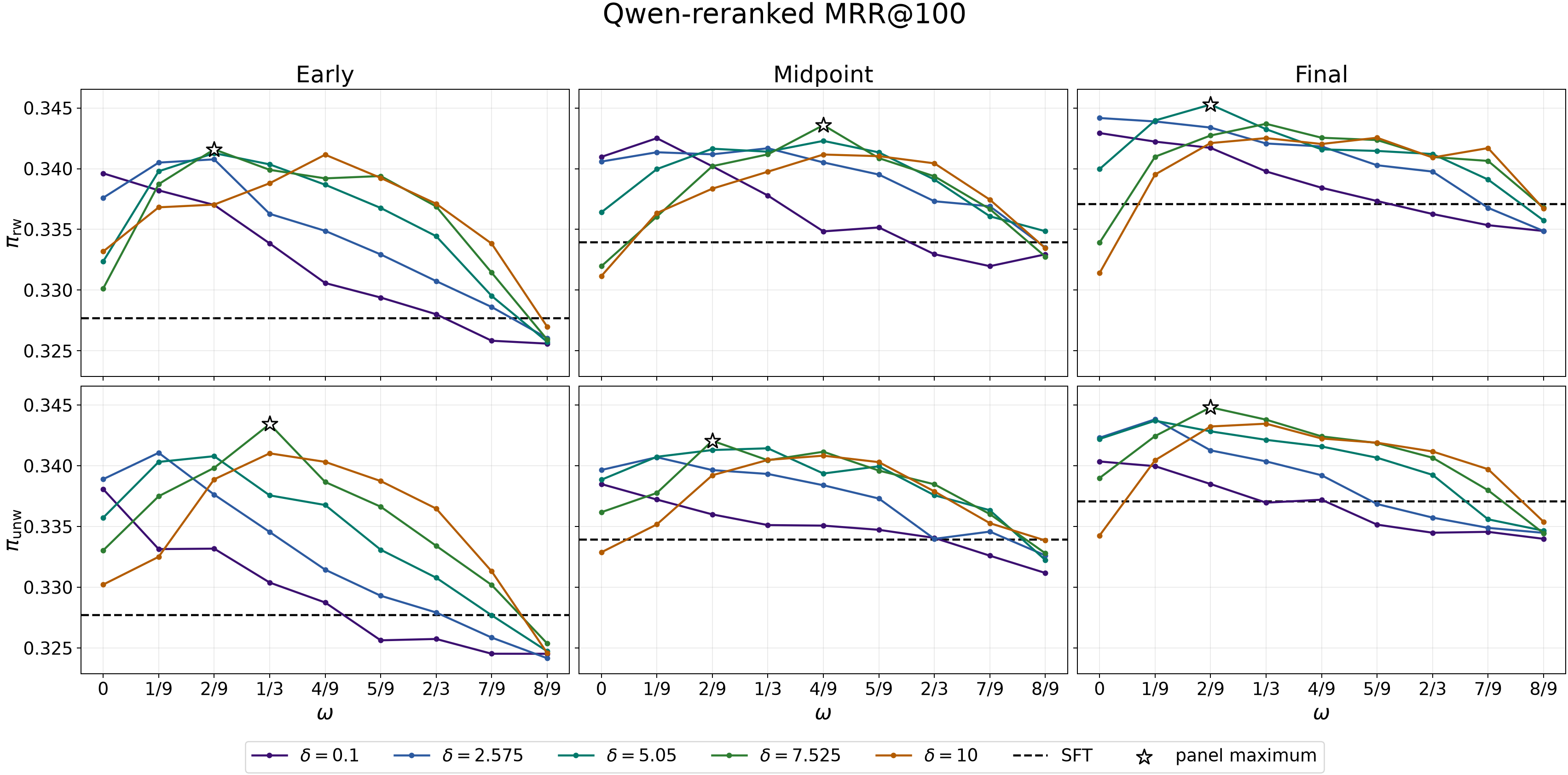}
    \caption{Variation of Qwen-reranked MRR with $\omega$ across different initial $\base$. Each curve is one margin $\delta$. Dashed line is Reward SFT.}
    \label{fig:qwen_reranked_mrr_plot_minder}
\end{figure}

\begin{figure}[H]
    \centering
    \includegraphics[width=\linewidth]{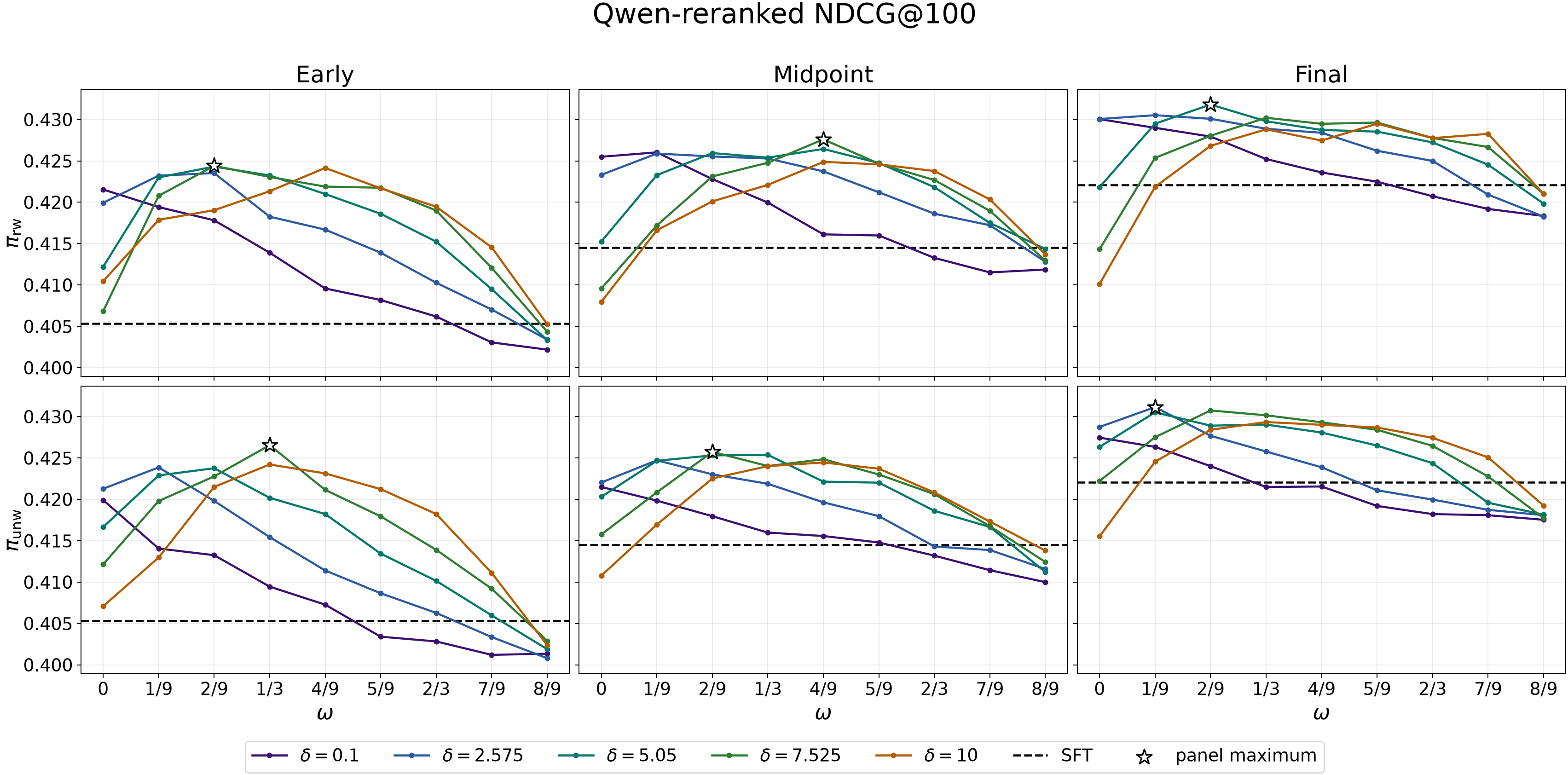}
    \caption{Variation of Qwen-reranked NDCG with $\omega$ across different initial $\base$. Each curve is one margin $\delta$. Dashed line is Reward SFT.}
    \label{fig:qwen_reranked_ndcg_plot_minder}
\end{figure}

\begin{figure}[H]
    \centering
    \includegraphics[width=\linewidth]{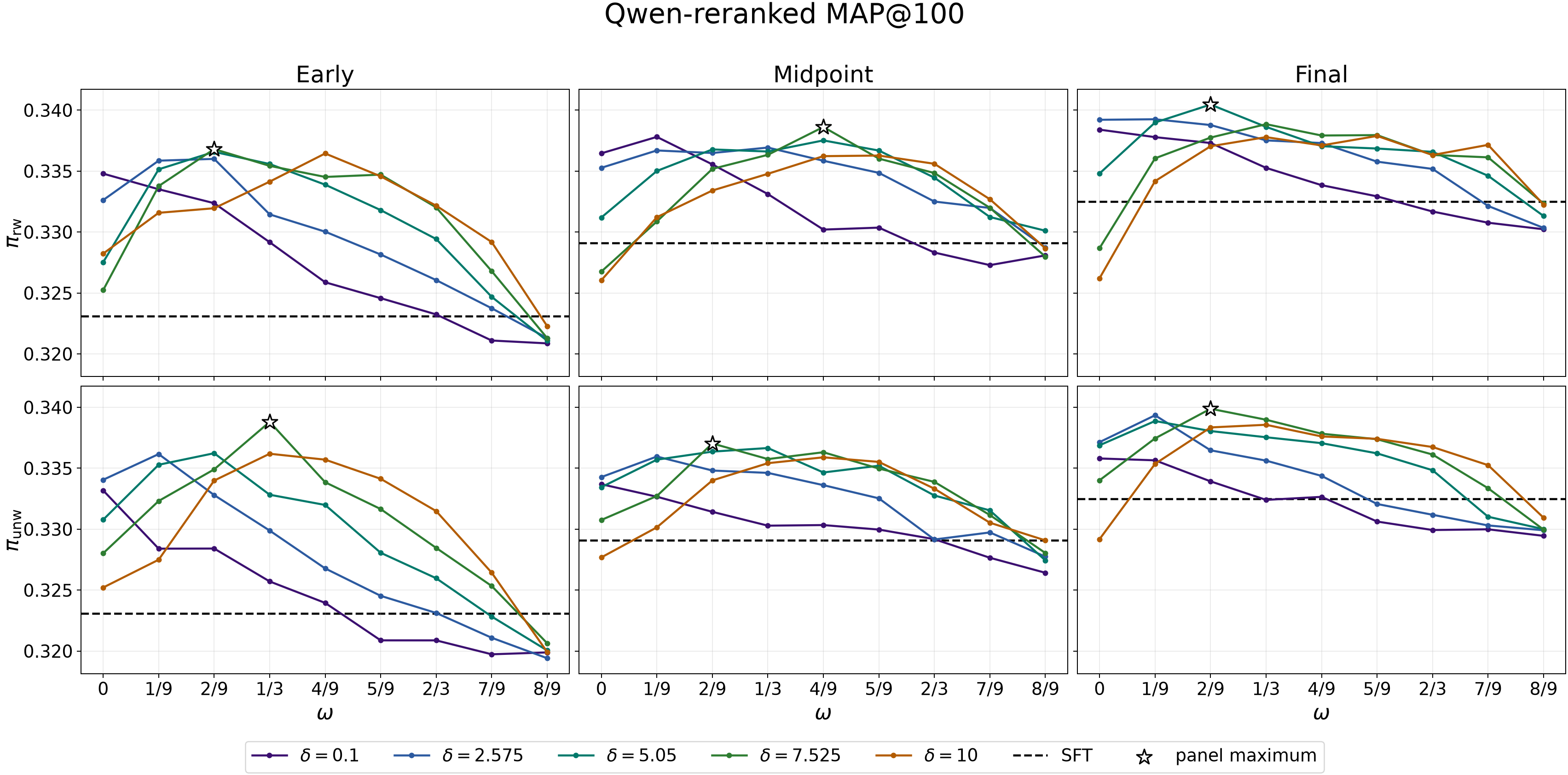}
    \caption{Variation of Qwen-reranked MAP with $\omega$ across different initial $\base$. Each curve is one margin $\delta$. Dashed line is Reward SFT.}
    \label{fig:qwen_reranked_map_plot_minder}
\end{figure}

\paragraph{Reward-based reranking.}
A retrieval metric computed on the native list conflates two things the generator does: which passages it puts into its top-100 list, and how it orders them. The second is what Theorem~\ref{thm:rankings} is about; the first is closer to the value. To look at the first alone we hand the ordering to the reward model: for each model and query we take the $100$ passages of the native list, reorder them by the beam-slot rewards $\widehat Q(q,d)$ of \cref{eq:minder-passage-reward}, and evaluate against the qrels. The candidate set is unchanged, so Recall@100 and Precision@100 are the native ones by construction, and only the order-sensitive metrics (MRR, NDCG, MAP and the @10 metrics) move.  We report only the order-sensitive MRR, NDCG and MAP of the reward-reranked version. \Cref{fig:qwen_reranked_mrr_plot_minder,fig:qwen_reranked_ndcg_plot_minder,fig:qwen_reranked_map_plot_minder} show the result on the same grid.

We make the following observation: in the reward-based reranking, the cells that produce the best sets are thus the cells with an interior $\omega$ and a large margin, the same cells that produce the highest reward, while the best native order comes from the pure hinge.

\subsection{\add{Instruction following with Qwen2.5-7B-Instruct}}
\label{app:qwen-ultrafeedback}

\paragraph{\add{Setup.}}
\add{We evaluate the objectives on instruction following using Qwen2.5-7B-Instruct and 8,000 UltraFeedback training prompts.
UltraFeedback supplies the instructions; all training responses are
generated by the frozen initial policy $\pi_0$, rather than taken
from the dataset's supplied preference pairs.}

\paragraph{\add{Reward and feedback collection.}}
\add{We use a frozen Skywork-Reward-Llama-3.1-8B-v0.2 model
to score each prompt--response pair $(x,y)$.
Its score $s_\phi(x,y)$ is transformed into a success probability
$Q(x,y)=0.05+0.90\,\sigma((s_\phi(x,y)-b)/T_r)$,
where $\sigma$ is the logistic sigmoid.
The offset $b$ and temperature $T_r>0$ are fixed using a separate
calibration split.
Each generated response occurrence receives an independent binary
terminal reward
$R\mid x,y\sim\operatorname{Bernoulli}(Q(x,y))$.
For each training prompt, we collect four successful and four
unsuccessful response occurrences.
All pairwise methods use this frozen feedback dataset, and SFT
uses the same successful responses.}

\paragraph{\add{Training and evaluation.}}
\add{
We sweep learning rates $\{5\times10^{-6},10^{-5},3\times10^{-5}\}$
and weight decay values $\{0,0.01\}$ for all four methods:
RW, UNW, SFT, and DPO.
For DPO, we additionally sweep
$\beta\in\{0.01,0.1,0.3,1\}$. Each config was given a training budget of three epochs.
For RW and UNW, we sweep the additive SFT coefficient
$\lambda\in\{0,1/4,1,4,16\}$ and hinge margin
$\delta\in\{27.8,278.1\}$.
UNW uses the score $\log\pi_\theta(y\mid x)$, while RW uses
$\log\pi_\theta(y\mid x)-\log\pi_0(y\mid x)$;
both use summed sequence log likelihoods.
We use three random seeds and select configurations and checkpoints
using separate validation prompts.
We report expected reward under the fixed reward function
on held-out test prompts:
$V(\pi)=
\mathbb{E}_{x\sim\mathcal{D}_{\mathrm{test}}}
\mathbb{E}_{y\sim\pi(\cdot\mid x)}[Q(x,y)]$.
}

\begin{table}[h]
    \centering
    \small
    \renewcommand{\arraystretch}{1.08}
    \caption{\add{Expected test reward with SEM (across 3 seeds) for the validation-selected
    configuration of each method on the instruction-following task.
    Bold marks the highest reported reward.}}
    \label{tab:qwen-main-reward}
    \add{\begin{tabular}{lc}
        \toprule
        Method & Expected reward \\
        \midrule
        Base policy
            & $0.525 \pm 0.001$ \\
        SFT
            & $0.547 \pm 0.001$ \\
        DPO
            & $0.588 \pm 0.001$ \\
        RW
            & $\mathbf{0.636 \pm 0.001}$ \\
        UNW
            & $0.592 \pm 0.001$ \\
        \bottomrule
    \end{tabular}}
\end{table}

\paragraph{\add{Comparison with baselines.}}
\add{Table~\ref{tab:qwen-main-reward} reports the test rewards of
the validation-selected configurations.
RW achieves the highest reported reward, exceeding SFT by $0.089$
and DPO by $0.048$.
UNW exceeds SFT by $0.045$ and DPO by $0.004$.}

\begin{table}[h]
    \centering
    \small
    \setlength{\tabcolsep}{5pt}
    \renewcommand{\arraystretch}{1.08}
    \caption{\add{Expected test reward across SFT coefficients and hinge
    margins. RW and UNW use the same frozen feedback dataset.
    Bold marks the highest reported reward in each column.}}
    \label{tab:qwen-margin-grid}
    \add{\begin{tabular}{ccccc}
        \toprule
        & \multicolumn{2}{c}{RW}
        & \multicolumn{2}{c}{UNW} \\
        \cmidrule(lr){2-3}
        \cmidrule(lr){4-5}
        SFT coefficient $\lambda$
        & $\delta=27.8$
        & $\delta=278.1$
        & $\delta=27.8$
        & $\delta=278.1$ \\
        \midrule
        $0$
        & $\mathbf{0.636 \pm 0.001}$
        & $0.509 \pm 0.001$
        & $\mathbf{0.560 \pm 0.001}$
        & $\mathbf{0.592 \pm 0.001}$ \\
        $1/4$
        & $0.580 \pm 0.001$
        & $\mathbf{0.581 \pm 0.001}$
        & $0.555 \pm 0.001$
        & $0.574 \pm 0.002$ \\
        $1$
        & $0.558 \pm 0.002$
        & $0.554 \pm 0.001$
        & $0.559 \pm 0.001$
        & $0.560 \pm 0.001$ \\
        $4$
        & $0.553 \pm 0.001$
        & $0.550 \pm 0.001$
        & $0.547 \pm 0.002$
        & $0.554 \pm 0.001$ \\
        $16$
        & $0.545 \pm 0.001$
        & $0.549 \pm 0.002$
        & $0.543 \pm 0.001$
        & $0.548 \pm 0.001$ \\
        \bottomrule
    \end{tabular}}
\end{table}

\paragraph{\add{Effect of the margin and SFT coefficient.}}
\add{Table~\ref{tab:qwen-margin-grid} shows that the effect of the
margin differs between the two objectives.
For UNW, the larger margin gives a higher reported reward at
every tested SFT coefficient.
For RW, the effect depends on $\lambda$.
In particular, at $\lambda=0$, increasing the margin reduces
reward from $0.636$ to $0.509$.}

\add{The high-margin RW results also show a benefit from combining
contrastive training with SFT.
At $\delta=278.1$, adding an SFT anchor with $\lambda=1/4$
raises reward from $0.509$ to $0.581$, exceeding both
the pure-contrastive result at that margin and the SFT
baseline of $0.547$.
The best overall RW result nevertheless occurs at the smaller
margin with $\lambda=0$.}